\documentclass[twocolumn]{article}

\usepackage{moreverb,url}

\usepackage[colorlinks,bookmarksopen,bookmarksnumbered,citecolor=red,urlcolor=red]{hyperref}

\usepackage[utf8]{inputenc}

\usepackage{graphicx}
\usepackage{multirow}
\usepackage{natbib}

\usepackage{appendix}
\usepackage{xcolor}
\usepackage{comment}

\usepackage{pdfpages}
\newcommand\BibTeX{{\rmfamily B\kern-.05em \textsc{i\kern-.025em b}\kern-.08em
T\kern-.1667em\lower.7ex\hbox{E}\kern-.125emX}}

\newcommand{\ie}{\textit{i}.\textit{e}.}
\newcommand{\eg}{\textit{e}.\textit{g}.}

\makeatletter
\let\@fnsymbol\@arabic
\makeatother

\begin{document}

%\runninghead{Takamatsu, Saito, Ikeuchi, Kanehira, Sasabuchi, Wake, and Wu}

%\title{Designing skill library for hardware-level re-usability}
\title{Designing Library of Skill-Agents for Hardware-Level Reusability}

\author{Jun~Takamatsu\thanks{Applied Robotics Research, Microsoft, Redmond, WA, 98052, USA} \and  Daichi~Saito\thanks{Applied Robotics Research, Microsoft, Tokyo, Japan} \and Katsushi~Ikeuchi$^1$ \and  Atsushi~Kanehira$^2$ \and Kazuhiro~Sasabuchi$^1$ \and Naoki~Wake$^1$} % and Tommy~Wu\affilnum{3}}
%\thanks{Applied Robotics Research, Microsoft, Redmond, WA, 98052, USA \\ \affilnum{2}Applied Robotics Research, Microsoft, Tokyo, Japan}
\date{}

\maketitle

\begin{abstract}

To use new robot hardware in a new environment, it is necessary to develop a control program tailored to the specific robot in the environment. Considering the reusability of software among robots is crucial to minimize the effort involved in this process and maximize software reuse across different robots in different environments. Although recent generative AI has made it possible to automatically generate some software, robot programs are still difficult to generate because of robot's physical interaction with the environment. This paper proposes a method to remedy this process by considering hardware-level reusability, using a Learning-from-observation (LfO) framework with a pre-designed skill-agent library. 
The LfO framework represents the required actions in hardware-independent representations, referred to as task models, from observing human demonstrations, capturing the necessary parameters for the interaction between the environment and the robot~(\cite{ikeuchi2021semantic}). When executing the desired actions from the task models, a set of skill agents is employed to convert the representations into robot commands. This paper focuses on the latter part of the LfO framework, utilizing the skill-agent set to generate robot actions from the task models, and explores a hardware-independent design approach for these skill agents. These skill agents are described in a hardware-independent manner, considering the relative relationship between the robot's hand position and the environment. As a result, it is possible to execute these actions on robots with different hardware configurations by simply swapping the inverse kinematics solver. This paper, first, defines a necessary and sufficient skill-agent set corresponding to cover all possible actions, and considers the design principles for these skill agents. %  in the library. 
We provide concrete examples of such skill agents and demonstrate the practicality of these skill agents by showing that the same representations can be executed on two different robots, Nextage and Fetch, using the proposed skill-agent set.

\end{abstract}

\section{Introduction}
Robot developers develop various types of robots, such as bipedal robots and mobile manipulators for satisfying users' various demands. If we consider manipulation aspects, there are various types: a single-arm robot or a dual-arm robot, and the degrees of freedom (DOF) of an arm is from 5 (\eg, HSR, Toyota) to 7 (\eg, Fetch Mobile Manipulator, Fetch Robotics). Some robots have additional DOF on their waist.

Users' demands are related to their backgrounds and robots suitable for users may vary. 
If a certain developer would adopt a new robot from the previously used one, robot-specific software has to be changed. On the other hand, robot-software developers would like to reuse their developed software as much as possible to reduce their efforts. It is desirable to satisfy those two conflicting demands.

We have been developing a robot system based on a {\em Learning-from-Observation (LfO)} framework (\cite{ikeuchi2021semantic})\footnote{In the machine-learning (ML) community, Learning-from-Observation (LfO) is used with a slightly different definition. However, following the rationale mentioned in the introduction of \cite{ikeuchi2021semantic}, LfO is specifically referred to as a method that transforms input into symbolic representations based on top-down knowledge and subsequently maps them to robot actions.}. Our goal is to have a robot reproduce the target behavior by simply demonstrating it in front of the robot.
Thus, even users without robotics knowledge can have the robot perform tasks that they desire. Unlike similar frameworks, such as learning-from-demonstration and imitation learning (\cite{schaal1999imitation,schaal2003imitation,billard2008robot,asfour2008imitation,dillmann2010advances,akgun2012keylfd}), the LfO system preforms indirect behavior mimicry by first a task-encoding step converting the demonstration into an abstract intermediate representation, referred to as a task model, and then a task-decoding step converting this intermediate representation into the behavior of each robot. The task-encoding recognizes behavior as symbolic task sequences (such as a sequence of pick up, place, and release actions) and then extracts from the demonstration the parameters that are pre-defined in each task (such as where to grasp and where to put). %Hereafter, a task is considered as a single unit action, such as pick up, etc., and the entire actions in behavior is considered as a task sequence. 
Because only task-specific information is extracted from the demonstration, unnecessary parts are ignored, allowing the demonstrator to focus only on the important demonstration parts that need to be taught. For task-decoding, we prepare in advance the skill agents, which correspond to the agents performing symbolic actions, following the design of the task model. Then, the reproduction by a robot is realized by activating the corresponding skill agent with the observed parameters.

This paper focuses on the second step of the LfO, to build the robot-independent task-decoder from task models to actions. For task-decoding, hardware-independent skill agents corresponding to each task are pre-designed and stored in the library. These skill agents represent the tasks only by hand motions to absorb the structural difference between each robot, and the robot arm and body is regarded as a carrier used to move the hand on the desired trajectory. %The hand motions corresponding to each skill are trained locally using reinforcement learning (RL) with a reward function that accounts for the drag force generated by the interaction with the environment in a desired manner. 
Given a target trajectory by a skill agent, a general inverse kinematics (IK) solver, body role division~(\cite{SasabuchiRAL2021}), is used to determine the robot's body motion to achieve this trajectory. % while maintaining reusability for a variety of robots. 

The contribution of this paper is threefold:
\begin{itemize}
    \item definition of a robot-independent, necessary and sufficient skill-agent set for manipulation tasks involving force/visual feedback

    \item proposal for design principles of skill agents and implementation examples of a skill-agent set using the principles
    
    \item demonstration of a reusable system using the skill-agent set
\end{itemize}

\section{Related work}

Efforts to increase reusability of robot programs, such as {\em Robot Operating System} (ROS)~(\cite{Quigley2009ROSAO}) and {\em OpenRTM}~(\cite{openrtm}), have been conducted thus far. These two pieces of middleware follow a so-called {\em subsumption architecture}~(\cite{Brooks}). In this architecture, a robot program are created by combining several nodes. The nodes are properly connected and communicate with each other to execute the program. These two pieces of middleware achieved reusability by 1) unifying the format of communication and 2) providing means of communication (\eg, publisher and subscriber in ROS). Switching between low-level nodes (\eg, nodes to output sensor reading) is very easy.

In high-level nodes, it is necessary to take into account the individual characteristics of the robot hardware. 
For example, to move a mobile robot on a floor, we can define a de-facto standard robot command (\eg, a pair of velocity and angular velocity) and combinations of nodes (the occupancy grid map~(\cite{MoravecICRA1985}) and Monte Carlo localization~(\cite{Dellaert-1999-14893})). These two pieces of middleware provide open-source nodes for various robots. Conversely, if we could define the robot-independent action representation, hardware-level reusability can be increased by converting that representation into a robot-specific control signal. \cite{Bachiller-Burgos2020} proposed the STEM education programming tool where the robot hardware can be changed without modifying programs created by students. In the hardware-level reusablity, the robot-independent representation and their conversion play important roles.

In manipulation aspects, which are our main targets, the control strategy of the manipulator differs from situations, such as simple position control, impedance control~(\cite{Hogan1984}), and machine-learning-based control~(\cite{JIN201823}). But to realize manipulation at a minimum, the end-effector must be brought to the desired position. % without colliding with any obstacles.
To bring to the desired position, it is necessary to decide on the robot joints to satisfy the target end-effector position. 
IK solver is a one of the solution and many proposals and implementations for IK are available, such as~\cite{Beeson2015,starke2017memetic}. \cite{cheng2018rmpflow} developed the globally stable controller within non-Euclidean spaces. They succeeded in the reactive motion generation (\eg, avoid obstacles) in real-time. The proposed skill agents output the target configuration of the end-effector and any methods to generate motions from the output are acceptable. In the sense of the hardware-level reusability,  \cite{pyrobot2019} proposed the open-source robotics framework that provides hardware independent mid-level APIs including FK/IK, robot vision, and planning for low-code development. By further considering task models and task decoders, this paper aims to provide a hardware-unaware robot programming environment.
%For collision avoidance, which is out of our main concern, the random-sampling-based motion planning, such as the probabilistic roadmap method~(\cite{KavrakiITRA1996}), Rapidly-exploring Random Trees~(\cite{LavalleAFR2000}), and their variants, are often used as a de-facto standard. And ROS also provides the comprehensive motion planning framework, {\em MoveIt}\footnote{https://moveit.ros.org/}. % These pieces of software require the target positions of the end-effector as input to achieve task sequences. That requires another software to generate these positions. Our main concern is to develop such software in order to achieve task sequences considering hardware-level reusability.

Recently, several papers showed the importance on the awareness of action primitives, which correspond to tasks/skills in the proposed system, in the learning-from-demonstration framework. \cite{LinRAL2022} succeeded in the complicated tasks with multiple action primitives by training each primitive independently and combining them with sub-goals. \cite{doi:10.1126/scirobotics.aay4663} succeeded to open medicine bottles with different locking mechanisms by learning discrete haptic states and grammatical representation of action primitives. The proposed system will prove that the goal oriented action primitives contribute to the hardware-level reusability, too.

\section{Designing re-usable skill-agent library}

\subsection{Overview}

This section aims to design the skill-agent library that enables hardware-level reuse. We assume that the task sequence starts with grasping a target object, continues by manipulating it, and ends to release it. Within this paper, we primarily focus on manipulation skill agents, assuming the grasping skill agents can be addressed through a separate methodology presented in~\cite{saito2022taskgrasping}. Therefore, we assume that the manipulated object is already grasped, and the object and hand are integrated. Assuming a robot with redundant DOF %degrees of freedom 
and a situation where hand motions and arm motions can be independently resolved through the body role division~(\cite{SasabuchiRAL2021}), we focus solely on hand motions when designing each skill agent.

In this paper, two terms, {\em task} and {\em skill}, are frequently used. A task is defined as a unit operation of what-to-do, obtained in the encoding part of LfO and on the decoding side, the procedure for a robot to perform this is referred to as a skill and the skill agent are responsible for the execution of a skill. Tasks and skills correspond one-to-one under an assumption of usage of one particular robot hardware, and in this paper, unless otherwise specified, they are used interchangeably without confusion. This paper aims to remove this constraint, \ie, differences of skills among robots, and to design skill agents that can be reused among a wide range of robot hardware. In each skill agent, the object displacement due to the hand motion is calculated to satisfy constraints from the environment. The displacement in unconstrained subspace is assumed to be obtained from the demonstration. The parameters required to perform these skills are called {\em skill parameters}.

In this paper, we assume that the environment between in demonstration and in robot execution is not dramatically changed. The demonstration includes a hint for collision avoidance and a robot can execute the target task by following the demonstration. Of course, we admit a slight difference in the environment in robot execution; the skill agent can absorb the difference using sensor feedback.  

%We implemented the skill library, a collection of skills, in advance. Each skill corresponds to each task and given the skill parameters and robot states (\eg, joint state and tactile), each skill outputs the target hand configuration in each control step. Most of the skills were implemented as the reinforcement-learning (RL) agent; when training, the RL agent learns the motion not to violate the constraint (\eg~avoiding infeasible displacement) using force feedback and visual feedback~\footnote{For this purpose, we prepared an environment for parallel reinforcement learning using OpenAI PPO. Details are defferred to another paper.}. In short, given the skill parameters and robot states, skills return the target hand configuration in each control step.

The grasped object is in contact with the environment. A single operation of the robot, consisting of translational and rotational motions, induces contact-state transitions between the object's surface and the environment. 
%As described earlier, we refer to this single robot operation as a {\it task}, and the program to execute it on actual hardware as a {\it skill}. Under the assumption of reusability, one skill corresponds to one task. 
%In other words, in this paper, tasks and skills are considered interchangeable. 
In a previous paper (\cite{ikeuchi2021semantic}), we defined the necessary and sufficient set of tasks that should be prepared as robot manipulations based on translational and rotational transitions. In this paper, we design skill agents to execute these defined tasks, assuming force feedback and visual feedback. During the design of each skill agent, we utilize changes in forces from the environment and/or visual features of the environment according to surface contact transitions and derive reward functions that ensure successful transitions using them. These skill agents are pre-trained through reinforcement learning based on these reward functions.

Traditionally, reinforcement learning (RL) has been employed to adjust the trajectory of a robot's hand, taking into account drag forces from the environment. Previous RL methods have predominantly focused on the design of reward functions for specific operations % , limiting their applicability 
and requiring individual learning for each new operation coming. We propose grouping multiple operations based on the types of physical constraints, deriving general guidelines, and proposing reward functions applicable to various operations based on these guidelines.

In the following discussion, to maintain continuity with the previous paper, we will first revisit the necessary concepts. Subsequently, we will outline the design principles and then proceed to the design of each skill agent.

\subsection{Preliminary}
\label{label:preliminary}

\subsubsection{Surface contact and contact transition}
\begin{figure}
    \centering
    \includegraphics[width=\linewidth]{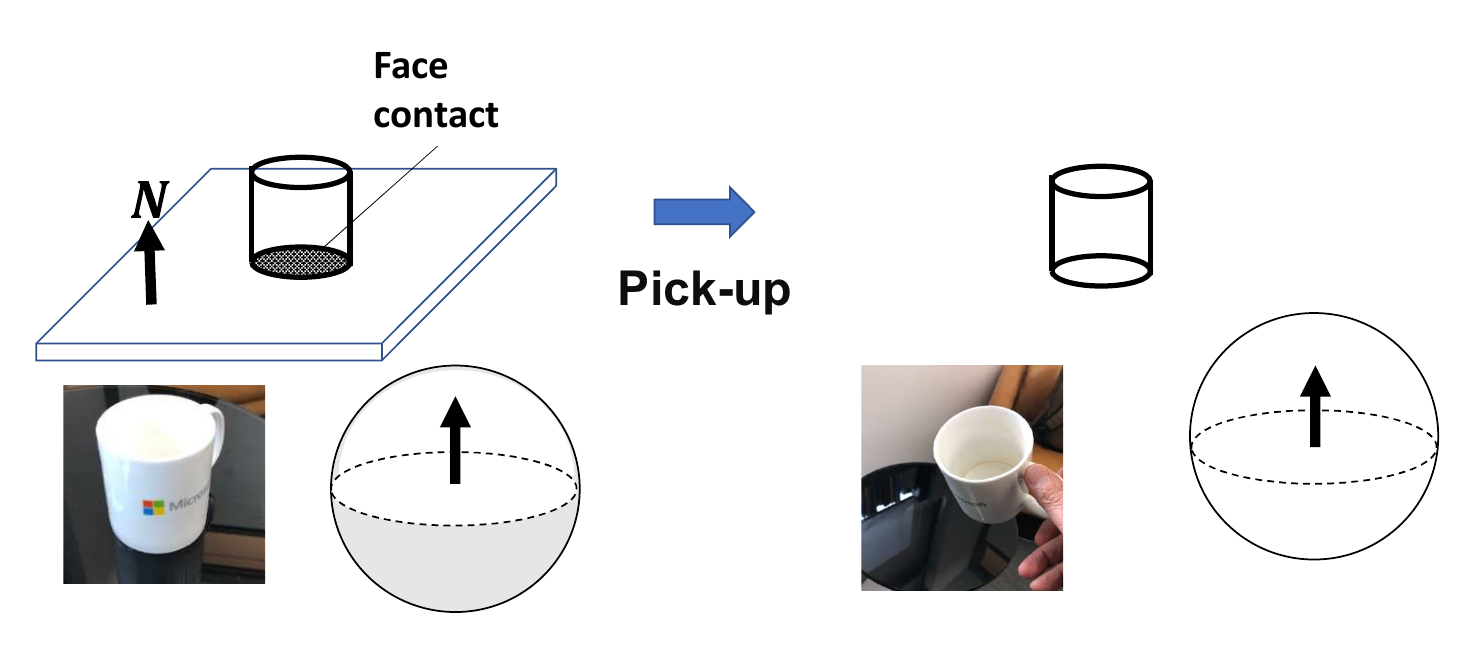}
    \caption{An example of a pick-up task. Object states can be defined by possible directions of motion for the object. The distribution of possible directions of motion can be represented as a region on a Gaussian sphere. When an object sits on a table, the Northern hemisphere represents the possible directions for the object, with the normal direction of the table as the North Pole of the sphere. The pick-up task can be defined as one causing the transition of the region from a hemisphere to a whole sphere.}
    \label{fig:pickup}
\end{figure}

The grasped object and the environment come into contact at the object's surface and environmental constraint points, resulting in restriction of possible motion directions of the object. 
\if 0
As detailed in the previous paper, the range of possible motion directions under these constraints can be expressed as the solution space of nonlinear simultaneous inequalities based on the Screw theory~\cite{roth}. According to Kuhn-Tucker theory~\cite{kuhn1957linear}, these solution spaces can be categorized into seven states. In this paper, we follow this categorization, defining tasks and skill agents as transitions between these states (including intrastate transitions). This definition serves as the starting point for the discussion, and the state transitions depicted in the graphs lead to the definition of 28 skill agents and tasks~\footnote{While some branches that rarely appeared in the manipulation operations were omitted for convenience in the previous paper, in this paper, we theoretically defined even those rare operations, resulting in a slightly increase number of branches.}.

We assume that the sequence of actions performed by a robot begins with grasping an object, manipulates it, and ends with releasing it. In other words, the units of the sequence of actions consist of grasping, multiple manipulations, and releasing. In this paper, we focus on standardizing the manipulation part across robot.
\fi
One unit of manipulation actions, \ie, one task in our terminology, can be defined as one that causes one transition in the motion constraint state of the object grasped. For example, when picking up an object on a table, in the initial state, the possible directions of motion for the object are limited to above the table surface. % due to the table's surface. 
Representing the possible motion directions on the Gaussian sphere, with the normal direction of the table surface as the North Pole, the Northern hemisphere, depicted as a white region in the left sphere in Figure~\ref{fig:pickup}, represents the possible motion directions. After picking up, there are no constraints from the environment, and the motion is possible in all directions, corresponding to the entire surface of the Gaussian sphere in the right side of Figure~\ref{fig:pickup}. The pick-up task can be defined as one to cause the transition of the motion constraint state of the object from a Hemispherical-constraint state to a No-constraint state. 

The constraints on the translational and rotational motion of an object, given by a contact point $\mathbf{p} $, can be expressed using the screw theory (\cite{roth1984screws}):

\begin{equation}
        \mathbf{n} \cdot \mathbf{t} + (\mathbf{p} \times \mathbf{n}) \cdot \mathbf{s}  \geq 0,
        \label{eq: screw}
\end{equation}
where $\mathbf{n}$ denotes the normal vector at the contact point and $\mathbf{s}$ denotes the screw axis vector. A translational motion occurs along $\mathbf{s}$, and a rotational motion occurs around $\mathbf{s}$. When the ratio between the translation and rotation is defined by the parameter $p$, $\mathbf{t} \equiv \mathbf{c} \times \mathbf{s} + p \mathbf{s}$, where $ \mathbf{c} $ is the center of rotation. Namely, one pair of an object surface and an environment contact point provides one linear inequality for the constraints on object motion.
In this paper, following the approach of the previous paper, we assume that robot manipulation involves only pure translation or pure rotation. We do not consider compound operations involving both. In the following, we will divide the analysis into translational or rotational motion, with the main analysis being translation.
\if 0
Regarding translation, we obtain:
\begin{eqnarray}
    \vec{N} \cdot \vec{S} & \geq & 0, \nonumber 
\end{eqnarray}
where $\vec{N}$ denotes the normal direction.

For rotation, we obtain:
\begin{eqnarray}
\vec{M} \cdot \vec{S} & \geq & 0
\nonumber
\end{eqnarray}
where $\vec{M} = (\vec{Q} \times \vec{N})$ is the normal vector to the plane spanned by $\vec{Q}$ and $\vec{N}$.  
See ~\cite{ikeuchi2021semantic} for detailed derivation.
\fi

\begin{figure}[ht]
    \centering
    \begin{tabular}{cc}
    \begin{minipage}{0.45\hsize}
    \includegraphics[width=\linewidth]{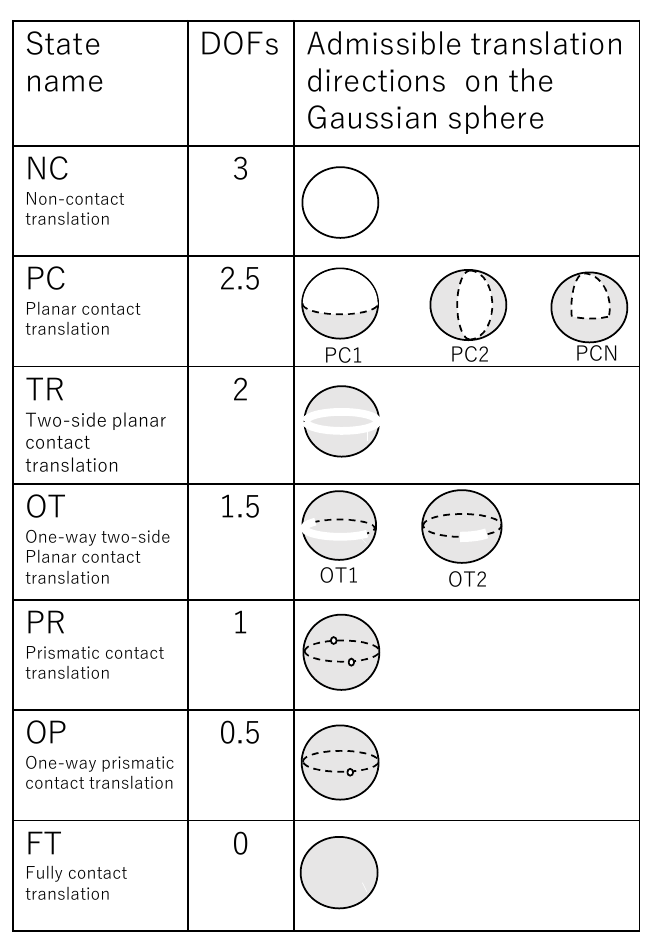}
    \end{minipage}
    &
    \begin{minipage}{0.45\hsize}
    \includegraphics[width=\linewidth]{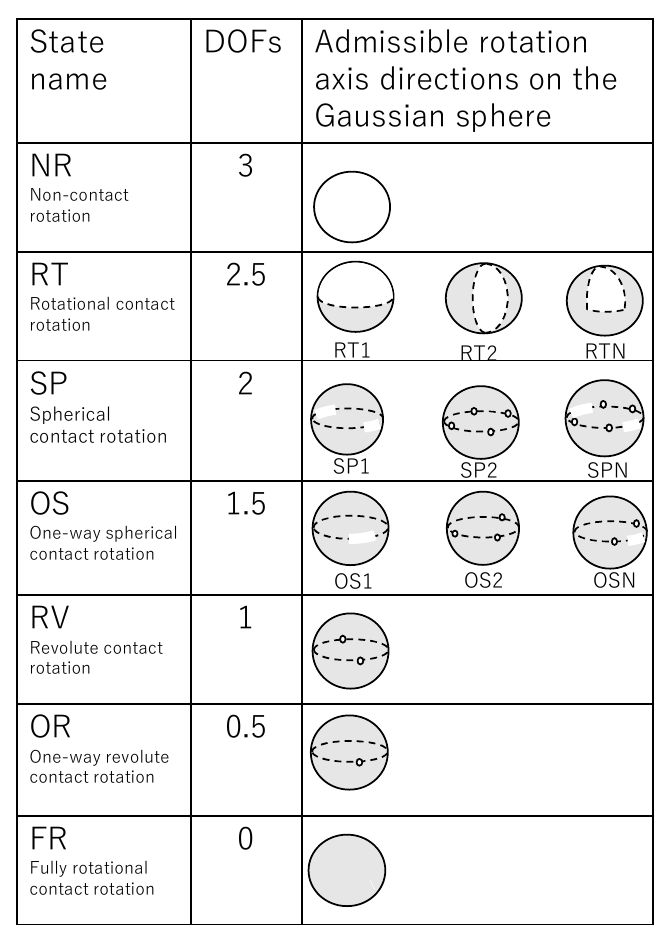}
    \end{minipage}
    \\
    (a) & (b)
    \end{tabular}
    \caption{Translational states and rotational states. (a)~Translational states. For the sake of simplicity, we grouped three partial translational states (\ie, a hemisphere (PC1), a crescent (PC2), and a polygonal shaped state (PCN)) into one PC state, and two one-way prismatic translational states (\ie, a hemi-circle (OT1) and an arc-shaped state (OT2)) into one OT state. (b) Rotational states. See \cite{ikeuchi2021semantic}.}
    \label{fig:contact-states}
\end{figure}

When multiple contact pairs exist, the solution space of these simultaneous inequalities given by them become the possible directions of the object's motion. Using the Kuhn-Tucker theory (\cite{kuhn1957linear}), the solution space of these simultaneous inequalities can be classified into 10 classes. 
For the sake of analysis simplification, these 10 classes of solution spaces are further grouped into 7 types based on the dimension of the DOF, as illustrated in Figure~\ref{fig:contact-states}. Note that PC1, PC2, and PCN were treated as the same PC state, and OT1 and OT2 were also processed as the same OT state. We use this classification as the states of an object.

\subsubsection{Maintenance, detachment, and constraint dimensions}
The relative DOF of an object with respect to the environment can be categorized into three translational and three rotational dimensions. These three dimensions are, further, classified into three types: maintenance dimension, detachment dimension,   and constraint dimension.

\begin{description}
    \item{\bf Maintenance dimension:} The maintenance dimension is a dimension in which small transnational or rotational motion do not face any constraints. For example, when an object is floating in the air, the object can move in any direction within these three dimensions without experiencing any resistance from the environment. Dimensions with such full degrees of freedom are referred to as {\it maintenance dimensions.}
    
    \item{\bf Detachment dimension:} The detachment dimension is a dimension where small translational or rotational motions in that dimension result in the loss of contact. In the opposite direction, any translational or rotational motions are constrained by drag from the environment. For example, a cup on a tabletop can move away from the table, breaking the contact between the surfaces. However, it cannot move towards the table due to the drag. Dimensions with these half degrees of freedom are defined as {\it detachment dimensions.}

    \item{\bf Constraint dimension:} The constraint dimension is a dimension where motions are constrained by resistance from the environment. For example, a drawer is constrained from moving in the direction of its side due to the resistance % it receives 
    from the surrounding walls. Dimensions lacking such degrees of freedom are termed {\it constraint dimensions.}
\end{description}

For states defined by translational motion, maintenance DOF, detachment DOF, and constraint DOF are assigned as shown on the left side of Table~\ref{tab:DOFs}. Since there are three DOF for translation, the total sum of the numbers is 3. Similarly, dimensions can be defined for rotational states as shown on the right side of the table.

\begin{table*}
    \centering
    \caption{DOF distribution}
    \begin{tabular}{|c|c|c|c||c|c|c|c|}
    \hline
       \multicolumn{4}{|c||}{Translation}&\multicolumn{4}{c|}{Rotation} \\ \hline
       State & Maintenance & Detachment & Constraint & State & Maintenance & Detachment & Constraint \\
       \hline
       NC  &  3 & 0 & 0 & NR & 3 & 0 & 0 \\
       PC1  &  2 & 1 & 0 & RT1 & 2 & 1 & 0 \\
       TR  &  2 & 0 & 1 & SP & 2 & 0 & 1 \\
       PC2 &  1 & 2 & 0 & RT2& 1 & 2 & 0 \\
       OT1  &  1 & 1 & 1 & OS1 & 1 & 1 & 1 \\
       PR  &  1 & 0 & 2 & RV & 1 & 0 & 2 \\
       PCN &  0 & 3 & 0 & RTN& 0 & 3 & 0 \\
       OT2 &  0 & 2 & 1 & OS2 & 0 & 2 & 1 \\
       OP &   0 & 1 & 2 & OR & 0 & 1 & 2 \\
       FT &   0 & 0 & 3 & FR & 0 & 0 & 3 \\
       \hline
    \end{tabular}
    \label{tab:DOFs}
\end{table*}

\begin{comment}
この次元分布を利用して、報酬関数を設計してゆく。この場合、状態内遷移のタスクと状態間遷移は、タスクの性質が異なるため、個別に検討する。遷移自体は微小並進や微小回転で発生するが、以下のスキル解析では、ロボットの実際のタスクを前提に、現在の状態がある有限区間継続し、状態遷移が発生し、その後新しい状態がある有限区間継続するものとして、報酬関数等を設計する。また、まず並進について考え、その後同様の考え方を回転にも適用する。  
\end{comment}
\subsubsection{State transitions and skill agents}
\begin{figure}
    \centering
        \includegraphics[width=\linewidth]{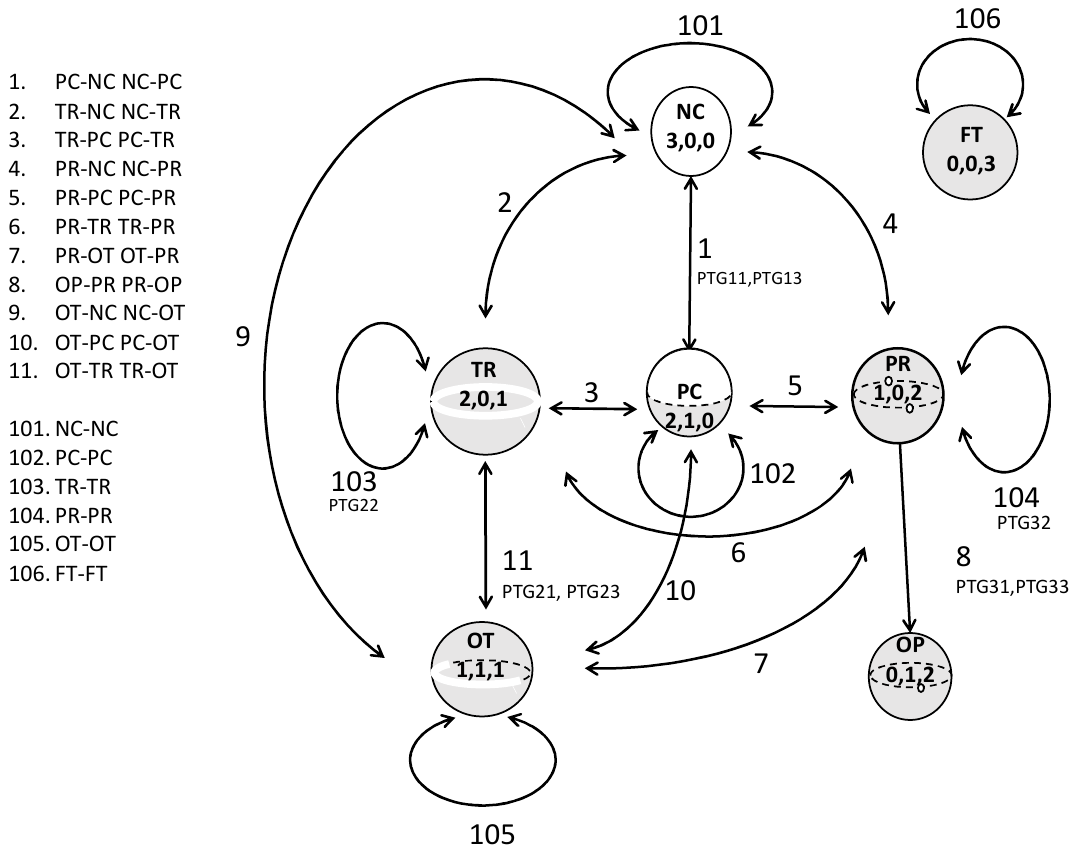}
        \\
        (a) Translational tasks

        \includegraphics[width=\linewidth]{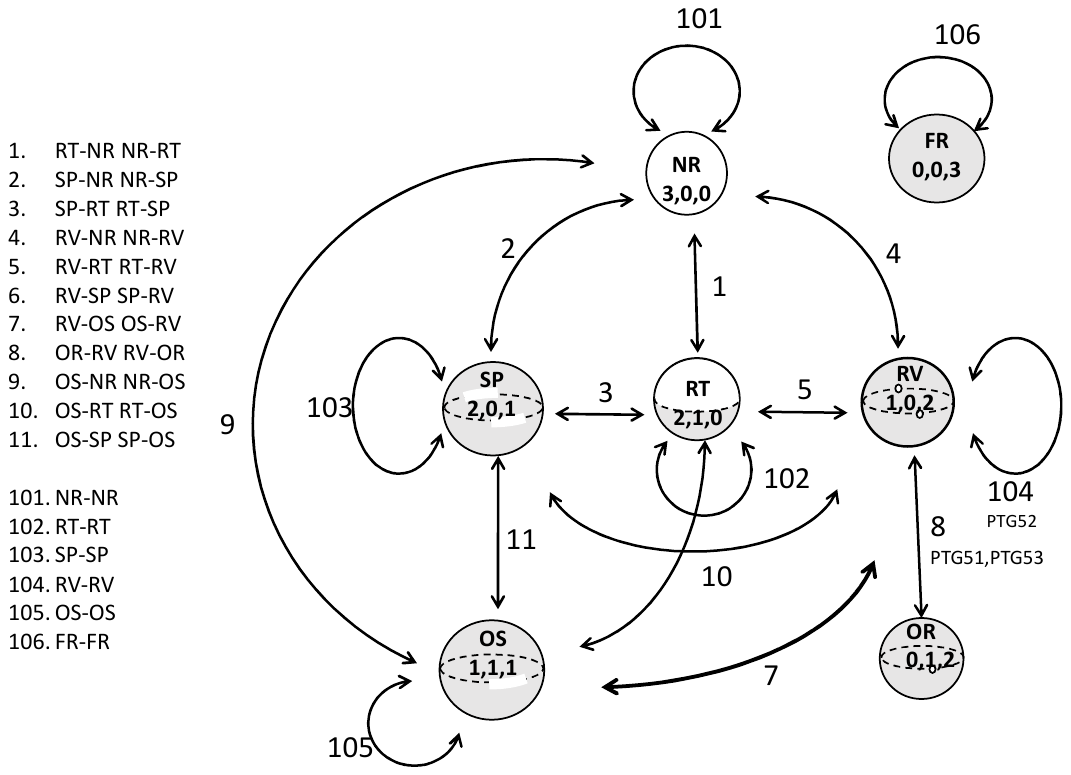}
        \\
       (b) Rotational tasks
    
    \caption{Translational tasks and rotational tasks}

    \label{fig:translation-tasks}
\end{figure}

Tasks are defined as transitions between these states. The transition of states are shown in Figure~\ref{fig:translation-tasks}. In other words, these branches in the graph are defined as tasks, and the purpose of this paper is to design the skill agents to perform these tasks.

\subsection{Design principles for translational skill-agents}
\label{sec:design_principles}

This subsection presents the design principles that are applied to all skill agents, and the next subsection applies these principles to all skill agents to design the reward functions for learning them. Although some of the skill agents may not be practical for robot execution, we design all the reward functions by applying the design principles to all the skill agents regardless of their practicality in order to informally prove the correctness of the design principles as well as to show the upper bound of the skill-agent set.   

We will design the reward functions based on the dimensional transitions. Although transitions themselves occur through infinitesimal translation or rotation, we design skill agents by assuming actual skills performed by a robot, that is, the current state persists for an finite (not infinitesimal) interval before a transition, then, the state transition occurs, and finally a new state persists for another finite interval. 

When designing the reward functions for translational skills, we will provide separate principles for the transition along the direction of motion and those orthogonal to the motion. This is because the transition along the direction of motion mainly affects the termination conditions of the skill agent, while directions orthogonal to the motion are related to motion control strategies during motion. We will first provide principles for the direction of motion and then consider principles for directions orthogonal to the motion.

\subsubsection{Transition along the motion}

\begin{figure}
    \centering
    \includegraphics[width=\linewidth]{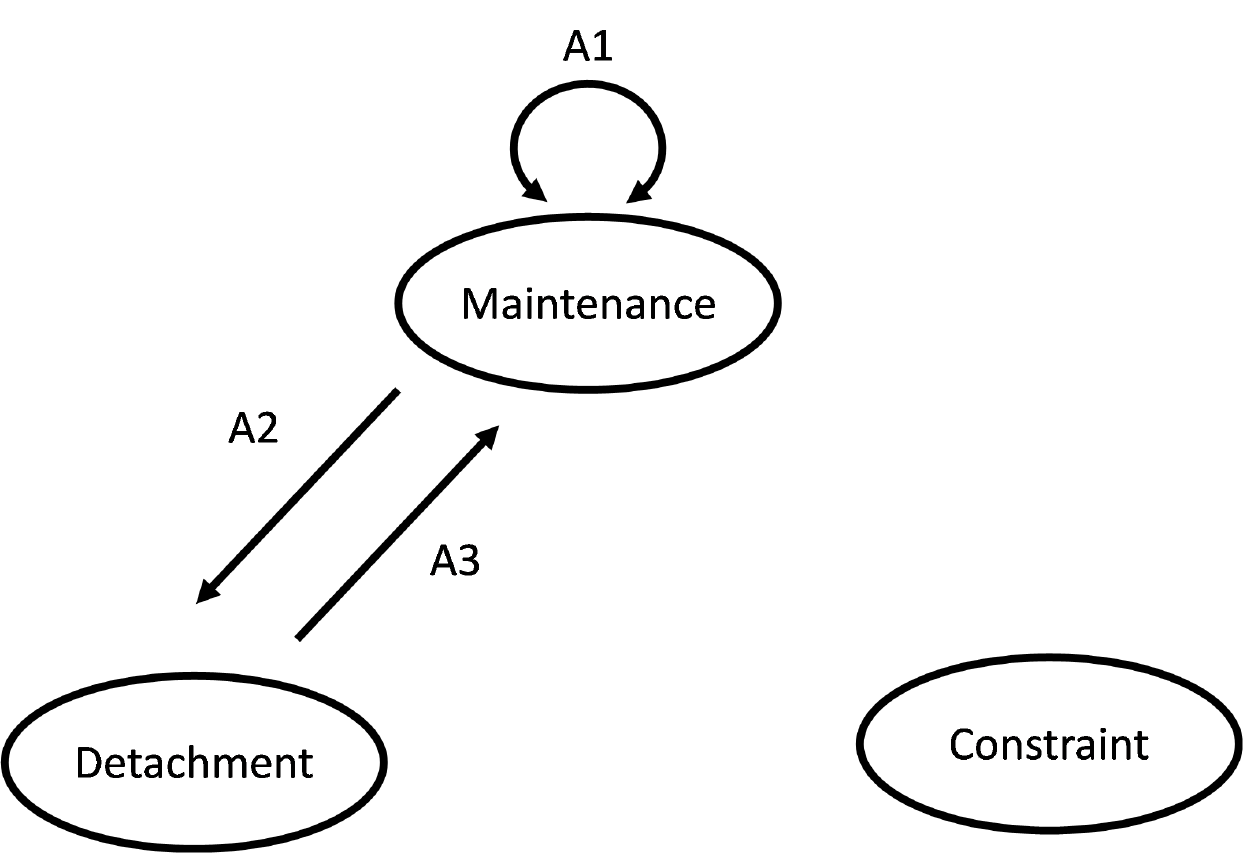}
    \caption{State transitions along motion}
    \label{fig:transalongmotion}
\end{figure}

The limitation on the direction of motion depends on which dimension the direction of motion belongs to. If it is in the maintenance dimension, the object can move in both directions. If it is in the detachment dimension, it can only move in one direction. If it is in the constraint dimension, the object cannot move in that direction. Thus, for state transitions in the motion direction, three cases occur: from the maintenance dimension to the maintenance dimension, from the maintenance dimension to the detachment dimension, and from the detachment dimension to the maintenance dimension: A1, A2 and A3. See Figure~\ref{fig:transalongmotion}.

Note that we denote the coordinate aligned with the direction of motion as S, the drag force of the opposing motion direction as F-s and the drag force along the motion direction as F+s. The two orthogonal directions to the motion direction and the drag forces in these directions are denoted as T, U, F-t, F-u, respectively. Also, let us denote the threshold for determining whether a collision has occurred or not based on the drag force as delta-collision, and the threshold for determining whether the contact has been lost or not as delta-zero. We also note the threshold for determining whether the position of visual features matches or not as delta-gap.

In the following discussion, we define one direction belonging to a certain dimension as a directional state. For example, when one direction is in the maintenance dimension, we call that the direction is the maintenance directional state (d-state). This d-state is defined to distinguish it from the state of the entire object. As long as there is no confusion, dimensions and d-states are used interchangeably.

\paragraph{A1: maintenance to maintenance}

When the maintenance d-state is maintained along the motion direction, no drag force occurs at the end of the motion as well as during the motion. Therefore, the skill agent can be defined solely based on positional information (\ie, reach to the goal position in the S coordinate) given by the demonstration.
\begin{small}
\begin{verbatim}
    if S = goal-s, then reward 
\end{verbatim}
\end{small}
The term goal-s is the goal position in the S coordinate. 

\paragraph{A2: maintenance to detachment}

Drag force that was absent in a maintenance d-state arises upon contact with an environment surface, occurring at the point of transitioning to the detachment d-state. Therefore, the occurrence of this drag force serves as the termination condition for this skill. Under normal circumstances with no errors, the contact position should align with the position given by the demonstration. However, for the sake of operational robustness to allow more gaps between the demonstration and the execution, the occurrence of the drag force from the environment is considered as the termination condition for the skill agent.

\begin{small}
\begin{verbatim}
    if F-s > delta-zero, then reward
\end{verbatim}
\end{small}

\paragraph{A3: detachment to maintenance} 

By moving in a admissible semi-direction in the detachment dimension, the object moves away from the environment contact surface, leading to the transition from the detachment d-state to the maintenance d-state. The disappearance of the drag force that exists in the detachment d-state can be considered as the termination condition for the skill. However, given the existence of motion in the maintenance d-state within a finite interval, the disappearance of the drag force and the the achievement of goal position are considered as the terminal conditions for this skill agent.

\begin{small}
\begin{verbatim}
    if F+s < delta-zero AND S = goal-s,
      then reward
\end{verbatim}
\end{small}

\subsubsection{Transition in the dimension orthogonal to the motion}

\begin{figure}
    \centering
    \includegraphics[width=\linewidth]{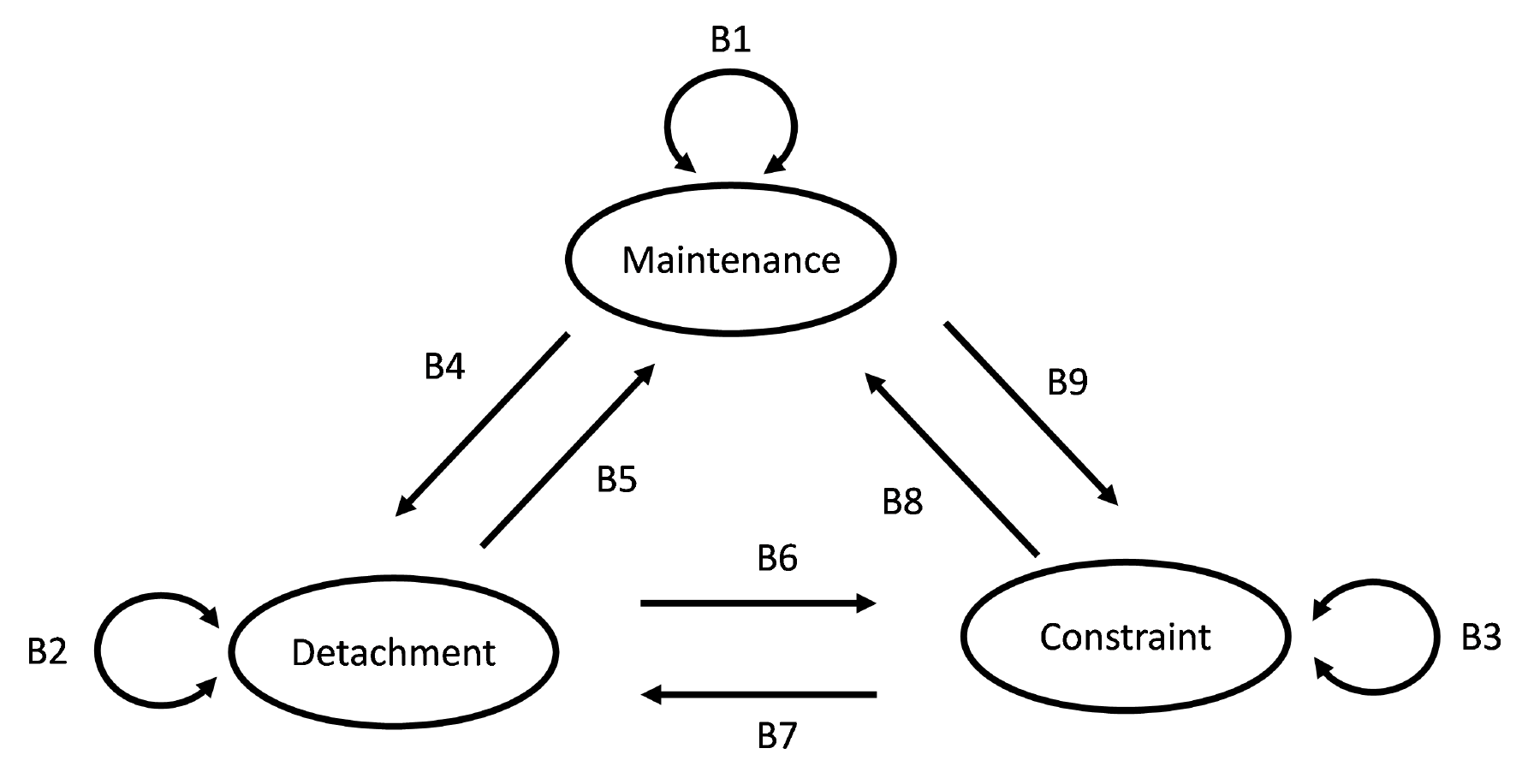}
    \caption{State transitions in the dimension orthogonal to the motion}
    \label{fig:transorthmotion}
\end{figure}

When considering transitions in the dimension orthogonal to the motion direction, nine cases can occur. See Figure~\ref{fig:transorthmotion}. In the following, the dimension considered is denoted as T, and the drag force encountered from the environment along this direction is represented as F-t.

\paragraph{B1: maintenance to maintenance} 
\begin{comment}
運動に直交する方向への拘束がないため、指示された位置情報がもつ誤差は許容でき、位置情報の指定のみでロボットの終端情報を定義できる。
\end{comment}
A moving object to have a maintenance d-state in the orthogonal direction to the motion is not constrained by the environment in that direction. Therefore, the position error in that direction can be tolerated and the end position of the skill in this direction is solely determined by the demonstration.

\begin{small}
\begin{verbatim}
    if T = goal-t, then reward
\end{verbatim}
\end{small}
The term goal-t is the goal position in the T coordinate. 

\paragraph{B2: detachment to detachment} 

Maintaining the detachment d-state in an orthogonal direction to the motion is equivalent to maintaining surface contact in the orthogonal direction during the motion. 
Therefore, throughout the motion, it is necessary to keep the drag force from the environment within a certain range. In other words, the adjustment of the moving direction is required to ensure that the drag force does not become too large, leading to collision, and also to prevent it from reaching zero, resulting in separation from the surface.

\begin{small}
\begin{verbatim}
    if F-t > delta-collision, then penalty
    if F-t < delta-zero, then penalty
\end{verbatim}
\end{small}

\paragraph{B3: constraint to constraint}

To maintain the constraint d-state in the orthogonal direction, similar to B2, it is necessary to adjust the motion direction to minimize the drag force in the orthogonal direction. However, there is no detachment from the constraint d-state in the orthogonal direction, so the second condition corresponding to the delta-zero condition in the case of B2 is not necessary.

\begin{small}
\begin{verbatim}
    if F-t > delta-collision, then penalty
\end{verbatim}
\end{small}

\paragraph{B4: maintenance to detachment} 

For the transition, it is necessary to adjust the motion direction to achieve surface contact in the detachment d-state. Since there is no surface contact in the orthogonal direction in the maintenance d-state, the adjustment needs to be done using positional information from visual sensors. Let us denote the position of the contact surface in the orthogonal direction as \hbox{\em feature-t}. For example, adjusting the position in the T coordinate of the manipulating object to the boundary edge of the surface that will be in contact gives an advantage in accomplishing the task. The skill agent aligns the positional information in the orthogonal direction using this value. 

After the transition, the motion direction is adjusted to maintain the detachment d-state in the same as in the B2 case. In order to specify whether the transition occurs or not, we introduce a flag referred to {\em AfterTransition}.

\begin{small}
\begin{verbatim}
  if NOT(AfterTransition):
    if |T - feature-t| > delta-gap, 
      then penalty
  else:
    if F-t > delta-collision, then penalty
    if F-t < delta-zero, then penalty
\end{verbatim}
\end{small}
\paragraph{B5: detachment to maintenance} 

In the finite interval before the transition, the motion direction is adjusted to maintain the surface contact in the orthogonal direction, \ie, detachment d-state in this direction. For this, the drag force F-t should be no greater than delta-collision so that the object dips into the environmental surface. In strict sense, the detachment d-state should be kept until the surface contact is disappeared to the motion (\ie, detach it from the boundary as shown in Figure~\ref{fig:two-motion-directions}~(b)). Then an additional condition imposing non-zero drag to prevent the separation is applied. % Note that there is a possibility not to apply the additional condition if the task is just achieved.
%However, an additional condition imposing non-zero drag to prevent the separation is not applied because the early departure in the orthogonal to the motion is acceptable. 

After the transition, in the maintenance d-state, there are no constraints from the environment in this orthogonal direction, allowing the termination condition based on positional information given by the demonstration. % However, to prevent errors due to positional inaccuracies, a condition to confirm that this direction is in the maintenance d-state is added.
\if 0
\begin{verbatim}
    if F-t > delta-collision, then penalty 
    if F-t < delta-zero AND T = goal-t,
      then reward
\end{verbatim}
\fi
\begin{small}
\begin{verbatim}
    if NOT(AfterTransition):
      if F-t > delta-collision, then penalty
      if F-t < delta-zero, then penalty
    else:
      if F-t < delta-zero AND T = goal-t,
        then reward
\end{verbatim}
\end{small}

\paragraph{B6: detachment to constraint}

Generally, the transition from the detachment d-state to the constraint d-state incurs costs. Fortunately, by implementing control to retain the detachment d-state, it is possible to achieve the constraint d-state. Before the transition, retaining the detachment d-state is realized by maintaining the contact with one of the surface. And the contact with the other is automatically achieved after the transition because of the geometries of the object and environment, and the original contact is still maintained. Therefore, the skill agent maintains the detachment d-state throughout the entire interval and to keep the drag within a certain range.

\begin{small}
\begin{verbatim}
    if F-t > delta-collision, then penalty
    if F-t < delta-zero, then penalty
\end{verbatim}
\end{small}

\paragraph{B7: constraint to detachment}

Similar to the case of B6, by implementing control to retain the detachment d-state, it is possible to maintain the constraint d-state. Therefore, the approach is adopted to maintain the detachment d-state throughout the entire interval and to keep the drag force from the environment constant in the orthogonal direction.
\begin{small}
\begin{verbatim}
    if F-t > delta-collision, then penalty
    if F-t < delta-zero, then penalty
\end{verbatim}
\end{small}

\paragraph{B8: constraint to maintenance}

The constraint d-state before the transition requires to ensure that the drag force in the orthogonal direction does not exceed a threshold, delta-collision. After the transition, since the direction becomes the maintenance d-state, there is no need to check this condition. %However, regardless, this condition is satisfied anyway during the maintenance d-state, there is no need to void this condition after the transition by checking the transition.
In the finite interval after the transition, the maintenance d-state allowing the use of positional information given by the demonstration. However, the condition to confirm the attainment of the maintenance d-state should be included.
\begin{small}
\begin{verbatim}
    if F-t > delta-collision, then penalty
    if F-t < delta-zero AND T = goal-t, 
      then reward
\end{verbatim}
\end{small}

\paragraph{B9: maintenance to constraint}

Immediately before the transition, it is necessary to obtain positional information, labeled as feature-t, from the visual data to initiate contact for the constraint d-state. % (Generally, measuring edges orthogonal to the orthogonal direction with a visual sensor is advantageous.) 
After the transition, to adjust the motion direction is required to maintain the constraint d-state. %To clearly define this transition, a flag called AfterTransition is introduced.

\begin{small}
\begin{verbatim}
  if NOT(AfterTranstion):
    if |T - feature-t| > delta-gap, 
      then penalty
  else:
    if F-t > delta-collision, then penalty
\end{verbatim}
\end{small}

\subsection{Interstate transition}

In this subsection, we apply the design principles obtained in the previous section to the interstate transitions of an object and derive the reward functions. Penalty conditions in the design principles are applied with OR logic, since the task has failed if even one penalty condition is satisfied. On the other hand, the reward conditions are applied with AND logic, since the task has reached to the goal when all the conditions are satisfied. In the design of each state transition below, we first consider the transitions from states with more constraints to states with fewer constraints and then complete the reverse transitions.

\subsubsection{PC-NC and NC-PC}

\begin{figure}
    \centering
    \begin{tabular}{cc}
    \begin{minipage}{0.43\hsize}
    \includegraphics[width=\linewidth]{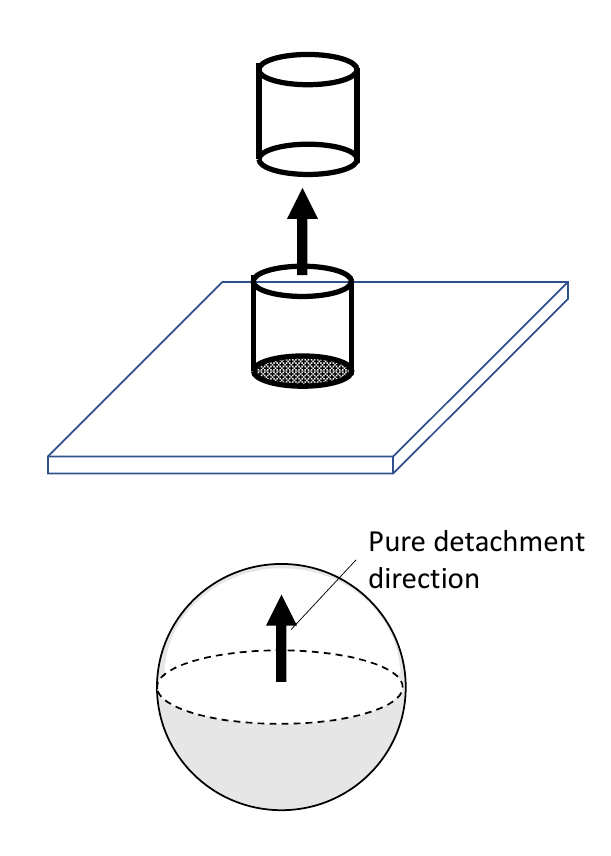}
    \end{minipage} &
    \begin{minipage}{0.43\hsize}
    \includegraphics[width=\linewidth]{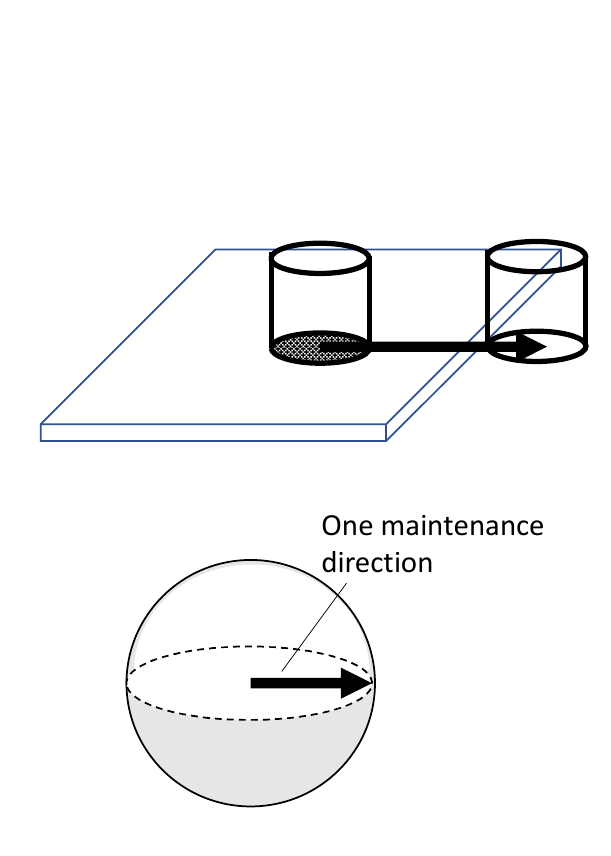}
    \end{minipage}
    \\
    (a)
    &
    (b)
    
    \end{tabular}
    \caption{Two types of motion directions. (a) motion in the detachment direction. (b) motion in the maintenance direction. }
    \label{fig:two-motion-directions}
\end{figure}

An object in PC (M=2, D=1, C=0) has two maintenance d-states and one detachment d-state. For example, it could be a cube on a desk. The normal direction of the desk surface serves as the pure detachment direction (detachment d-state) and any direction along the desk surface is a maintenance direction (maintenance d-state). In PC, we can consider two types of tasks to move the object: in the detachment direction and in the maintenance direction as shown in Figure~\ref{fig:two-motion-directions}.

{\it PC-NC detachment motion:} the pure detachment direction of the object is along the normal direction of the contact surface. The motion of the object along the direction results in no contact with surface, causing the transition to the maintenance d-state from the detachment d-state and the state of the object becomes NC (M=3, D=0, C=0) from PC. This is a typical example of the Pick-up task.

{\it PC-NC maintenance motion:} When moving in the maintenance direction, the infinitesimal motion does not cause any transition of dimensions. The transition to NC and loss of contact only occurs due to the shape of the environment surface. Namely, the transition occurs by moving of the object along the contact surface with maintaining contact with the surface and then reaching the edge of the surface due to the overall shape of the surface. After this point, there is no contact with the surface, and the direction orthogonal to the motion transits from the detachment d-state to the maintenance d-state.

In typical robot operations, the first scenario is much more common, and the second one is rare and not very practical. However, for the sake of comprehensive descriptions, the second scenario is also included.

\paragraph{PC-NC-a: detachment motion}

This scenario corresponds to a typical {\it Pick-up} task. Concerning the motion direction~S, a transition occurs from the detachment d-state to the maintenance d-state, and A3 can be applied:
\begin{small}
\begin{verbatim}
  A3: if F+s < delta-zero AND S = goal-s,
        then reward 
\end{verbatim}
\end{small}

In the two orthogonal directions to the motion, T and U, the maintenance d-state is preserved. Therefore, B1 can be applied for the two directions:
\begin{small}
\begin{verbatim}
  B1: if T = goal-t, then reward
  B1: if U = goal-u, then reward
\end{verbatim}
\end{small}

Combining these reward conditions with AND logic yields the following reward function. In other words, the goal is to eliminate drag force in the motion direction by an infinitesimal motion and reach a specified position, given from the demonstration, by a finite motion.
\begin{small}
\begin{verbatim}
  Reward PC-NC-a (PTG11 (Pick) task)
    if F+s < delta-zero AND S = goal-s AND 
       T = goal-t AND U = goal-u,
      then reward 
\end{verbatim}
\end{small}
Note that this corresponds to PTG11 in \cite{IkeuchiIJCV2018}.

\paragraph {PC-NC-b: maintenance motion}

This case occurs when surface contact disappears at the edge of the environment surface (\ie, table surface) as an example scenario shown in Figure~\ref{fig:two-motion-directions}~(b). 

Regarding the motion direction, the maintenance d-state remains after the transition, and A1 can be applied:
\begin{small}
\begin{verbatim}
  A1: if S = goal-s, then reward  
\end{verbatim}
\end{small}

On one hand, one of the two directions orthogonal to the motion, in the example shown in the figure, the direction parallel to the table surface (here we call the T direction) does not undergo a transition in the maintenance d-state before and after the transition. Therefore, B1 can be applied:
\begin{small}
\begin{verbatim}
  B1: if T = goal-t, then reward
\end{verbatim}
\end{small}

On the other hand, the other dimension, the vertical direction in the example (here we call the U direction) undergoes a transition from the detachment d-state to the maintenance d-state. B5 can be applied. In the finite interval before the transition, the detachment d-state is preserved, and in the finite interval after the transition, position control becomes relevant. However, it is necessary to include the condition to confirm that departure from the surface has occurred.
\begin{small}
\begin{verbatim}
  B5: if NOT(AfterTransition):
        if F-u > delta-collision, 
          then penalty
        if F-u < delta-zero, then penalty
      else:
        if F-u < delta-zero AND U = goal-u,
          then reward
\end{verbatim}
\end{small}
\if 0
\begin{small}
\begin{verbatim}
  B5: if F-u > delta-collision, then penalty 
      if F-u < delta-zero AND U = goal-u, 
        then reward
\end{verbatim}
\end{small}
\fi

Combining these, the following reward function is obtained.
\begin{small}
\begin{verbatim}
  Reward PC-NC-b
    if NOT(AfterTransition):
      if F-u > delta-collision, then penalty
      if F-u < delta-zero, then penalty
    else:
      if S = goal-s AND T = goal-t AND 
         F-u < delta-zero AND U = goal-u,
        then reward  
\end{verbatim}
\end{small}

\paragraph{NC-PC-a: attachment motion}

In the same way as PC-NC, NC-PC also has two scenarios. Here, please imagine cases where the directions of the arrows are reversed in Figure~\ref{fig:two-motion-directions}. One corresponds to the common {\it Place} task where the detachment d-state is achieved by moving from the direction that will become the detachment direction after the transition, \ie, placing an object from above the contact surface. The other is a race case where the detachment d-state is achieved by approaching the contact surface from the side, causing contact.

When placing an object from above the contact surface, along the motion direction, the maintenance d-state transits to the detachment d-state. A2 can be applied:
\begin{small}
\begin{verbatim}
  A2: if F-s > delta-zero, then reward
\end{verbatim}
\end{small}
In the two orthogonal directions to the motion, the maintenance d-state remains unchanged. B1 can be applied:
\begin{small}
\begin{verbatim}
  B1: if T = goal-t, then reward
  B1: if U = goal-u, then reward
\end{verbatim}
\end{small}
Therefore, the reward function is as follows:
\begin{small}
\begin{verbatim}
  Reward NC-PC-a (PTG13 (Place) task)
    if F-s > delta-zero AND T = goal-t AND 
       U = goal-u, then reward
\end{verbatim}
\end{small}
Note that this is named PTG13 in~\cite{IkeuchiIJCV2018}.

\paragraph{NC-PC-b: maintenance motion}

In the example shown in the figure, the object moves parallel to the desk surface from the outside of the desk to precisely induces the surface contact on the desk. This task is not advantageous for robot operations, so its frequency of use may not be high. However, it is included here to ensure overall completeness and necessity. 

Regarding the motion direction, it remains in the maintenance d-state, and A1 can be applied:
\begin{small}
\begin{verbatim}
  A1: if S = goal-s, then reward
\end{verbatim}
\end{small}
One dimension orthogonal to the motion also remains in the maintenance d-state, and B1 can be applied:
\begin{small}
\begin{verbatim}
  B1: if T = goal-t, then reward
\end{verbatim}
\end{small}

On the other hand, for the remaining orthogonal dimension, a transition from a maintenance d-state to a detachment d-state occurs; B4 can be applied. Before the transition, as there is no physical contact, visual feedback becomes necessary to adjust the position in this direction. After the transition in a finite interval, it is necessary to maintain the detachment d-state.
\begin{small}
\begin{verbatim}
  B4: if NOT(AfterTransition):
        if |U - feature-u| > delta-gap, 
          then penalty
      else:
        if F-u > delta-collision,
          then penalty
        if F-u < delta-zero, then penalty
\end{verbatim}
\end{small}

\begin{figure*}[ht]
    \centering
    \includegraphics[width=\linewidth]{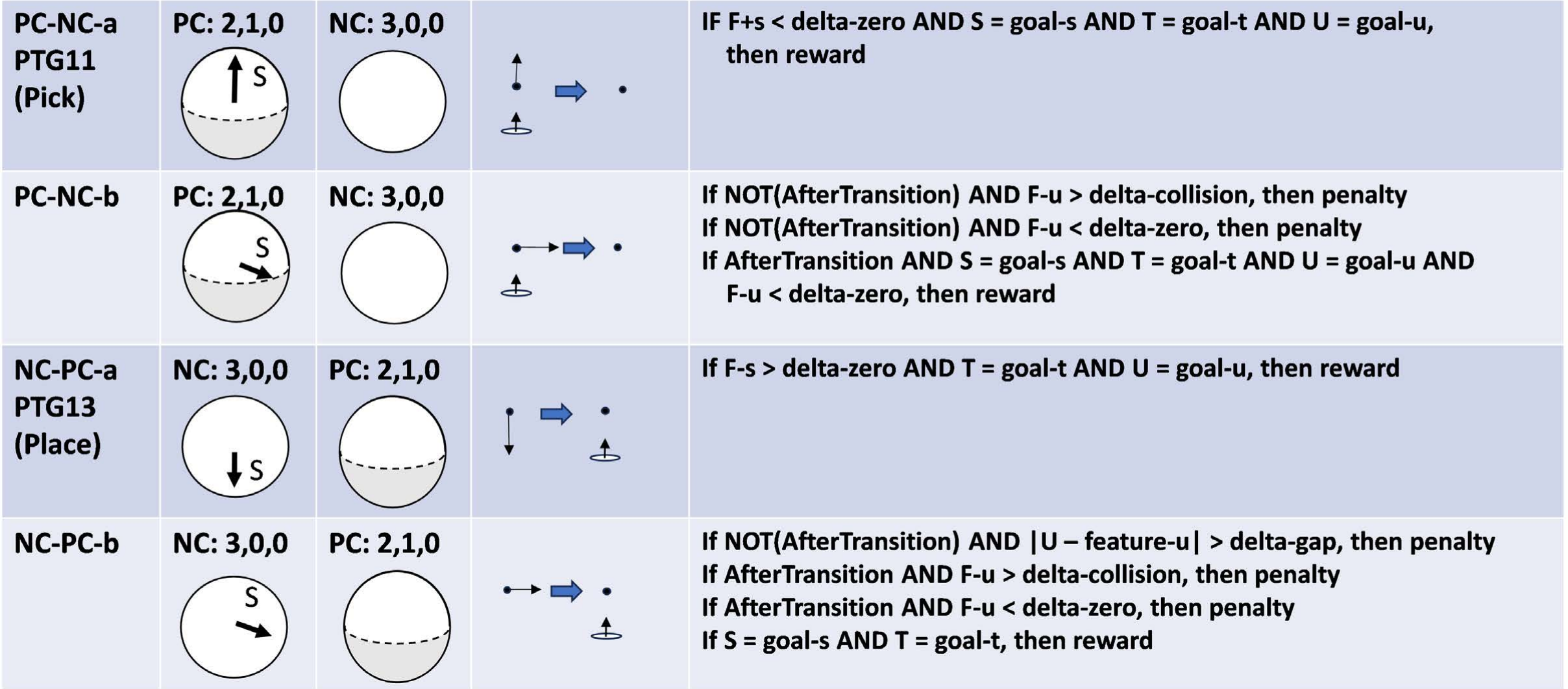}
    \caption{PC-NC and NC-PC skills}
    \label{fig:pc-nc}
\end{figure*}

Therefore, the reward function is as follows:
\begin{small}
\begin{verbatim}
  Reward NC-PC-b
    if NOT(AfterTransition):
      if |U - feature-u| > delta-gap, 
        then penalty
    else:
      if F-u > delta-collision, then penalty
      if F-u < delta-zero, then penalty
      if S = goal-s AND T = goal-t, 
        then reward
\end{verbatim}
\end{small}
Figure~\ref{fig:pc-nc} summarizes the skills related to PC-NC and NC-PC.

\subsubsection{TR-NC and NC-TR}
\paragraph{TR-NC}

TR (M=2, D=0, C=1) has two maintenance dimensions and one constraint dimension and a typical example of an object in this state is a cube sandwiched between two parallel walls. In this case, the normal direction of the wall is in the constraint dimension. The cube can move in the maintenance directions (\ie, in the directions parallel to the walls). This maintenance motion itself does not cause a state transition of the object in an infinitesimal interval, but the transition occurs due to the shape of the constraint surfaces when it moves for a certain finite interval. 

\begin{figure}
    \centering
    \begin{tabular}{cc}
    \begin{minipage}{0.45\hsize}
    \includegraphics[width=\linewidth]{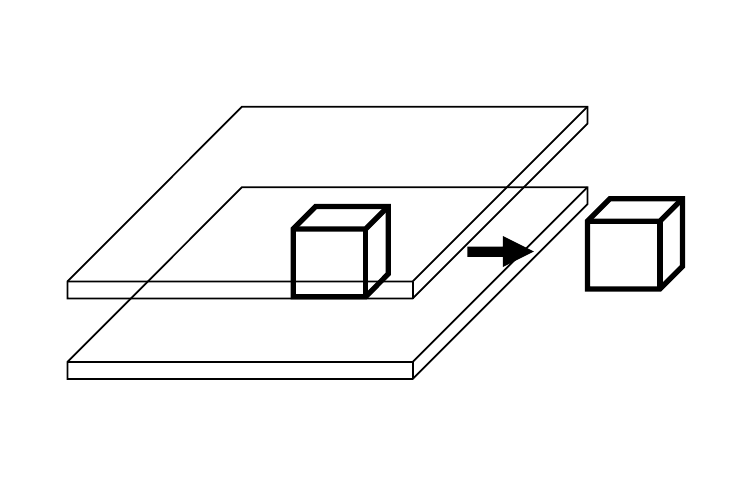}
    \end{minipage} &
    \begin{minipage}{0.45\hsize}
    \includegraphics[width=\linewidth]{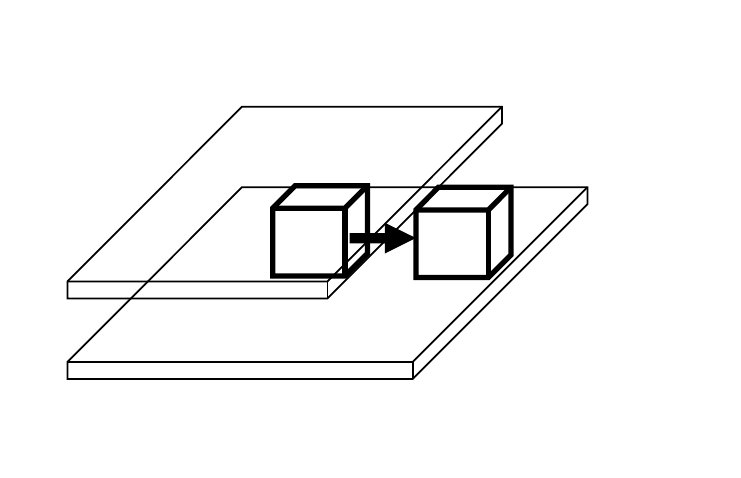}
    \end{minipage}
    \\
    (a)
    &
    (b)
    
    \end{tabular}
    \caption{Two scenarios. (a) TR-NC. (b) TR-PC.}
    \label{fig:cube-sand}
\end{figure}

Depending on the shape of the constraint surface, it could transit to NC (M=3, D=0, C=0) or PC (M=2, D=0, C=1) as shown in Figure~\ref{fig:cube-sand}. The TR-NC transition occurs when the edges of the two constraint surfaces are at the same position, and the moving cube simultaneously loses contact with these faces.

Regarding the motion direction, it maintains the maintenance d-state before and after the transition, satisfying A1:
\begin{small}
\begin{verbatim}
  A1: if S = goal-s, then reward   
\end{verbatim}
\end{small}
One of the dimensions orthogonal to the motion direction maintains the maintenance d-state % (\ie, no contact) 
before and after the transition, satisfying B1:
\begin{small}
\begin{verbatim}
  B1: if T = goal-t, then reward   
\end{verbatim}
\end{small}

The remaining orthogonal dimension experiences a sudden disappearance of the constraint surfaces, transiting from the constraint d-state to the maintenance d-state. B8 is applicable:
\begin{small}
\begin{verbatim}
  B8: if F-u > delta-collision, then penalty
      if F-u < delta-zero AND U = goal-u,
        then reward  
\end{verbatim}
\end{small}

In summary,
\begin{small}
\begin{verbatim}
  Reward TR-NC
    if F-u > delta-collisoin, then penalty
    if S = goal-s AND T = goal-t AND 
       F-u < delta-zero AND U = goal-u, 
      then reward
\end{verbatim}
\end{small}

\paragraph{NC-TR}

This NC (M=3, D=0, C=0) - TR (M=2, D=0, C=1) is the reverse skill of the earlier TR-NC, and for example, a skill such as placing a book in the air into the space between two books would cause the transition from one maintenance dimension to one constraint dimension. Along this dimension, this skill requires the use of vision, as there is no surface contact before the transition.

\begin{figure*}[ht]
    \centering
    \includegraphics[width=\linewidth]{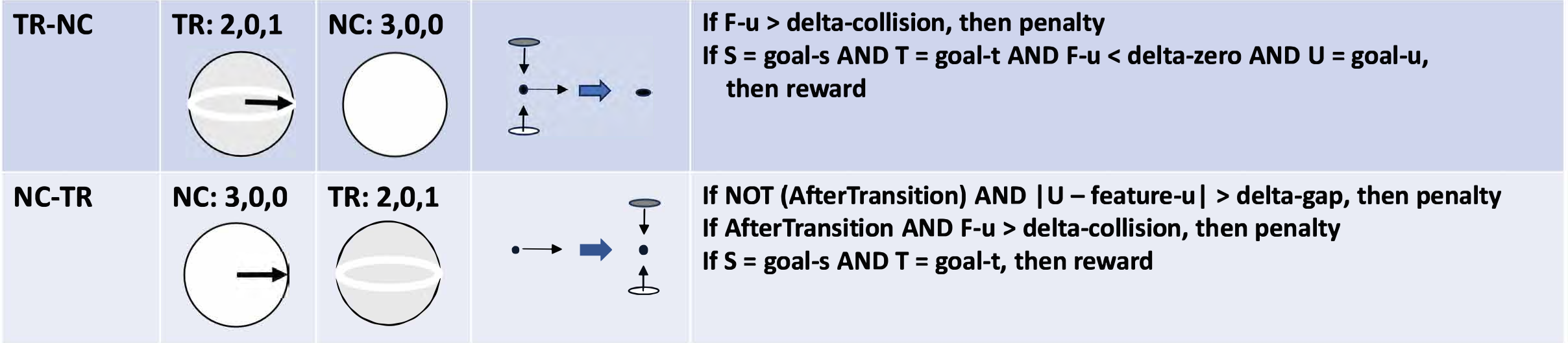}
    \caption{TR-NC and NC-TR skills}
    \label{fig:tr-nc}
\end{figure*}

As for the direction of motion, A1 is applicable since it remains a maintenance d-state.
\begin{small}
\begin{verbatim}
  A1: if S = goal-s, then reward 
\end{verbatim}
\end{small}
One direction orthogonal to the motion direction remains in a maintenance d-state before and after the transition, and B1 is applied:
\begin{small}
\begin{verbatim}
  B1: if T = goal-t, then reward 
\end{verbatim}
\end{small}
For the dimension for which the maintenance d-state changes to the constraint d-state (the normal direction of the book in the above example), B9 can be applied and visual information must be used. %Edge information orthogonal to this dimension must be extracted using a visual sensor.
\begin{small}
\begin{verbatim}
  B9: if NOT(AfterTransition):
        if |U - feature-u| > delta-gap, 
          then penalty
      else:
        if F-U > delta-collision, 
          then penalty  
\end{verbatim}
\end{small}
In summary,
\begin{small}
\begin{verbatim}
  Reward NC-TR
    if NOT(AfterTransition):
      if |U - feature-u| > delta-gap, 
        then penalty
    else: 
      if F-u > delta-collision, then penalty
      if S = goal-S AND T = goal-t, 
        then reward
\end{verbatim}
\end{small}
Figure~\ref{fig:tr-nc} summarises the skills of TR-NC and NC-TR.

\subsubsection{TR-PC and PC-TR}

\paragraph{TR-PC}

As shown in Figure~\ref{fig:cube-sand}~(b), the TR-PC transition occurs in the case that the edge positions of the top and bottom surfaces are different and the top surface loses contact while the bottom surface is still in contact in the finite motion.

Along the motion direction, since the maintenance d-state is preserved before and after the transition, A1 can be applied and the terminal point of the skill can be specified as the positional information given from the demonstration.
\begin{small}
\begin{verbatim}
  A1: if S = goal-s, then reward  
\end{verbatim}
\end{small}
One of the two dimensions orthogonal to the motion remains in the maintenance d-state before and after the transition. Therefore, with B1, positional information can also be specified for this dimension.
\begin{small}
\begin{verbatim}
  B1: if T = goal-t, then reward
\end{verbatim}
\end{small}

In the remaining dimension, the constraint d-state transits to the detachment d-state. In the above example, before the transition, the cube is constrained to the top and the bottom surfaces, and after the transition, it detaches from the top surface by sliding the bottom surface. In this direction, B7 can be applied. % Before the transition, the motion is adjusted to maintain the constraint d-state. After the transition the motion is adjusted to keep the detachment d-state. %However, since detecting this transition is usually difficult, in B7, 
%The control for the detachment d-state is performed throughout the transition. In other words, 
\begin{small}
\begin{verbatim}
  B7: if F-u > delta-collision, then penalty
      if F-u < delta-zero, then penalty
\end{verbatim}
\end{small}

In summary,
\begin{small}
\begin{verbatim}
  Reward TR-PC
    if F-u > delta-collision, then penalty
    if F-u < delta-zero, then penalty
    if S = goal-s AND T = goal-t, 
      then reward  
\end{verbatim}
\end{small}

\paragraph{PC-TR}

Let us consider a cube sliding on a table into a gap between the table surface and a parallel upper wall. A state transition occurs from PC (M=2, D=1, C=0) on the table to TR (M=2, D=0, C=1) in the gap. 
As described above, this sliding motion automatically causes such transition and if this cube collides the top wall due to the difference in size, the skill is fundamentally infeasible due to the difference between the sizes of the object (the cube) and the environment (the walls).

The motion direction remains in the maintenance d-state. Therefore, A1 is applicable: 
\begin{small}
\begin{verbatim} 
  A1: if S = goal-s, then reward
\end{verbatim}
\end{small}
One dimension orthogonal to the motion direction remains in the maintenance d-state, \ie, the horizontal direction in the above example, allowing for the application of B1:
\begin{small}
\begin{verbatim}
  B1: if T = goal-t, then reward
\end{verbatim}
\end{small}
On the other hand, for the dimension transiting from the detachment d-state to the constraint d-state, \ie, the normal direction of the table surface in the above example, B6 can be applied. Essentially, by controlling to maintain the detachment d-state, \ie, keeping the surface contact, the detachment d-state naturally transits into the constraint d-state.
\begin{small}
\begin{verbatim}
  B6: if F-u > delta-collision, then penalty
      if F-u < delta-zero, then penaly
\end{verbatim}
\end{small}

\begin{figure*}[ht]
    \centering
    \includegraphics[width=\linewidth]{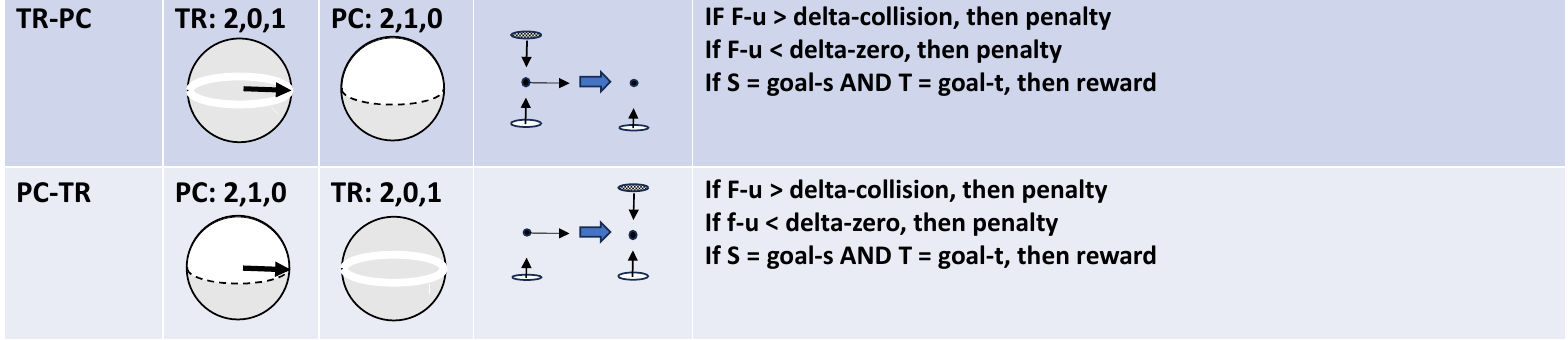}
    \caption{TR-PC and PC-TR skills}
    \label{fig:pc-tr}
\end{figure*}

In summary, 
\begin{small}
\begin{verbatim}
  Reward PC-TR
    if F-u > delta-collision, then penalty
    if F-u < delta-zero, then penalty
    if S = goal-s AND T = goal-t,
      then reward    
\end{verbatim}
\end{small}
Figure~\ref{fig:pc-tr} summarises the skills of TR-PC and PC-TR.

\subsection{Other interstate transitions}

The similar discussions can be applied for other interstate transitions. For these transitions, only brief descriptions and results are listed below. Detailed derivations of the reward functions are given in Appendix A and Figures~\ref{fig:pr-nc} to~\ref{fig:ot-tr} show their reward functions.

\subsubsection{PR-NC and NC-PR}
PR (M=1, D=0, C=2) - NC (M=3, D=0, C=0) involves such as completely pulling a peg out of the hole, while NC-PR involves the opposite skill of swiftly inserting a peg into a hole. Since there is no physical contact with the environment before the transition, it is necessary to determine the hole position using visual sensors. % The reward functions are shown in Figure~\ref{fig:pr-nc}. % Regarding the derivation of the reward functions, please refer to the Appendix.

\begin{figure*}[ht]
    \centering
    \includegraphics[width=\linewidth]{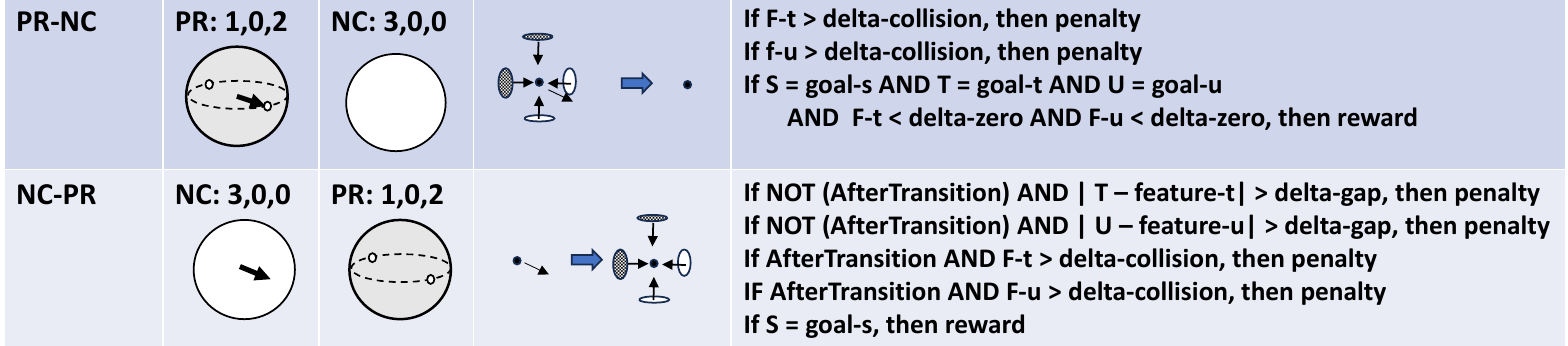}
    \caption{PR-NC and NC-PR skills}
    \label{fig:pr-nc}
\end{figure*}

\subsubsection{PR-PC and PC-PR}
The PR (M=1, D=0, C=2) - PC (M=2, D=1, C=0) transition occurs, for example, when pulling a peg inside a hole, of which a part of the side is extended externally. Thus, when the peg comes out, contact with the extended surface persists, leading to a detachment d-state instead of a maintenance d-state in that direction unlike the PR-NC skill. %The reward functions are shown in Figure~\ref{fig:pr-pc}. %(i.e., C to D transition) in that direction.

\begin{figure*}[ht]
    \centering
    \includegraphics[width=\linewidth]{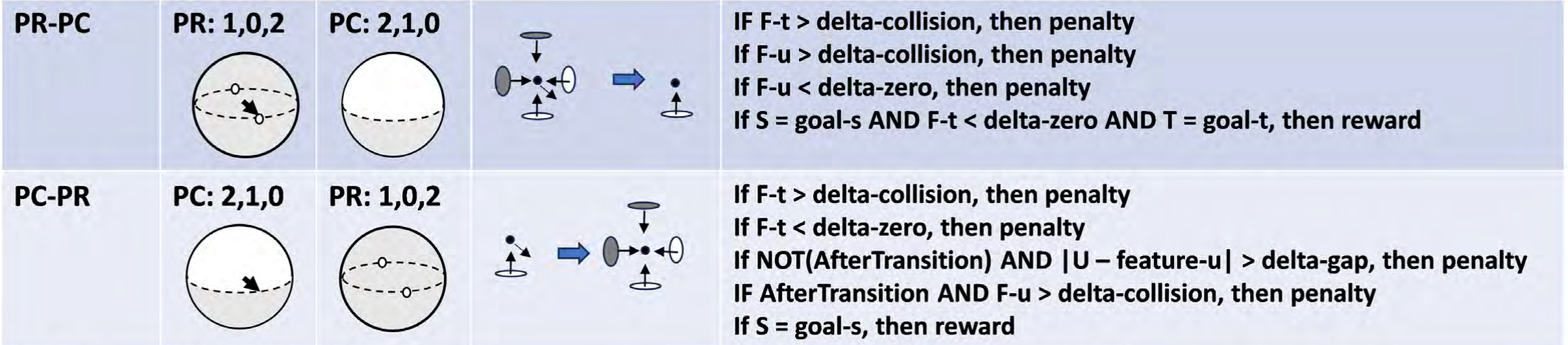}
    \caption{PR-PC and PC-PR skills}
    \label{fig:pr-pc}
\end{figure*}

\subsubsection{PR-TR and TR-PR}
PR (M=1, D=0, C=2) - TR (M=2, D=0, C=1) is a skill required in situations where, using the previous example, a pair of opposite surfaces extend outside the hole, and despite the transition from the constraint d-state to the maintenance d-state occurs in the other orthogonal direction, this direction remains in the constraint d-state. % The reward functions are shown in Figure~\ref{fig:pr-tr}.

\begin{figure*}[ht]
    \centering
    \includegraphics[width=\linewidth]{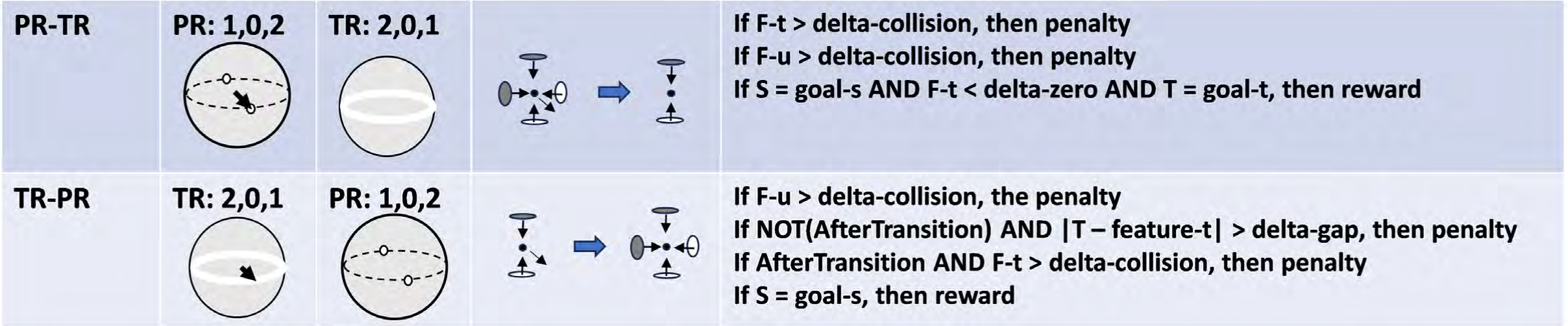}
    \caption{PR-TR and TR-PR skills}
    \label{fig:pr-tr}
\end{figure*}

\subsubsection{PR-OT and OT-PR}
PR (M=1, D=0, C=2) - OT (M=1, D=1, C=1) is a skill required in situations where surface contact continues in three directions. % The reward functions are shown in Figure~\ref{fig:pr-ot}.

\begin{figure*}[ht]
    \centering
    \includegraphics[width=\linewidth]{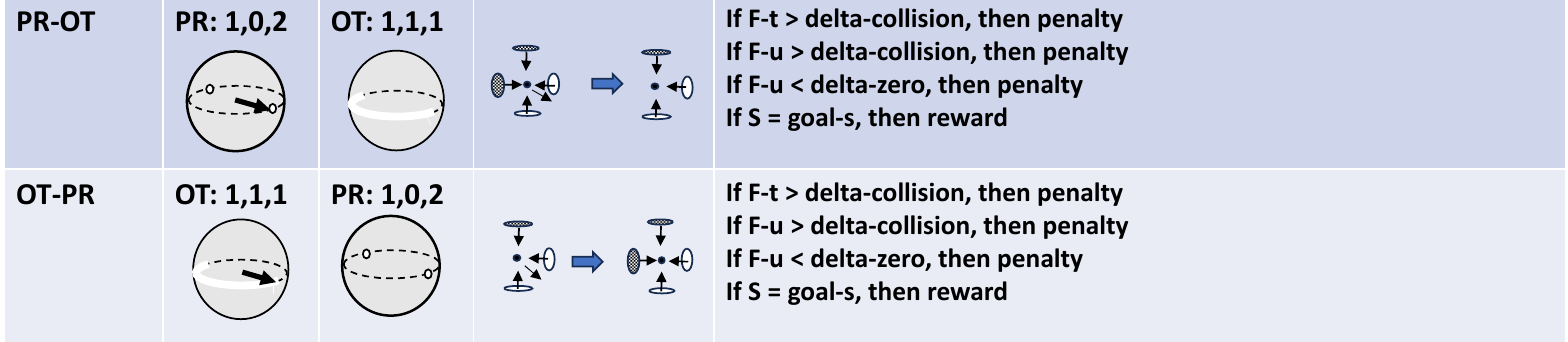}
    \caption{PR-OT and OT-PR skills}
    \label{fig:pr-ot}
\end{figure*}

\subsubsection{OP-PR and PR-OP}
OP (M=0, D=1, C=2) is like a peg that has reached the bottom of a hole and can move only in the half direction away from the bottom, with two orthogonal directions to this direction constrained, while PR (M=1, D=0, C=2) is like a peg that remains in the middle of a hole and can move along the hole in both directions. In the previous paper, we name OP-PR and PR-OP as PTG31 (Drawer-opening) and PTG33 (Drawer-closing). % We can obtain those reward functions as shown in Figure~\ref{fig:op-pr}. % Regarding the derivation of the reward functions, please refer to the Appendix.

\begin{figure*}[ht]
    \centering
    \includegraphics[width=\linewidth]{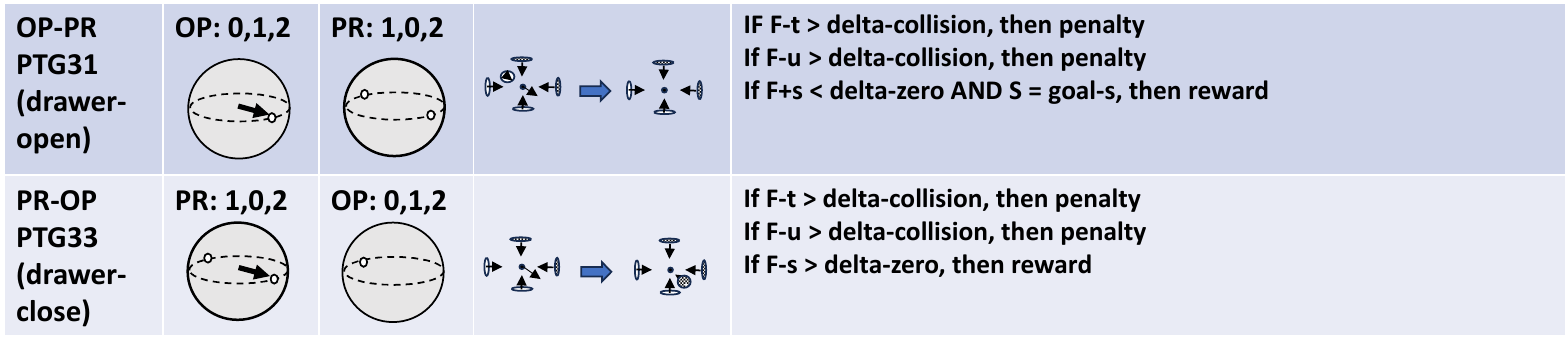}
    \caption{OP-PR and PR-OP skills}
    \label{fig:op-pr}
\end{figure*}

\subsubsection{OT-NC and NC-OT}
OT (M=1, D=1, C=1) - NC (M=3, D=0, C=0) is a skill that involves pulling an object along the detachment surface, not in the normal direction of the detachment surface, when the object is surrounded by a pair of opposing directions and another direction. In the reverse skill, NC-OT, visual feedback is required. % The reward functions are shown in Figure~\ref{fig:ot-nc}

\begin{figure*}[ht]
    \centering
    \includegraphics[width=\linewidth]{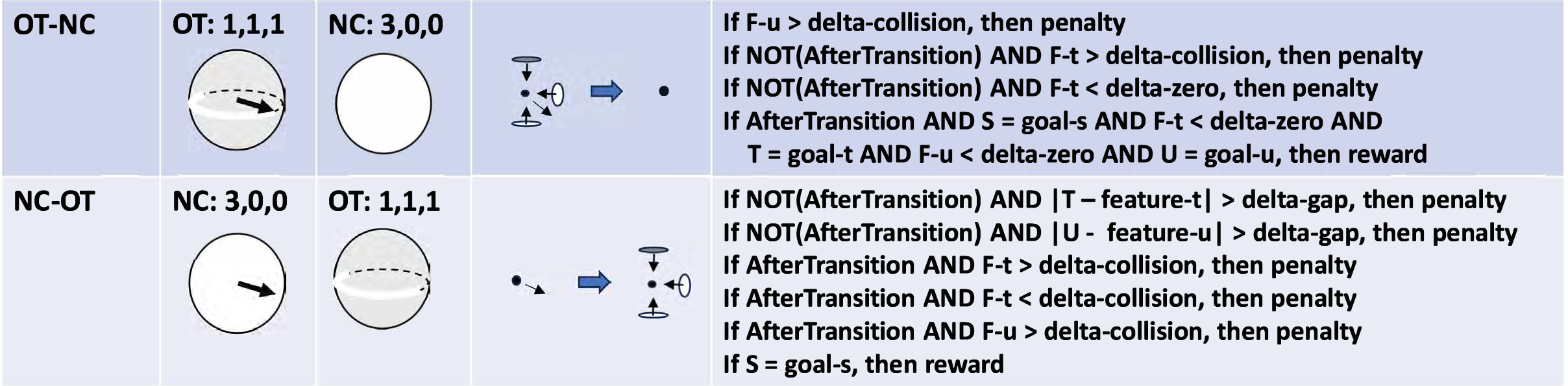}
    \caption{OT-NC and NC-OT skills}
    \label{fig:ot-nc}
\end{figure*}

\subsubsection{OT-PC and PC-OT}
\begin{comment}
先のOT-NC遷移は、中心にある物体が周辺環境と接触を継続しながら運動している間に、この物体取り囲む環境との接触が同時にわるような形状の際に発生する。一方、OT-PC遷移は、この境の接触が、ある１つの方向で継続するような場合に発生する。この際、もともと離脱状態であった面との接触が継続するか、拘束状態を作り出していた面の一部と接触が継続するのかで制御の内容がことなる。（１）は離脱状態が継続する場合、（２）は拘束状態が離脱状態になる場合である。
\end{comment}
The previous OT-NC transition occurs while an object in the center is in motion with maintaining contact with the surrounding environment, the contact ends simultaneously due to the shape of the environment. On the other hand, the OT (M=1, D=1, C=1)-PC (M=2, D=1, C=0) transition occurs, while the object is moving, contact with the environment continues in one direction. In this OT - PC transition, the control differs depending on whether the contact with original detachment surface continues or the contact with a part of the surfaces that created the constraint d-state persists: (a) the detachment d-state continues and (b)~the constraint d-state transits to the detachment d-state. See the details in Appendix.

\begin{figure*}[ht]
    \centering
    \includegraphics[width=\linewidth]{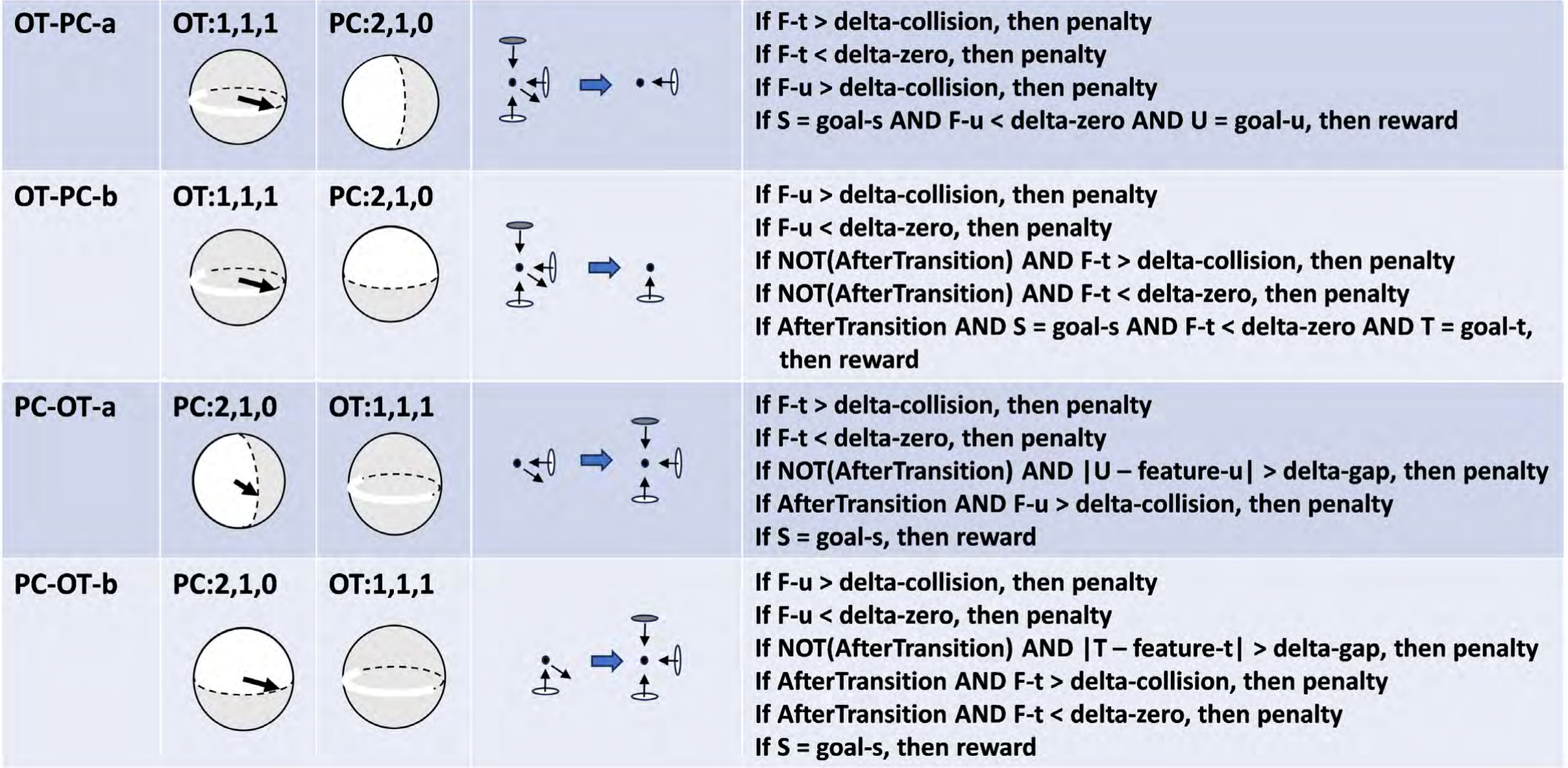}
    \caption{OT-PC and PC-OT skills}
    \label{fig:ot-pc}
\end{figure*}

\subsubsection{OT-TR and TR-OT} 
Regarding the transition of OT (M=1, D=1, C=1) to TR (M=2, D=0, C=1), there are two cases: (a)~motion to break the face contact of the detachment surface and (b)~motion to maintain the face contact to the detachment surface. Also see the details in Appendix.

\begin{figure*}[ht]
    \centering
    \includegraphics[width=\linewidth]{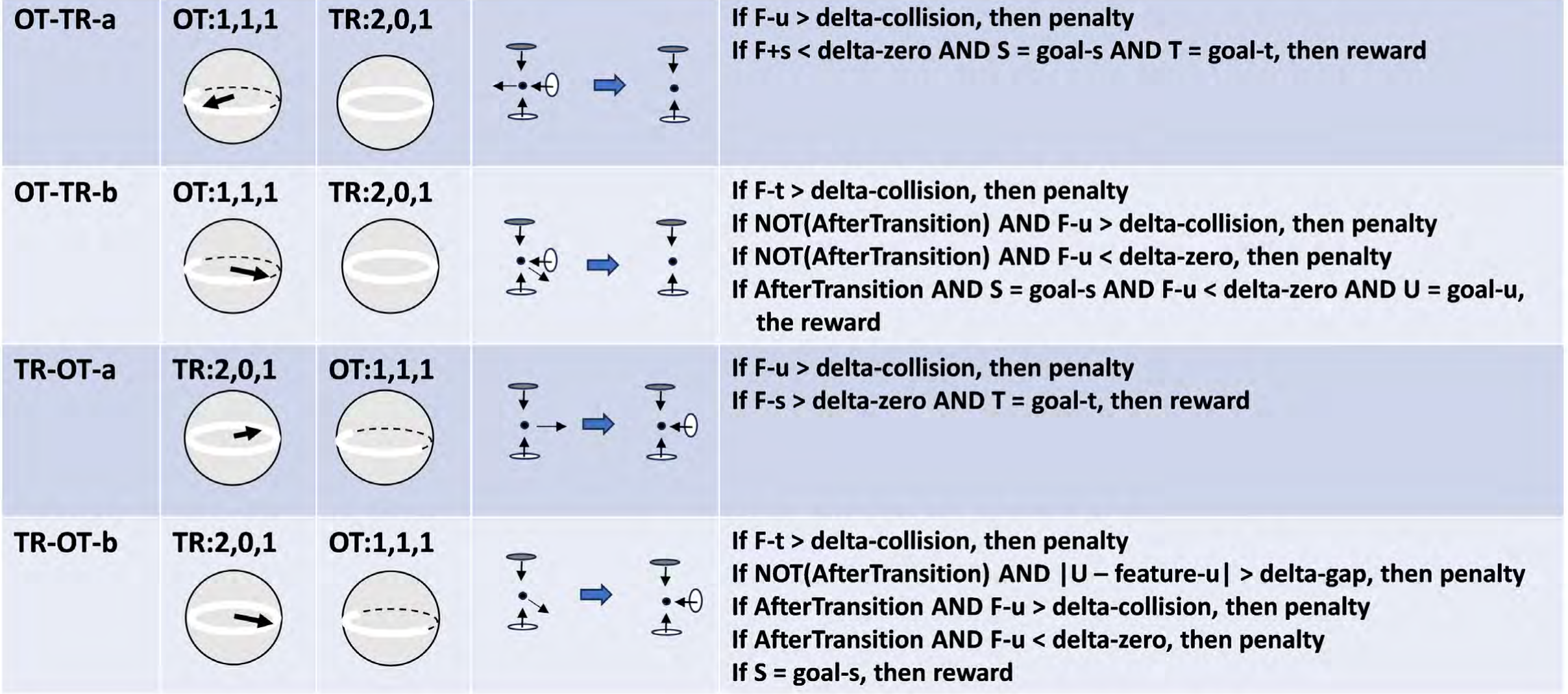}
    \caption{OT-TR and TR-OT skills}
    \label{fig:ot-tr}
\end{figure*}

\subsection{Intrastate transition}
\begin{comment}
本論文では、面接触状態の変化をさせるものがタスクであると定義した。広い意味で、同一の状態に遷移するものも状態遷移の一種であると定義すると、同一状態に移る動きも広義のタスクと定義できる。特に、同一状態を維持する動きは、意味拘束に対応するタスクの実装として利用できる。たとえば、PCからPCに遷移するタスクは、意味的作業のSTG2に対応する。以下では、これらの同一状態に遷移する際の動きを実装する際の報酬関数について解析する。  
\end{comment}

This paper defines a task as one unit of robot motion that causes a transition in surface contact states. In a broader sense, we define a transition to the same state as one type of state transitions. A unit of motion to the same state is also defined as a task. This is because some motions that maintain the same states are important for the implementation of some of semantic tasks~(\cite{ikeuchi2021semantic}). For example, a task that transits from the PC state to the PC state corresponds to the STG2 of semantic tasks, {\it Wiping} task. In the following, we will analyze reward functions for the implementation of these intrastate transition motion.

\subsubsection{NC-NC}

The skill of the robot corresponding to NC (M=3, D=0, C=0) - NC is a {\it Bring} task and PTG12 in \cite{IkeuchiIJCV2018}. It is the unconstrained motion to bring an object from one position to another. 

\begin{figure*}[ht]
    \centering
    \includegraphics[width=\textwidth]{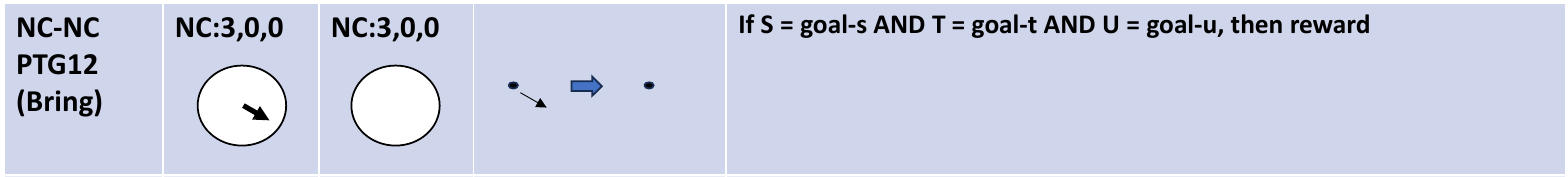}
    \caption{NC-NC skill}
    \label{fig:nc-nc}
\end{figure*}

For the motion direction, the maintenance d-state is maintained, A1 can be applied: 
\begin{small}
\begin{verbatim}
  A1: if S = goal-s, then reward  
\end{verbatim}
\end{small}
For the two dimensions orthogonal to the motion direction, B1 is also applicable since the maintenance d-state is maintained.
\begin{small}
\begin{verbatim}
  B1: if T = goal-t, then reward
  B1: if U = goal-u, then reward
\end{verbatim}
\end{small}
Putting these together:
\begin{small}
\begin{verbatim}
  Reward NC-NC (PTG12 (Bring) task)
    if S = goal-s AND T = goal-t AND
       U = goal-u, then reward
\end{verbatim}
\end{small}
Figure~\ref{fig:nc-nc} shows the summary of the NC-NC skill.

\subsubsection{PC-PC}

The PC state includes three classes of solutions from the Kuhn-Tucker theory (\cite{kuhn1957linear}): PC1 (M=2, D=1, C=0), PC2 (M=1, D=2, C=0), and PCN (M=0, D=3, C=0), whose solution areas are hemispherical, cresentic, and polygonal regions on the Gaussian sphere, respectively. For the sake of analytical simplicity, this paper assumes that interstate transitions only occur from/to PC1 and transitions among PC1, PC2, and PCN are treated as intrastate transitions, which will be analyzed in this subsection.

\begin{figure*}[ht]
    \centering
    \includegraphics[width=\textwidth]{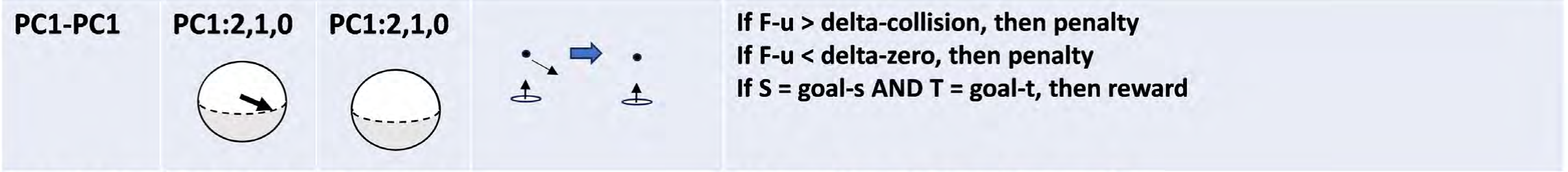}
    \caption{PC1-PC1 skill}
    \label{fig:pc1-pc1}
\end{figure*}

\paragraph{PC1-PC1}
In the intrastate transition from PC1 (M=2, D=1, C=0) to PC1, the object must remain in contact with the contact surface during the motion. In the motion direction, A1 is applicable because the maintenance d-state continues in that direction.
\begin{small}
\begin{verbatim}
  A1: if S = goal-s, then reward
\end{verbatim}
\end{small}
In one of the orthogonal dimension to the motion, B1 is applicable since the maintenance d-state is maintained.
\begin{small}
\begin{verbatim}
  B1: if T = goal-t, then reward
\end{verbatim}
\end{small}
In the remaining dimension, B2 is applicable since the detachment d-state is maintained.
\begin{small}
\begin{verbatim}
  B2: if F-u > delta-collision, then penalty
      if F-u < delta-zero, then penalty
\end{verbatim}
\end{small}
Putting these together, we obtain:
\begin{small}
\begin{verbatim}
  Reward PC1-PC1 (STG2 (Wipe) task)
    if F-u > delta-collision, then penalty
    if F-u < delta-zero, then penalty
    if S = goal-s AND T = goal-t, 
      then reward
\end{verbatim}
\end{small}
Figure~\ref{fig:pc1-pc1} shows the summary of the PC1-PC1 skill.

\paragraph{PC1-PC2}
The transition from PC1 (M=2, D=1, C=0) to PC2 (M=1, D=2, C=0) is caused when an object in motion while maintaining contact with one surface collides with another surface and the motion direction becomes the detachment d-state. Thus, as for the motion direction, A2 is applicable:
\begin{small}
\begin{verbatim}
  A2: if F-s > delta-zero, then reward
\end{verbatim}
\end{small}
In one orthogonal dimension to the motion, B1 is applicable because the maintenance d-state is maintained.
\begin{small}
\begin{verbatim}
  B1: if T = goal-t, then reward
\end{verbatim}
\end{small}
In the remaining dimension that maintains the contact, the detachment d-state is maintained and B2 is applicable.
\begin{small}
\begin{verbatim}
  B2: if F-u > delta-collision, then penalty
      if F-u < delta-zero, then penalty
\end{verbatim}
\end{small}
We obtain:
\begin{small}
\begin{verbatim}
  Reward PC1-PC2
    if F-u > delta-collision, then penalty
    if F-u < delta-zero, then penalty
    if F-s > delta-zero AND T = goal-t,
      then reward
\end{verbatim}
\end{small}

\paragraph{c) PC2-PC1}

\begin{figure*}[ht]
    \centering
    \includegraphics[width=\linewidth]{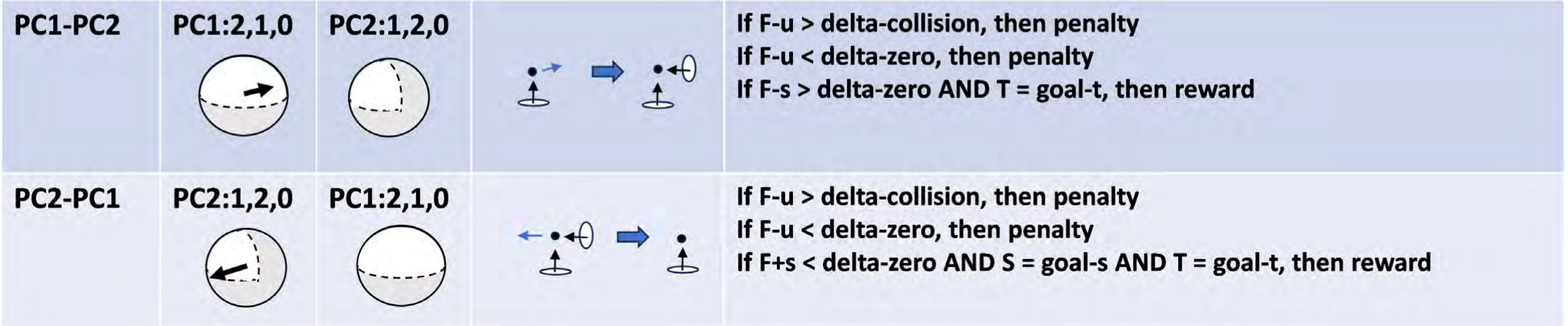}
    \caption{PC1-PC2 and PC2-PC1 skills}
    \label{fig:pc1-pc2}
\end{figure*}

A3 is applicable to the motion direction because the contact surface will be broken due to the motion and the transition from the detachment d-state to the maintenance d-state occurs:
\begin{small}
\begin{verbatim}
  A3: if F+s < delta-zero and S = goal-s, 
        then reward
\end{verbatim}
\end{small}
In one orthogonal dimension to the motion, the maintenance d-state is maintained and B1 is applicable:
\begin{small}
\begin{verbatim}
  B1: if T = goal-t, then reward
\end{verbatim}
\end{small}
In another orthogonal dimension to the motion, the object maintains the surface contact during the motion and the detachment d-state is maintained. B2 is applicable:
\begin{small}
\begin{verbatim}
  B2: if F-u > delta-collision, then penalty
      if F-u < delta-zero, then penalty
\end{verbatim}
\end{small}
We can summarize:
\begin{small}
\begin{verbatim}
  Reward PC2-PC1
    if F-u > delta-collision, then penalty
    if F-u < delta-zero, then penalty
    if F+s < delta-zero AND S = goal-s AND
       T = goal-t, then reward
\end{verbatim}
\end{small}
Figure~\ref{fig:pc1-pc2} shows the summary of the skills of PC1-PC2 and PC2-PC1.

\paragraph{PC2-PC2}

\begin{figure*}[ht]
    \centering
    \includegraphics[width=\linewidth]{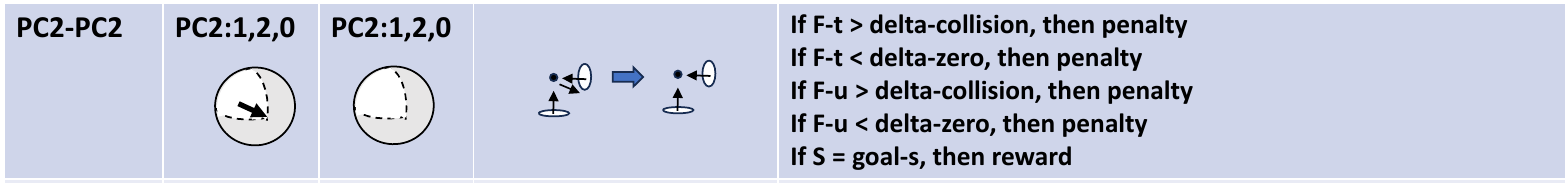}
    \caption{PC2-PC2 skill}
    \label{fig:pc2-pc2}
\end{figure*}

In PC2 (M=1, D=2, C=0), a one-dimensional maintenance d-state exists. Along this dimension, a motion to maintain the state is possible. Thus, A1 is applicable.
\begin{small}
\begin{verbatim}
  A1: if S = goal-s, then reward
\end{verbatim}
\end{small}
Concerning the two dimensions orthogonal to the motion, the detachment d-state is maintained with keeping contact to the environment surfaces. B2 is applicable.
\begin{small}
\begin{verbatim}
  B2: if F-t > delta-collision, then penalty
      if F-t < delta-zero, then penalty
  B2: if F-u > delta-collision, then penalty
      if F-u < delta-zero, then penalty
\end{verbatim}
\end{small}
In summary,
\begin{small}
\begin{verbatim}
  Reward PC2-PC2
    if F-t > delta-collision, then penalty
    if F-t < delta-zero, then penalty
    if F-u > delta-collision, then penalty
    if F-u < delta-zeron, then penalty
    if S = goal-s, then reward
\end{verbatim}
\end{small}
Figure~\ref{fig:pc2-pc2} shows the PC2-PC2 skill.

\paragraph{PC2-PCN}

PC2 (M=1, D=2, C=0) has one maintenance dimension that allows motion while maintaining contact with two different surfaces. This dimension corresponds to the base of the crescent-shaped cone of a Gaussian sphere. When the object collides with a third surface during motion along this maintenance dimension, the transition from PC2 to PCN (M=0, D=3, C=0) occurs.

Regarding the motion direction, since the transition from the maintenance d-state to the detachment d-state occurs, A2 is applicable:
\begin{small}
\begin{verbatim}
  A2: if F-s > delta-zero, then reward
\end{verbatim}
\end{small}
In the two dimensions orthogonal to the motion direction, the detachment d-state is maintained, so B2 is applicable:
\begin{small}
\begin{verbatim}
  B2: if F-t > delta-collision, then penalty
      if F-t < delta-zero, then penalty
  B2: if F-u > delta-collision, then penaly
      if F-u < delta-zero, then penalty
\end{verbatim}
\end{small}

In summary,
\begin{small}
\begin{verbatim}
  Reward PC2-PCN
    if F-t > delta-collision, then penalty
    if F-t < delta-zero, then penalty
    if F-u > delta-collision, then penalty
    if F-u < delta-zero, then penalty
    if F-s > delta-zero, then reward
\end{verbatim}
\end{small}

\paragraph{PCN-PC2}

There is no maintenance dimension in PCN (M=0, D=3, C=0), all the possible motions are in the detachment dimensions. Among those possible motions, the transition from PCN to PC2 occurs when the object departs from one surface while maintaining contact with the remaining surfaces. Therefore, concerning the motion direction, the transition from a detachment d-state to a maintenance d-state takes place. A3 is applicable:
\begin{small}
\begin{verbatim}
  A3: if F+s < delta-zero AND S = goal-s,
        then reward
\end{verbatim}
\end{small}
In the two dimensions orthogonal to the motion direction, the detachment d-state is maintained, so B2 is applicable.
\begin{small}
\begin{verbatim}
  B2: if F-t > delta-collision, then penalty
      if F-t < delta-zero, then penalty
  B2: if F-u > delta-collision, then penalty
      if F-u < delta-zero, then penalty
\end{verbatim}
\end{small}

\begin{figure*}[ht]
    \centering
    \includegraphics[width=\linewidth]{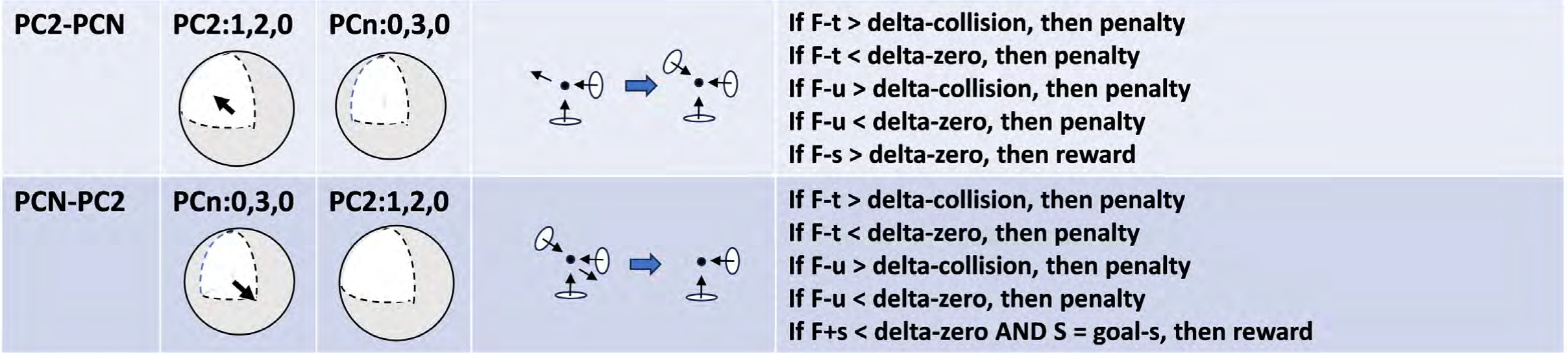}
    \caption{PC2-PCN and PCN-PC2 skills}
    \label{fig:pc2-pcn}
\end{figure*}

In summary,
\begin{small}
\begin{verbatim}
  Reward PCN-PC2
    if F-t > delta-collision, then penalty
    if F-t < delta-zero, then penalty
    if F-u > delta-collision, then penalty
    if F-u < delta-zero, then penalty
    if F+s < delta-zero and S = goal-s,
      then reward
\end{verbatim}
\end{small}
Figure~\ref{fig:pc2-pcn} shows the skills of PC2-PCN and PCN-PC2.

\subsubsection{TR-TR}

\begin{figure*}[ht]
    \centering
    \includegraphics[width=\linewidth]{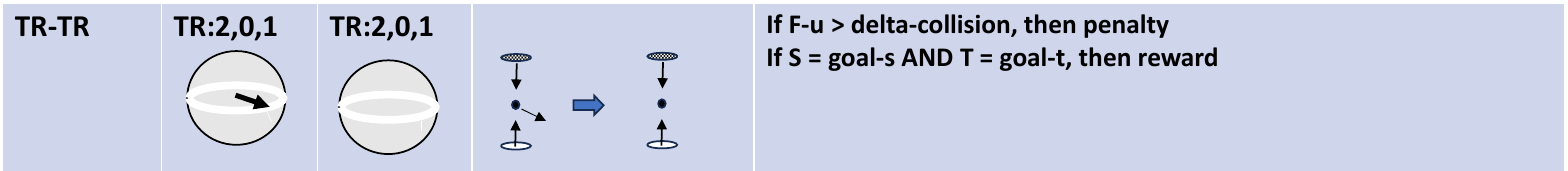}
    \caption{TR-TR skill}
    \label{fig:tr-tr}
\end{figure*}

TR (M=2, D=0, C=1) - TR is a transition where an object sandwiched between two walls moves between them. The reward function is given in Figure~\ref{fig:tr-tr}. For the detailed derivation, please refer to Appendix B.

\subsubsection{OT-OT}
\begin{figure*}[ht]
    \centering
    \includegraphics[width=\linewidth]{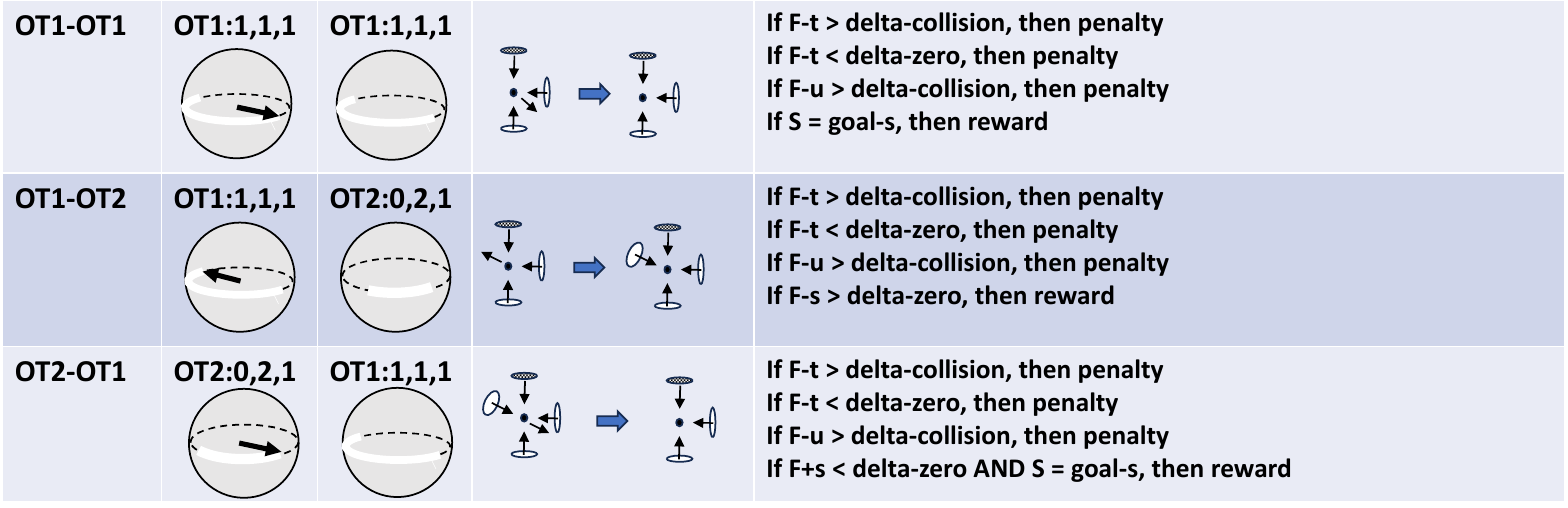}
    \caption{OT-OT skills}
    \label{fig:ot-ot}
\end{figure*}

In the OT state, there are two sub-classes: OT1 (M=1, D=1, C=1) and OT2 (M=0, D=2, C=1). Similar arguments to the case of the PC state can be applied to transitions between these sub-classes. Reward functions for these skills are given in Figure~\ref{fig:ot-ot}. For detailed derivation, please refer to Appendix B.

%\par
%\newpage

\subsection{Rotational skills}

For rotational motion, states can be defined in Figure~\ref{fig:contact-states}~(b), and skills can be defined for transitions between these states shown in Figure~\ref{fig:translation-tasks}~(b).

In the case of translational motion, the direction of the screw axis coincided with the direction of motion of the object. In the case of rotational motion, on the other hand, the axis direction and the direction of motion are orthogonal, and in a finite interval of motion, the trajectory is a curvilinear motion. Nevertheless, the infinitesimal motion at each infinitesimal unit time can be assumed to be a translational motion orthogonal to the axis. By considering a local coordinate system in which the direction of motion at that time is S and the orthogonal directions to the motion direction are T and U, we can develop an argument similar to that for the translational case.

We will take an example, the OR (M = 0, D = 1, C = 2) - RV (M = 1, D = 0, C = 2) transition, corresponding to a Door-opening task, which appears particularly frequently. OR corresponds to a state where the door is closed, and rotation is possible only in the direction of detachment (\ie, opening direction). The transition in the direction of rotation is from the detachment d-state to the maintenance d-state, and A3 can be applied. In other words, the drag force in the direction of rotation, F+s, is eliminated, and the skill concludes when a certain predetermined rotation angle, \hbox{goal-s}, given by the demonstration, is reached.
\begin{small}
\begin{verbatim}
  A3: if F+s < delta-zero AND S = goal-s, 
        then reward
\end{verbatim}
\end{small}

The two other dimensions are both constrained, resulting in B3. In other words, attempting to rotate with infeasible displacement forcibly will generate drag forces, F-t and F-u. Therefore, it is necessary to maintain these forces below a certain threshold.
\begin{small}
\begin{verbatim}
  B3: if F-t > delta-collision, then penalty
  B3: if F-u > delta-collision, then penalty
\end{verbatim}
\end{small}

In summary, 
\begin{small}
\begin{verbatim}
  Reward OR-RV (PTG51 (Door-open) task)
    if F-t > delta-collision, then penalty
    if F-u > delta-collision, then penalty
    if F+s < delta-zero AND S = goal-s
      then reward
\end{verbatim}
\end{small}

A similar consideration yields the following reward functions for RV-OR and RV-RV skills.
\begin{small}
\begin{verbatim}
  Reward RV-OR (PTG53 (Door-close) task)
    if F-t > delta-collision, then penalty
    if F-u > delta-collision, then penalty
    if F-s > delta-zero, then reward
\end{verbatim}
\end{small}
\begin{small}
\begin{verbatim}
  Reward RV-RV (PTG52 (Door-adjust) task)
    if F-t > delta-collision, then penalty
    if F-u > delta-collision, then penalty
    if S = goal-s, then reward
\end{verbatim}
\end{small}

\section{Implementation of skill-agent library}

\subsection{Current skill-agent library}

The manipulation skill-agent library currently consists of a commonly used set of skill agents, including 
\begin{itemize}
    \item PC-NC-a (PTG11, Pick) skill agent
    \item NC-NC   (PTG12, Bring) skill agent
    \item NC-PC-a (PTG13, Place) skill agent
    \item OP-PR (PTG31, Drawer-open) skill agent
    \item PR-PR (PTG32, Drawer-adjust) skill agent
    \item PR-OP (PTG33, Drawer-close) skill agent
    \item OR-RV (PTG51, Door-open) skill agent
    \item RV-RV (PTG52, Door-adjust) skill agent
    \item RV-OR (PTG53, Door-close) skill agent
    \item PC1-PC1 (STG2, Wipe) skill agent
\end{itemize}
This set adequately allows our home service robot to perform tasks. However, if needed, it can be extended using the method outlined in this paper.

In order to build an end-to-end system, grasping skill agents are also necessary. The grasping skill agents described in~\cite{saito2022taskgrasping} have been implemented for Shadow-hand and Fetch Parallel-gripper in the grasp skill-agent library. Currently, this library includes:
\begin{itemize}
    \item Active-force grasp skill agent
    \item Passive-force grasp skill agent
    \item Lazy-closure grasp skill agent
\end{itemize}

\begin{figure}
    \centering
    \includegraphics[width=\linewidth]{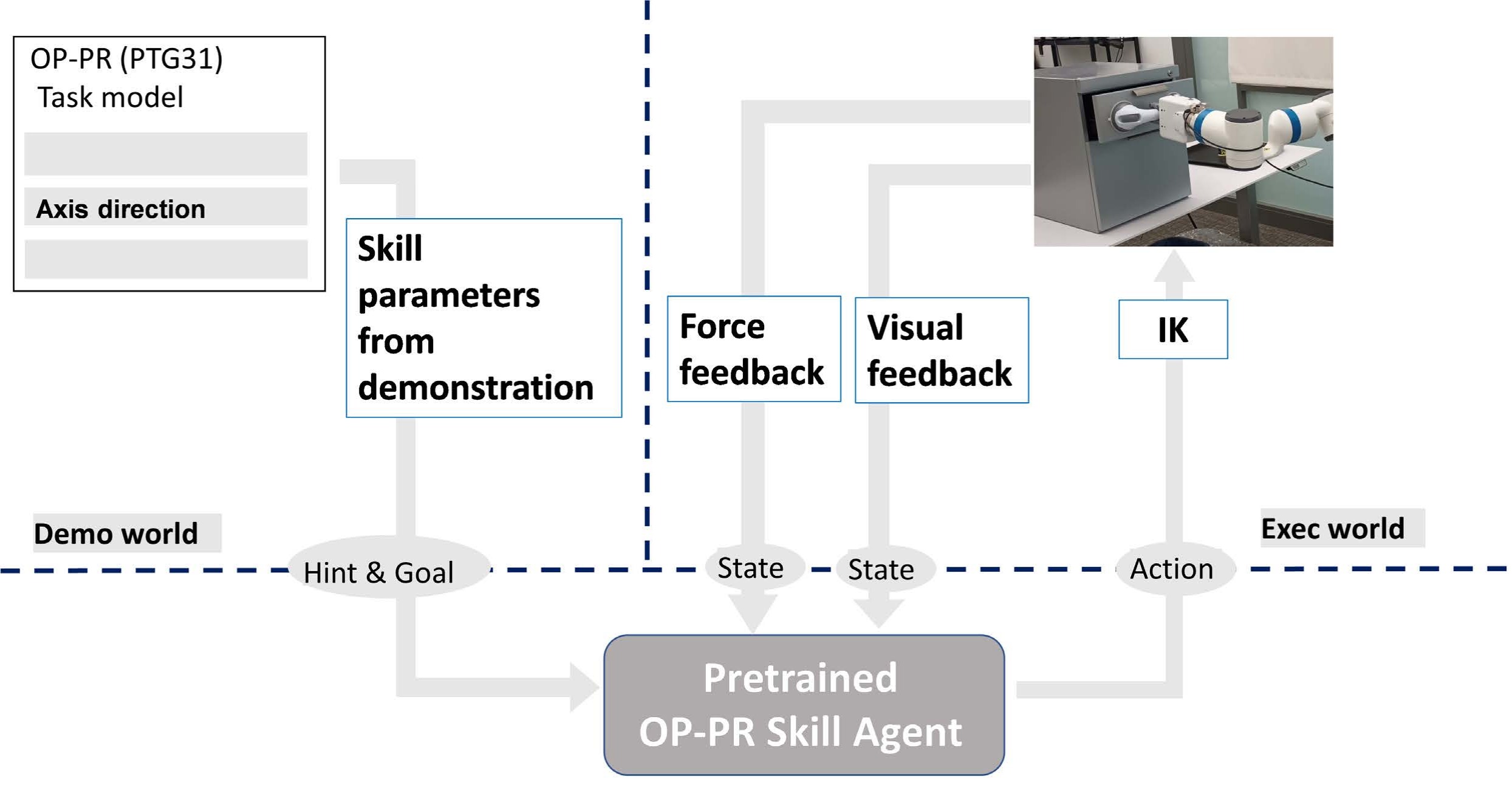}
    \caption{Task and skill agent}
    \label{fig:skill-exec}
\end{figure}

The execution of a skill agent, in the case of the OP-PR task, is illustrated in Figure~\ref{fig:skill-exec}. First, a sequence of task models is generated from the human demonstration. Each task model has the skill parameters to execute that task based on the demonstration. In the OP-PR task, the direction of pushing the drawer is stored as Axis direction.

%The required skills corresponding to the task model sequence are retrieved from the skill library, and a corresponding skill sequence is generated. Each skill receives parameters from the task model that are necessary at runtime. For example, in the case of OP-PR skill, the direction in which the demonstrator pulled the drawer is passed to it as the initial pulling direction.

\begin{comment}
During the execution of the OP-PR skill, 
\begin{verbatim}
    if F-t > delta-collision, then penalty
    if F-u > delta-collision, then penalty
\end{verbatim}
under this policy, even if there is an error in the initial drawer-pulling direction stored in the Axis direction slot, the skill continuously obtains the drag force perpendicular to the pulling direction and determines the next position of the hand for correcting the direction. Once the next hand position is obtained, this is given to the IK solver to determine the overall posture of the robot. 

\begin{verbatim}
    if F+s < delta-zero AND S = goal-s, then reward
\end{verbatim}
From the reward policy, once the final position is achieved, the OP-PR skill finishes.
\end{comment}

\subsection{Reinforcement learning environment}

\begin{figure}
    \centering
    \includegraphics[width=0.8\linewidth]{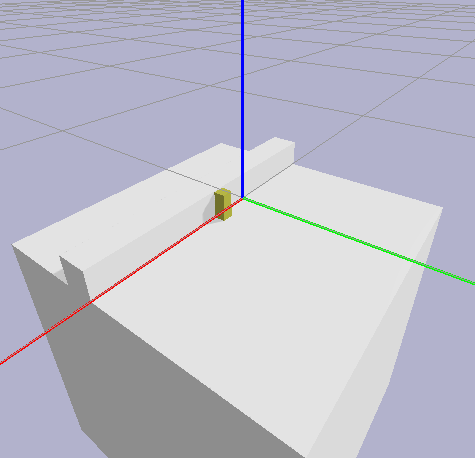}
    \caption{Simulation environment of PC2-PC2: the yellow box is the manipulated object and the white object is the environmental object.}
    \label{fig:sim_env}
\end{figure}

Many skill agents require prior learning policies to adjust motion directions based on the force feedback with reward functions. For training those skill agents, we developed a reinforcement learning environment that parallelized the PPO algorithm in {\em Stable Baselines3}\footnote{\url{https://stable-baselines3.readthedocs.io/en/master/}}.

To build the learning environment, we used {\em PyBullet}\footnote{\url{https://pybullet.org/wordpress/}}
as a simulator and obtained the necessary policies for each skill agent through reinforcement learning. For example, we set up the environment shown in Figure~\ref{fig:sim_env} in the simulator that satisfies the surface contact state. Gravity is set to 0 to keep the environment as simple as possible. We assumed that the object is already grasped by a hand and they are integrated into one unit. Since the estimation of the normal direction of the contact surface may contain some errors, domain randomization was used to add the errors in the training. The contact was assumed to be represented by an elastic body, and hardware compliance was simulated by setting the spring and damper coefficients. 

The control side of each skill agent is described by a force relationship. It is assumed that the actual robot is equipped with a 6-axis force sensor. However, due to the difference in physical parameters between the simulator and the actual machine, it is difficult to simulate the magnitude of the force accurately. Therefore, to ensure that the policy obtained in the simulation can be implemented on the actual robot, we used the unit vector of the force $\mathbf{f}$ %, and torque, $\mathbf{\tau}$, 
instead of the original value as the states for the reinforcement learning.
\begin{eqnarray}
\mathbf{f}_{n} & = & \mathbf{f} / |\mathbf{f} |.
% \mathbf{\tau}_{n} & = & \mathbf{\tau} / | \mathbf{f} |
\end{eqnarray}
The values of the force sensor shall be represented by converting them to the world coordinate system using the posture of the object obtained on the simulator and the FK of the actual machine.

In some cases, coarsely discretized force magnitudes were used to convey force magnitude information to the reinforcement learning.
\begin{equation}
f_{desc} = \lfloor | \mathbf{f} | / f_{step} \rfloor.
\end{equation}
Here, $f_{step}$ is an arbitrary constant that determines the degree of discretization, and different values are used for simulator and the actual machine to bridge the gap between the simulator and the actual machines.

\if 0
また、ハンドと物体や実機にはある程度のハードウェアコンプライアンスがあるものとし、各タイムステップで小さな移動量に基づく位置制御で行動が実現できると仮定する。
これはシミュレータと実機で異なる値とする。実機の場合は経験的に調整した値を利用した。論文~\cite{saito2024compmanip}と同じく、手と物体は一体化しているという仮定を置き、シミュレータでは物体にかかる力、実機では○○。
\fi

\begin{comment}
進行方向の状態変化の検は実機でも比較的容易であるため、学習は同一接触状態内の遷移を実現するskillを実装し、terminationの条件を変えることにより実現する。例えばPC1-PC1とPC1-PC2のスキルはPC1の状態を維持する動作をさせつつ、PC1-PC1の場合は目標位置まで到達（s = goal-s）すればterminate、PC1-PC2の場合は、反力を検出（F-s $>$ delta-zero）すればterminateすることとする。これにより、学習するべきスキルの数を減らすことができる。例えば、運動方向に直交する方向の状態が変化しない (B1, B2, B3) と仮定すれば、並進の場合、PC1-PC1, PC2-PC2, TR-TR, OR-OR、PR-PR の5つのスキルを学習させればよい。

運動に直交する方向の状態がB1の場合は何の制約もなく、B2の場合は接触を維持し続ける必要があること、すなわちF-uもしくはF-tがdelta-zero以上、delta-collision以下である必要がある。そこでそれらの間の適当な値$f_c$を目標力とすることで達成する。また、B3の場合はその方向にかかる力を最小にすることで達成する。
\end{comment}

\if 0
並進のスキルに関して述べる。状態としては接触を表現するに足りえる情報、つまり、接触している面それぞれの法線（PC2の場合は２つ）を状態として持たせ、それに合わせて、最終状態に至るまでの変位の方向、力の方向、粗々に離散化された大きさを状態とする。行動としては接触面の法線方向に関する行動修正量を出力し、目標変位＋修正量が毎回の移動量となる。報酬に関しては現在の力の値と目標力の差が離れるにつれて負の報酬を与え、離れすぎる（e.g.過剰な力の発生）と大きなペナルティを与えてエピソード終了とする。またB2に関しては接触を維持し続けることが必須であり、simulatorを用いた場合、接触の有無を毎回計算することができるので、目標とする接触が達成できなければ大きなペナルティを与えてエピソード終了とする。この方針により上述の5つのスキルの状態、行動、報酬は図~\ref{fig:rl_design}のように決定することができる。
\begin{figure}
\begin{center}
 Now printing
 \caption{State, action, and reward for each skill} % 各スキルにおける状態・行動・報酬
 \label{fig:rl_design}
\end{center}
\end{figure}
\fi

\subsection{Skill agents}

\subsubsection{PC-NC-a (Pick), NC-NC (Bring), NC-PC-a (Place) skill agents}
These skills are considered to be the most basic robot skills. Reinforcement learning is not required for this group. 

\begin{figure}[ht]
    \centering
    \includegraphics[width=\linewidth]{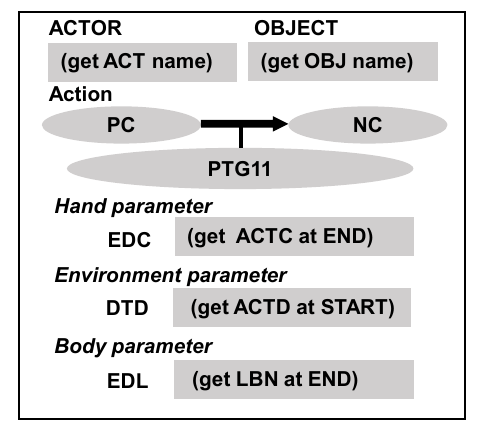}
    \caption{PTG11 (PC-NC-a) task model}
    \label{fig:ptg11-task}
\end{figure}

The PC-NC-a skill agent can be described by the following control rule.
\begin{small}
\begin{verbatim}
  if F+s < delta-zero AND S = goal-S AND 
     T = goal-t AND U = goal-U, then reward
\end{verbatim}
\end{small}
The skill corresponds to the task model shown in Figure~\ref{fig:ptg11-task}~(\cite{ikeuchi2021semantic}). In the figure, (get ACT name) {\it etc.} are daemon functions to obtain the respective values and to store in these slots:
\begin{itemize}
    \item ACTOR slot - right hand or left hand
    \item OBJECT slot - object name
    %\item STC slot - Hand configuration at the start of the task obtained from the demonstration (STart Configuration)
    \item EDC slot - Hand configuration at the end of the task obtained from the demonstration (EnD Configuration)
    \item DTD slot - Direction of hand movement at the start of the task obtained from the demonstration (DeTachment Direction)
    \item EDL Labanotation of human pose at the end of the task obtained from the demonstration (EnD Labanotation (\cite{IkeuchiIJCV2018}))
\end{itemize}
Since the position of the hand before the execution of the skill is known from the end position of the previous skill, the values of goal-s, goal-t, ant goal-u can be calculated using the value in the EDC slot. The skill ends when the target position is reached.
The NC-NC skill agent is implemented in the same way.

\begin{comment}
前述の通り、進行方向Maintain $ \rightarrow $Detachingの変化は瞬間であるため、NC-NCのスキルを実装し、終了条件 $ > delta-zero $を検出する部分をマニュアルでプログラムすることでNC-PCスキルを実装した。NC-NCにおいて力学的な制約は全く存在しないため、特にRLによる学習を必要としなかった。目標とする変位方向に動かしながら、$ > delta-zero $条件が満たされればタスク終了とするプログラムを作成した。

ここでは実装されたスキルが正しく動いていることを確認するためにFetch-Parallelで実行した結果を示す。図~\ref{fig:place_example}にPlaceスキル実行の様子を示す。Placeスキルはplaceされた瞬間を除きすべての運動自由度がmaintainである。目標変位はスキルパラメータとして与えられるとして、目標力がdelta-zeroを超えたかどうかを判断することで動作の終了を検知した。図~\ref{fig:force_profile_pl}は力センサの値の変化を示す。重力の影響を除去するため、PTG13開始時の力の状態を保持しておき、そこからの差分を計算する。また、FKにより得られたロボット座標系における力センサ座標系の関係を用いることで、世界座標系での力の値に変換してある。図はz軸方向（鉛直上向き）の力の変化を示している。把持物体が面と接触することにより反力が発生し、鉛直上向きに力が増加していることが確認できる。delta-zeroとして30[N]を設定したためタイムステップ22で接触を検知、Placeスキルを正しく終了することができた。
\end{comment}

The NC-PC-a skill agent, also called PTG13 (Place), is also almost the same as the PC-NC-a skill agent except for the termination condition. The end position obtained from the demonstration includes an observation error, thus, the control algorithm,
\begin{small}
\begin{verbatim}
    if F-s > delta-zero AND T = goal-t AND 
       U = goal-U, then reward
\end{verbatim}
\end{small}
moves the hand to the target position given by the demonstration, while the skill ends when the value of the drag force against the direction of the motion exceeds the threshold value.

\begin{figure}[ht]
    \centering
     \includegraphics[width=1.0\linewidth]{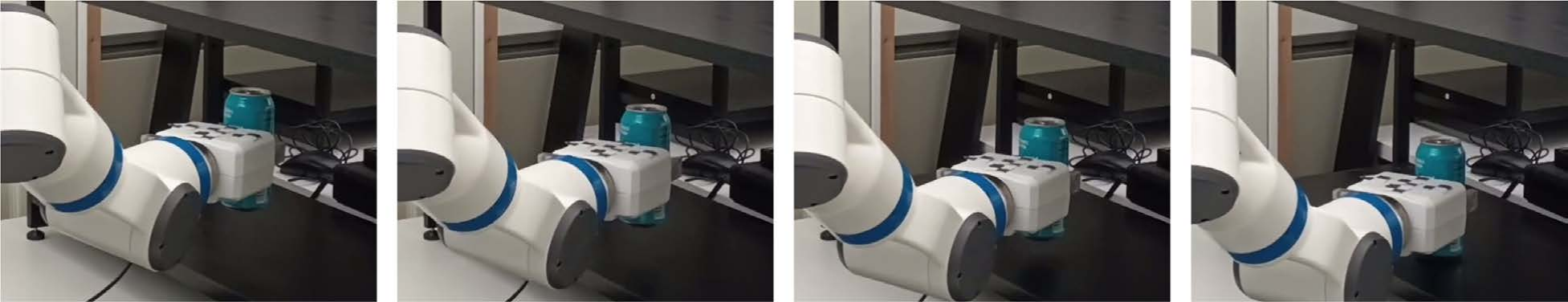}
    \caption{Execution of NC-PC-a (PTG13, Place) skill agent}
    \label{fig:place_example}
\end{figure}

\begin{figure}[ht]
    \centering
    \includegraphics[width=0.8\linewidth]{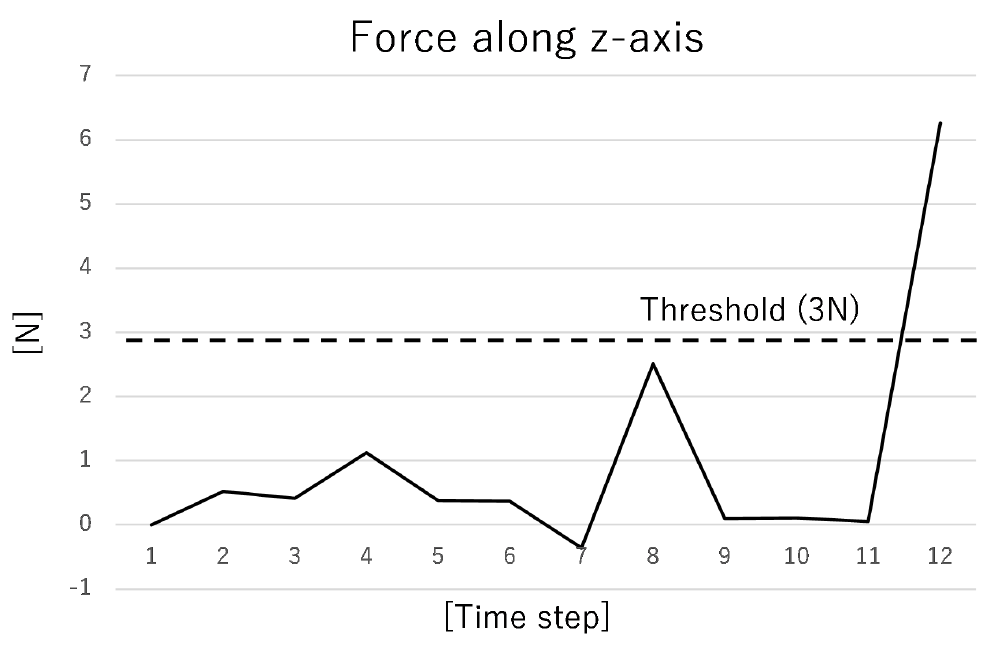}
    \caption{Force profile along z-axis in NC-PC-a skill}
    \label{fig:force_profile_pl}
\end{figure}

Here we show the results of the execution using a real robot to confirm if the implemented skill agents are working correctly. Figure~\ref{fig:place_example} shows the execution of the NC-PC-a skill agent. Figure~\ref{fig:force_profile_pl} shows the change in force sensor values. To remove the effect of gravity, the force at the start of the skill is retained and the difference from it is calculated; by using the relationship between the force sensor coordinate system in the robot coordinate system obtained from FK, the force values are converted to those in the world coordinate system. Figure~\ref{fig:force_profile_pl} shows the changes in the z-axis direction (vertical upward). It can be confirmed that the force increases due to the drag force generated when the grasping object comes into contact with the surface. Delta-zero was set to 3~[N]. The contact was detected at time 12 and this skill was completed correctly.

\subsubsection{OP-PR (Drawer-open), PR-PR (Drawer-adjust), PR-OP (Drawer-close) skill agents}

These skill agents can be described by the following control rule without the termination condition.
\begin{small}
\begin{verbatim}
    if F-t > delta-collision, then penalty
    if F-u > delta-collision, then penalty
\end{verbatim}
\end{small}
All the skills need to satisfy the same rule. Generally, the skill parameters (\eg, drawing direction) observed in the demonstration include errors. Thus, it is necessary to train these skill agents using RL. In order to reduce the training effort, we train the PR-PR skill agent first, and then add the program to decide the terminal condition. 

%なる条件を満たす必要があるという点は共通である。一般に、一般にデモから得られる引き出す方向や押し込むには観測誤差がのるため、ペナルティー項を利用したスキルの強化学習が必要となる。強化学習の回数を減らすため、まずこの部分、すなわち PR-PR、の部分を学習したのち、終了条件をくわえた。

The state in PR can be represented by setting the direction of the feasible displacement. The skill agnet needs the force information for feedback in the execution. Thus, the state in RL can be designed as follows:
\begin{itemize}
\item the direction of the feasible displacement: $ \mathbf{c}_t $,
\item the unit vector of the force: $ \mathbf{f}_t / | \mathbf{f}_t| $,
\end{itemize}
where $ \mathbf{f}_t $ is the force value in time $ t $. The action in RL is the modification of the currently estimated direction, $ \Delta \mathbf{c} \equiv (\Delta c_x, \Delta c_y, \Delta c_z) $. Given the modification, the displacement direction at time $ t + 1 $ is calculated by the following equation:
\begin{displaymath}
    \mathbf{c}_{t+1} = \frac{\mathbf{c}_t + \Delta \mathbf{c}}{|\mathbf{c}_t + \Delta \mathbf{c}|}.
\end{displaymath}

\begin{comment}
PRの状態は、可能な変位方向を指定することで表現できる。またフィードバックのため力の情報を必要とする。つまりRLにおける状態は以下のようになる。
\begin{itemize}
\item 可能な変位の方向: $ \mathbf{c}_t $
\item 力を正規化したもの: $ \mathbf{f}_t / | \mathbf{f}_t| $
\end{itemize}
ただし$ \mathbf{f}_t $は時刻$ t $における力センサの値である。
行動は現在推定された可能な変位方向に対する修正量$ \Delta \mathbf{c} \equiv (\Delta c_x, \Delta c_y, \Delta c_z) $である。実際に修正量が与えられたとき、時刻$ t + 1$の変位の方向$\mathbf{c}_{t+1}$は以下の式で計算される。
\begin{displaymath}
    \mathbf{c}_{t+1} = \frac{\mathbf{c}_t + \Delta c}{|\mathbf{c}_t + \Delta c|}
\end{displaymath}
\end{comment}

\begin{figure}
    \centering
    \includegraphics[width=0.8\linewidth]{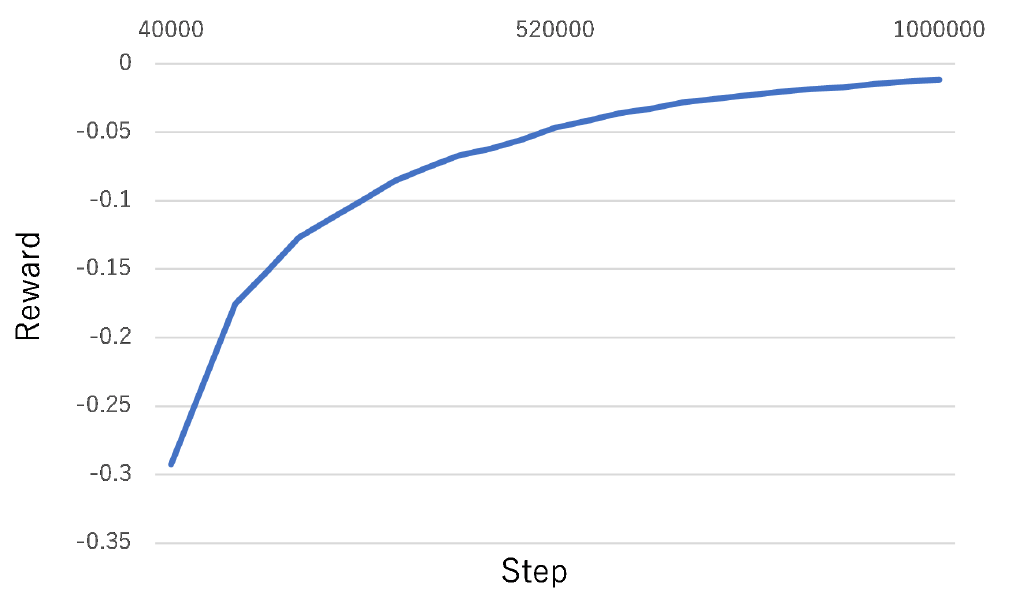}
    \caption{Reward curve in PR-PR skill agent}
    \label{fig:rc_tr}
\end{figure}

The PR-PR skill agent requires to satisfy F-t $\leq$ delta-collision and  F-u $\leq$ delta-collision. If the force exerted by the constraint would be reduced, these two conditions tend to be satisfied. Thus, the reward function $ r $ can be formulated as follows:
\begin{equation}
r = - |\mathbf{f}_{t}|.
\end{equation}
The episode in RL training is terminated after the predefined duration. Figure~\ref{fig:rc_tr} shows the reward curve. The training was terminated after one million steps. The reward curve looks converging.

\begin{comment}
PR-PRスキルはF-t $\leq$ delta-constraint, F-u $\leq$ delta-constraintを満たすことを要求する。接触により発生する力をできるだけ小さくすればこれらの条件が満たされる可能性が高い。
これより、報酬$ r $は
\begin{equation}
r = - |f_{s}|
\end{equation}
となる。強化学習の際は一定時間たつとエピソードを終了することとした。図~\ref{fig:rc_tr}に報酬曲線を示す。今回は経験的に100万ステップを回して学習終了ということとした。学習が収束していることが見て取れる。
\end{comment}

\begin{figure}
    \centering
    \includegraphics[width=\linewidth]{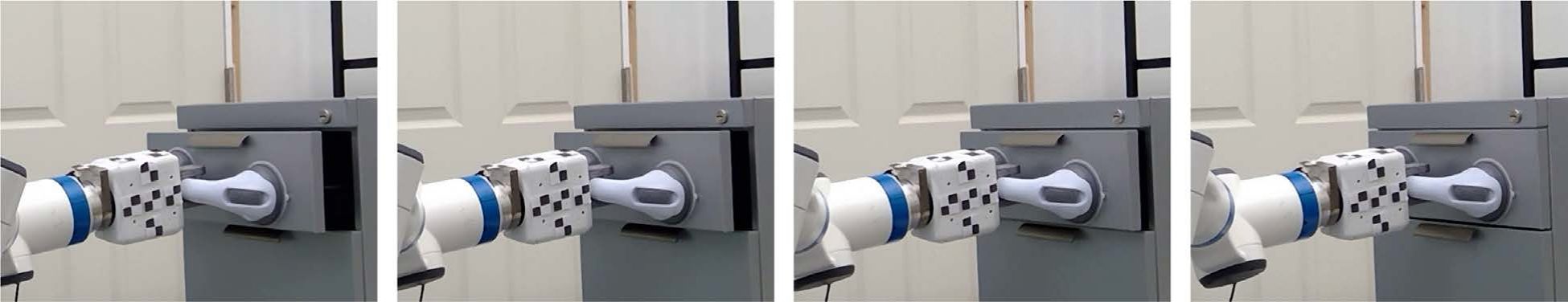}
    \caption{Execution of PR-OP (Drawer-close) skill agent}
    \label{fig:drawer_close}
\end{figure}

\begin{figure}
    \centering
    \includegraphics[width=0.8\linewidth]{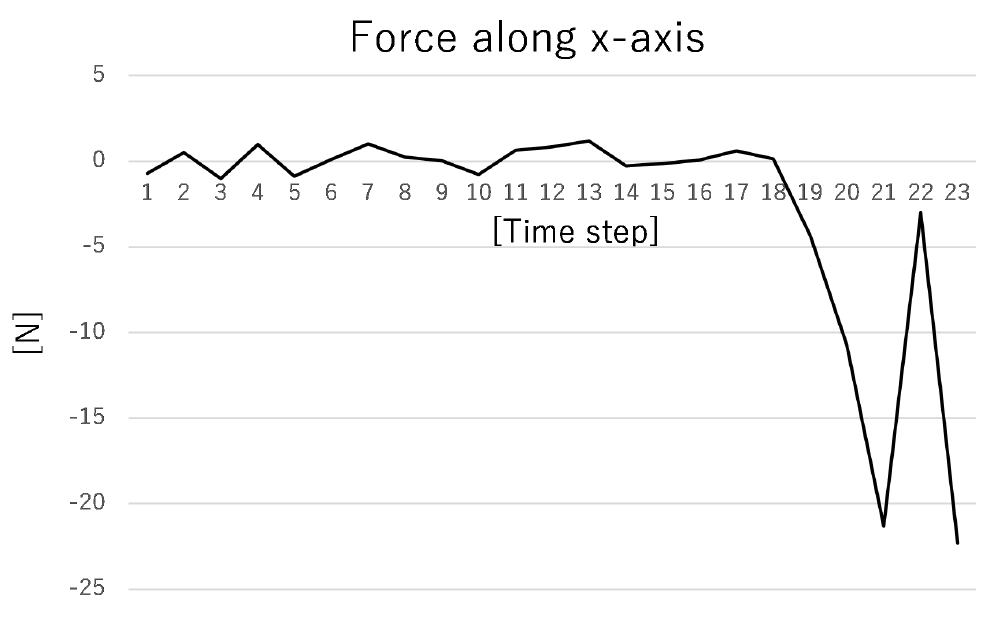}
    \caption{Force profile along x-axis (drawing direction) in PR-OP skill}
    \label{fig:force_profile_dc2}
\end{figure}

\begin{figure}
    \centering
    \includegraphics[width=\linewidth]{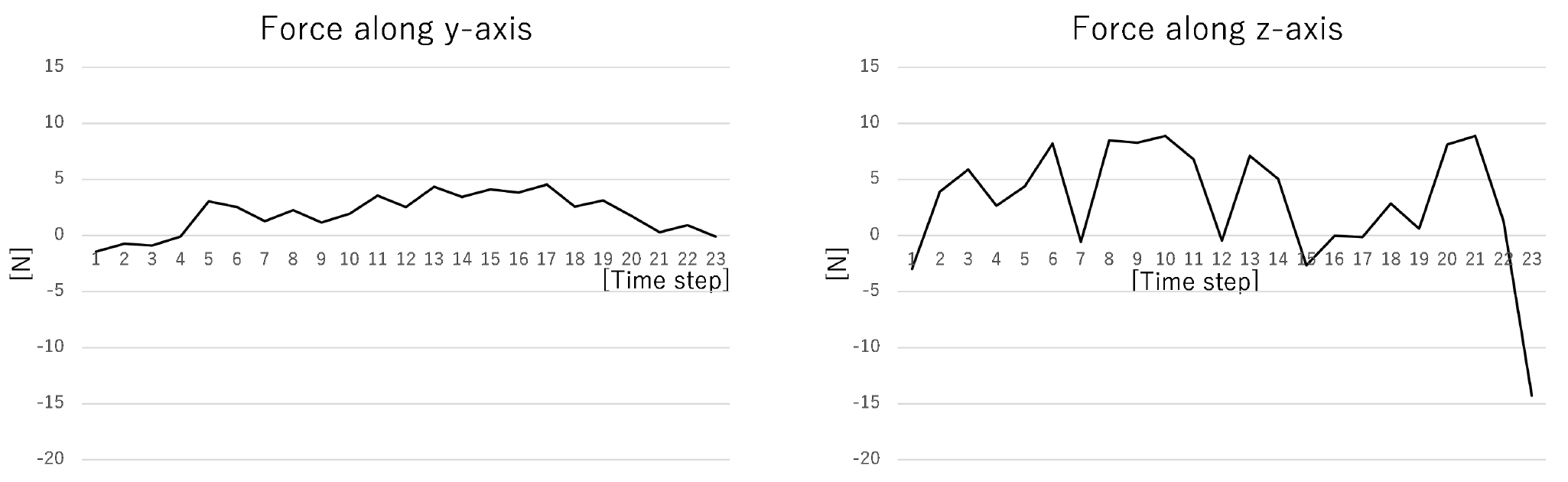}
    \caption{Force profiles along y-axis (horizontal direction) and z-axis (vertical direction) in PR-OP skill}
    \label{fig:force_profile_dc}
\end{figure}

Figure~\ref{fig:drawer_close} shows the execution of the PR-OP skill agent by an actual robot. %The drawing direction was estimated from the estimation result of the handle's 6D pose. 
We set the distance of the displacement at each step to 5 [mm]. Figure~\ref{fig:force_profile_dc2} shows the force along the drawing direction. Closing the drawer generates a large drag force, which allows us to determine the termination of the PR-OP skill. This drawer has a locking mechanism that prevents it from opening accidentally on its own. It can be seen that a large force is generated at time 21 before the lock, then the force decreases, and finally a large force is generated when the drawer is completely closed. Figure~\ref{fig:force_profile_dc} shows the changes in the force values in the horizontal and vertical directions, orthogonal to the drawing direction. Although there are some situations where unwanted force is generated during the execution, the drawing direction is modified to suppress the generation of force by the PR-OP skill agent. However, instability is confirmed because of the deviation from the physical condition of PR-PR around the time it approaches the locking mechanism (See force along the z-axis), but the PR-OP skill agent achieved the task without any troubles because of the short time from there to the closure.

\begin{comment}
図~\ref{fig:drawer_close}に実機によるPR-OPスキル実行の様子を示す。引き出しの向きは取っ手の6D poseの認識結果を用いて推定している。１タイムステップの並進量は5[mm]とした。
図~\ref{fig:force_profile_dc2}に進行方向の力の値を示す。引き出しを閉めたことによって大きな反力が発生しており、それによりPR-OP skillの終了を判定できた。ちなみにこの引き出しは勝手に開かないようなロック機構があり、ロックの手前で大きな力が発生した後、力が減少し、最後に完全に締め切った際に大きな力が発生していることが分かる。図~\ref{fig:force_profile_dc}に進行方向に直交する、水平・垂直方向の力センサの値の変化を示す。途中、不要な力が発生する場面もあるが、PR-OP（Drawer-close）のスキルに力の発生を抑えるように動作が修正されていることが分かる。ただしロック機構に差し掛かるあたりでPR-PRの物理条件から離れてしまうため不安定さが確認されるが、そこから締め切るまでの時間が短かったためにPR-OPは問題なく実現されている。
\end{comment}

\subsubsection{OR-RV (Door-open), RV-OR (Door-close), RV-RV (Door-adjust) skill agents}

These skill agents can be described by the following control rule without the termination condition.
\begin{small}
\begin{verbatim}
    if F-t > delta-collision, then penalty
    if F-u > delta-collision, then penalty
\end{verbatim}
\end{small}
All the skills need to satisfy this rule. Generally, the skill parameters (\eg, the configuration of the rotation axis) observed in the demonstration include errors. These skill agents also need to be trained using RL. As described above, the infinitesimal displacement % of the contact point relative to the environment 
at each moment can also be interpreted as an infinitesimal translation tangential to the rotation. The whole trajectory can be regarded as the pieces of the infinitesimal translation, where the translation direction is gradually changed. We can regard that the training of RV-RV is the same as that of PR-PR, since the target displacement at both skills needs to be modified following the constraint (\eg, force feedback) of the (infinitesimal) translation; the control rule is the same.

%In the figure, the constraint about the rotation axis, which is $M (\equiv Q \times N$), can be considered as a constraint to the pieces of the infinitesimal translation micro translation in the direction $N$ from $M = Q \times N$ when Q is fixed. Therefore, the learning of RV-RV can be formulated almost similarly to the PR-PR case.

\begin{comment}
なる条件を満たす必要があるという点は共通である。一般に、一般にデモから得られる回転軸方向には観測誤差がのるため、ペナルティー項を利用したスキルの強化学習が必要となる。一般に微小回転における接触点の環境に対する動きは、回転接線方向への微小並進とも解釈できる。回転軸まわりの拘束方向は、微小回転仮定のもとでは、接触点まわりの環境からの拘束方向に変換できる。図において、Mなる回転軸に関する拘束は、Qが固定された場合、cより、拘束方向$N$による微小並進への拘束と考えられる。従って、RV-RVの学習は、PR-PRの場合とほぼ同様の定式化が得られる。
\end{comment}

\begin{figure}
    \centering
    \includegraphics[width=\linewidth]{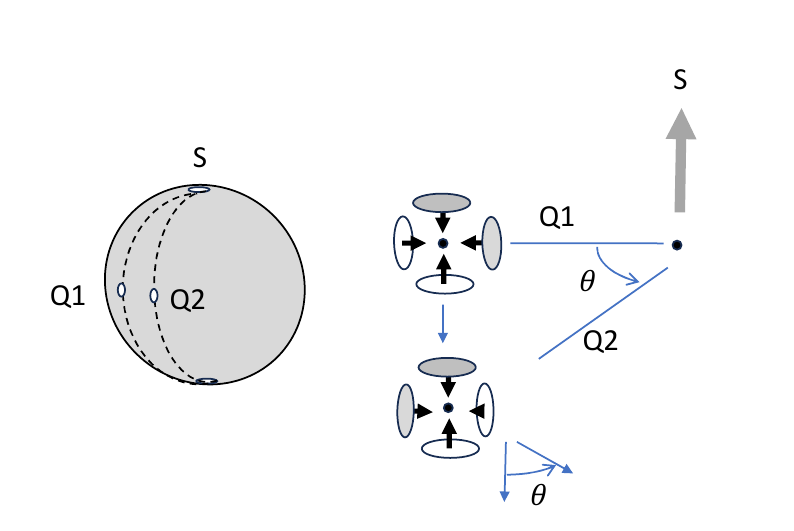}
    \caption{RV-RV transition. Q1 and Q2 are the contact points at each instant, and S is the axis of rotation. The motion direction and orientation should be corrected at each regular interval.}
    \label{fig:RV-RV-modification}
\end{figure}

The state of RV is represented by the infinitesimal feasible translation. The skill needs the force information of the feedback. The state in RL can be formulated as follows:  
\begin{itemize}
\item estimated infinitesimal translation: $ \mathbf{c}_t $,
\item normalized force: $ \mathbf{f}_t / | \mathbf{f}_t| $.
\end{itemize}
The action is also the same as that in the PR-PR skill agent as $ \Delta \mathbf{c} \equiv (\Delta c_x, \Delta c_y, \Delta c_z) $. The modification is also the same as 
\begin{displaymath}
    \mathbf{c}_{t+1} = \frac{\mathbf{c}_t + \Delta \mathbf{c}}{|\mathbf{c}_t + \Delta \mathbf{c}|}.
\end{displaymath}
The difference between the RV-RV and PR-PR skills is that the RV-RV skill involves the orientation changes with respect to the modification of the displacement. Thus, we add the orientation changes to the PR-PR skill agent; if the displacement is changed by $ \theta $, the orientation of the hand, as well as the next motion direction, are modified by $ \theta $ around the center of the grasping points. See Figure~\ref{fig:RV-RV-modification}.

\begin{figure}
    \centering
    \includegraphics[width=\linewidth]{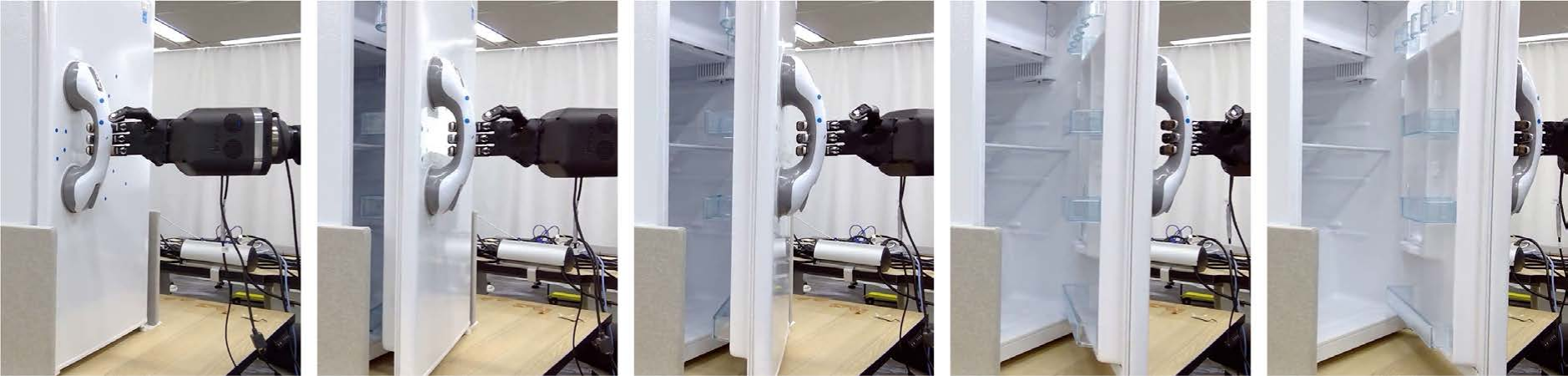}
    \caption{Execution of OR-RV (Door-opening) skill agent}
    \label{fig:door_open}
\end{figure}

\begin{figure}
    \centering
    \includegraphics[width=0.8\linewidth]{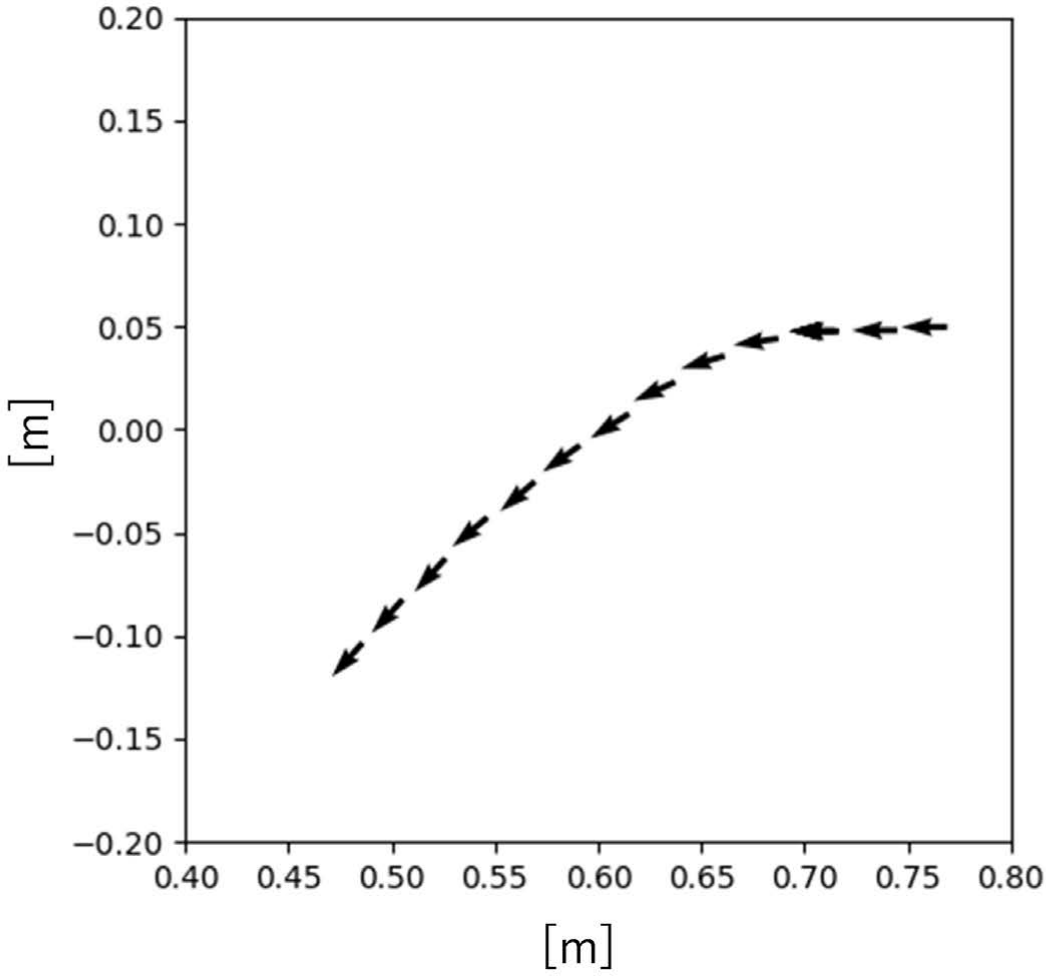}
    \caption{Trajectory of the grasping points (bottoms of arrows) and the estimated opening directions (directions of arrows)}
    \label{fig:ptg5_estimation}
\end{figure}

Figure~\ref{fig:door_open} shows the execution of the OR-RV skill agent by an actual robot. In this case, we also set the distance
of the displacement at each step to 5 [mm]. Figure~\ref{fig:ptg5_estimation} shows the trajectory of the center of the grasping points and the estimated opening directions. As can be seen, the robot tried to follows the curvilinear trajectory to open the door and the estimated opening direction is modified to follow the tangential direction of the trajectory.  

\begin{comment}
Q1とQ2は各瞬間のおけるコンタクトポイント。Sは回転軸である。一定間隔ごとに方向を修正する。
RVの状態は、可能な変位方向を指定することで表現できる。またフィードバックのため力の情報を必要とする。つまりRLにおける状態は以下のようになる。
\begin{itemize}
\item 可能な変位の方向: $ \mathbf{c}_t $
\item 力を正規化したもの: $ \mathbf{f}_t / | \mathbf{f}_t| $
\end{itemize}
ただし$ \mathbf{f}_t $は時刻$ t $における力センサの値である。
行動は現在推定された可能な変位方向に対する修正量$ \Delta \mathbf{c} \equiv (\Delta c_x, \Delta c_y, \Delta c_z) $である。実際に修正量が与えられたとき、時刻$ t + 1$の変位の方向$\mathbf{c}_{t+1}$は以下の式で計算される。
\begin{displaymath}
    \mathbf{c}_{t+1} = \frac{\mathbf{c}_t + \Delta c}{|\mathbf{c}_t + \Delta c|}
\end{displaymath}

ただし、RV-RVとPR-PRの違いとして、微小な変位では、並進と考えられても有限の大きさの変位では、回転成分が発生する。従って、図に示すように、微小並進したのち、微小回転を行うという処理を行った。
\end{comment}

\subsubsection{PC1-PC1 (Wipe) skill agent}
% PC1-PC1 skill is also named as STG2 (wiping) in~\cite{ikeuchi2021semantic}. 
The control rule is formulated as follows:
\begin{small}
\begin{verbatim}
  if F-u > delta-collision, then penalty
  if F-u < delta-zero, then penalty
  if S = goal-s AND T = goal-t, then reward
\end{verbatim}
\end{small}
Generally, the skill parameters (\eg, the surface normal) observed in the demonstration include errors. Thus, this also needs to be trained using RL.

\begin{comment}
PC1-PC1スキルは、STG22（Wiping)スキルとも呼ばれる。制御則としては、
\begin{verbatim}
        if F-u > delta-collision, then penalty
        if F-u < delta-zero, then penalty
        if S = goal-s AND T = goal-t, then reward
\end{verbatim}
とかける。一般にデモから得られた法線方向には観測誤差がのるため、PR-OPスキルの場合と同じく、ペナルティ項を利用したスキルの強化学習が必要となる。
\end{comment}

From the control rule, the force along the normal direction, F-u, should be less than delta-collision and more than delta-zero. 
Then, the PC1-PC1 skill is realized by controlling F-u to be an appropriate value $ f_c $ in between. Specifically, if the value of the force sensor just when contacting each other is $ \mathbf{f}_0 $, the target force $ \mathbf{f}_{d}$ is calculated by the following equation:
\begin{equation}
\mathbf{f}_d = \mathbf{f}_0 + (f_c - \mathbf{f}_0\cdot{\mathbf{n}}) \mathbf{n},
\end{equation}
where $ \mathbf{n} $ is the observed surface normal. The above equation will simply change the value of the force along the normal direction to $ f_c $. If the value of the force sensor at time $ t $ is $ \mathbf{f}_t $, the state of RL can be defined as follows: 
\begin{itemize}
\item surface normal: $ \mathbf{n} $,
\item target displacement direction: $ \Delta \mathbf{d}$,
\item Normalized difference between current and target forces: $ \mathbf{f}_{n} \equiv \Delta \mathbf{f}_t / |\Delta \mathbf{f}_t|$,
\end{itemize}
where $ \Delta \mathbf{f}_t \equiv \mathbf{f}_t - \mathbf{f}_0 $.
Unlike the cases of the PR-PR and RV-RV skill agents, which merely reduce the norm of the force, the PC1-PC1 skill agents need to consider the magnitude of the force in order to control F-u to approach to $f_c$. Therefore, a coarse discretization of the magnitude of $ \Delta \mathbf{f}_t $ is also included as a state. In other words, we add to the state the following element:
\begin{itemize}
\item The information on the magnitude of the force: $ f_{desk} \equiv \lfloor | \Delta \mathbf{f}_t | / f_{step} \rfloor $.
\end{itemize}
The action in RL is defined as the modification of the displacement along the normal direction, $ d_n $ and the reward function is as follows: 
\begin{equation}
r = \left\{
\begin{array}{ll}
    -f_{max} & (f_{desk} > f_{max})\\
    -f_{max} & (\hbox{the contact is detached})\\
    f_{max} / 2 - f_{desk} & (\hbox{otherwise})
\end{array}
\right..
\end{equation}
The first condition corresponds to the case of delta-collision and the second condition corresponds to the case of delta-zero. The third condition contributes to approaching to the target force. The termination condition of the episode is as follows: 
\begin{itemize}
\item F-u $ > $ delta-collision, \ie, $ f_{desk} > f_{max} $.
\item F-u $ < $ delta-zero, \ie, the contact is detached.
\item the predetermined duration $ t_{max} $ is past.
\end{itemize}
If the third condition is achieved, we regard that the skill is succeeded and $ f_{max} / 2 $ is further added to the reward. 
\begin{figure}
    \centering
    \includegraphics[width=0.8\linewidth]{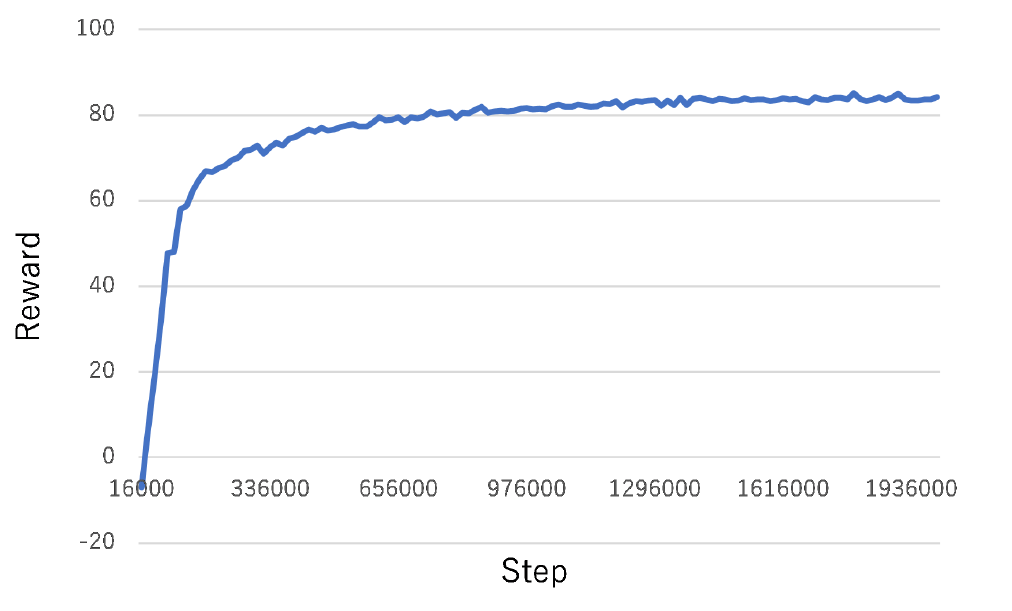}
    \caption{Reward curve in PC1-PC1 skill agent}
    \label{fig:rc_wipe}
\end{figure}
Figure~\ref{fig:rc_wipe} shows the reward curve. We finished training after two million steps. The reward looks converging.

\begin{figure}
    \centering
    \includegraphics[width=\linewidth]{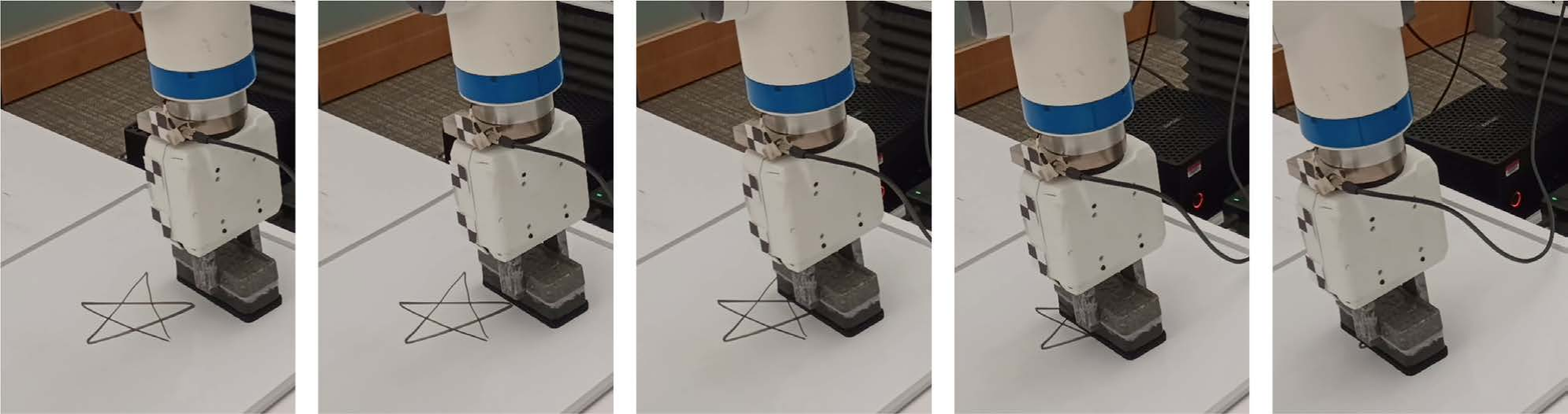}
    \caption{Execution of PC1-PC1 (Wipe) skill agents}
    \label{fig:wipe}
\end{figure}

\begin{figure}
    \centering
    \includegraphics[width=0.8\linewidth]{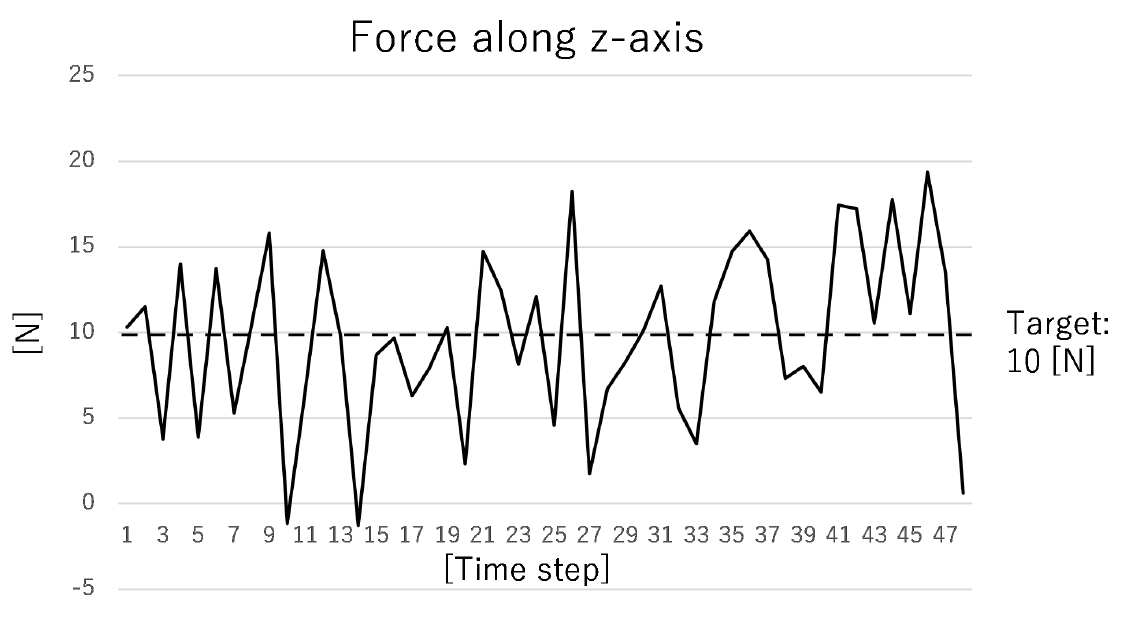}
    \caption{Force profile along z-axis in PC1-PC1 skill}
    \label{fig:force_profile_wp}
\end{figure}

\begin{figure}
    \centering
    \includegraphics[width=\linewidth]{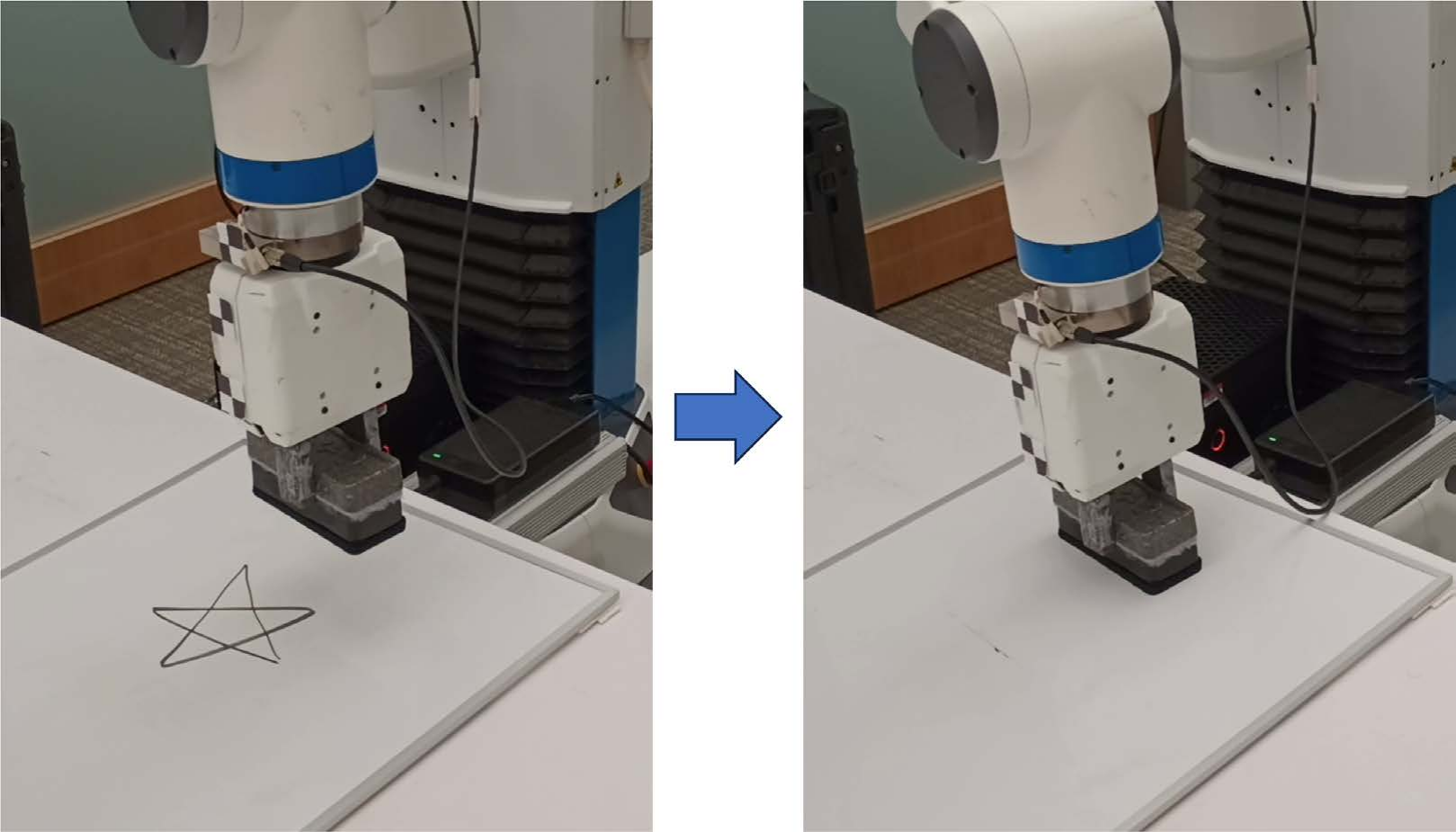}
    \caption{Success to wipe}
    \label{fig:wiping_result}
\end{figure}

Figure~\ref{fig:wipe} shows the execution of the PC1-PC1 skill agent by an actual robot. The displacement for each step is 5~[mm] in the target translational direction plus the modification outputted by the skill agent. % At each step IK calculates the robot motion from the target hand position that is obtained by adding the displacement per each step to the previous hand configuration.
It can be seen that the motion is modified to achieve the target of 10~[N]. As shown in Figure~\ref{fig:wiping_result}, the PC1-PC1 (Wipe) skill agent was actually able to erase the drawing on the white board. Theoretically, as $ f_{step} $ is increased, the degree of feedback to the force value can be reduced. Further improvement of control may be possible by adjusting $ f_{step} $.

%実際に学習されたskillを実機に接続する。その際、シミュレータと実機で物理的な環境が違うため、経験的に$ f_t = 100~[N], f_{step} = 20~[N] $とした。1タイムステップの並進量は目標並進方向へ5[mm]の変位に加え、RLが出力する補正領を加えたものである。図~\ref{fig:wipe}にPC1-PC1スキル実行の様子を示す。また図~\ref{fig:force_profile_wp}に力センサの値の変化を示す。1タイムステップあたりの変位量とskillにより出力される動作修正量を毎回の手先位置に加えIKを解くことによってロボットの動作を決定している。IKの解の精度など制御の追従性に影響を与える要因はあるものの、目標である100 [N]を実現するように動作が修正されていることが分かる。図~\ref{fig:wiping_result}に示すように、実際にWipeスキルによって白板の絵を消すことができた。理論的には$ f_{step} $を大きくすると、力センサに対するフィードバックの度合いを小さくすることができる。$ f_{step} $を調整することで更なる制御の向上が可能であるかもしれない。

\section{Working system}
This section describes how the implemented skill agents function in a reusable manner within the end-to-end Learning-from-observation system.

\begin{figure}
    \centering
    \vspace{2mm}
    \includegraphics[width=0.8\linewidth]{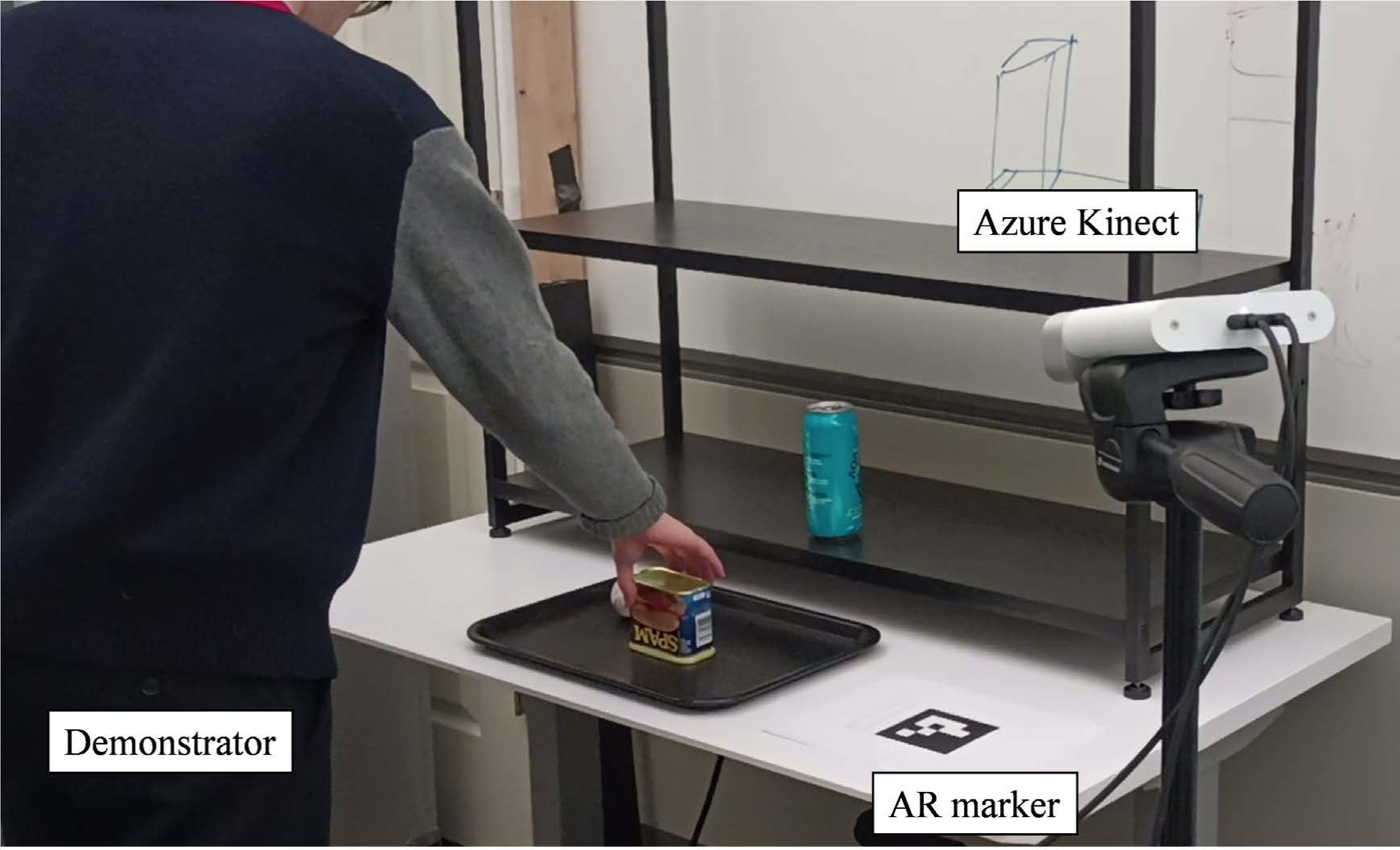}
    \caption{Observation station}
    \label{fig:demo_site}
\end{figure}

\subsection{Observation station}

Figure~\ref{fig:demo_site} shows the observation station.
For observation during the demonstration, we utilized an RGB-D camera, specifically {\em Azure Kinect} by Microsoft. To ensure alignment in the orientation between the robot coordinates and the demonstration coordinates, we employed an AR marker. This alignment enables the robot to replicate the demonstrated task sequence, achieving the same displacement while incorporating collision avoidance measures. Note that the locations of the objects do not have to be exactly the same, although they are assumed to be approximately the same during the demonstration and runtime. Any small differences are accommodated by the respective skill agents.

\begin{figure}
    \centering
    \includegraphics[width=\linewidth]{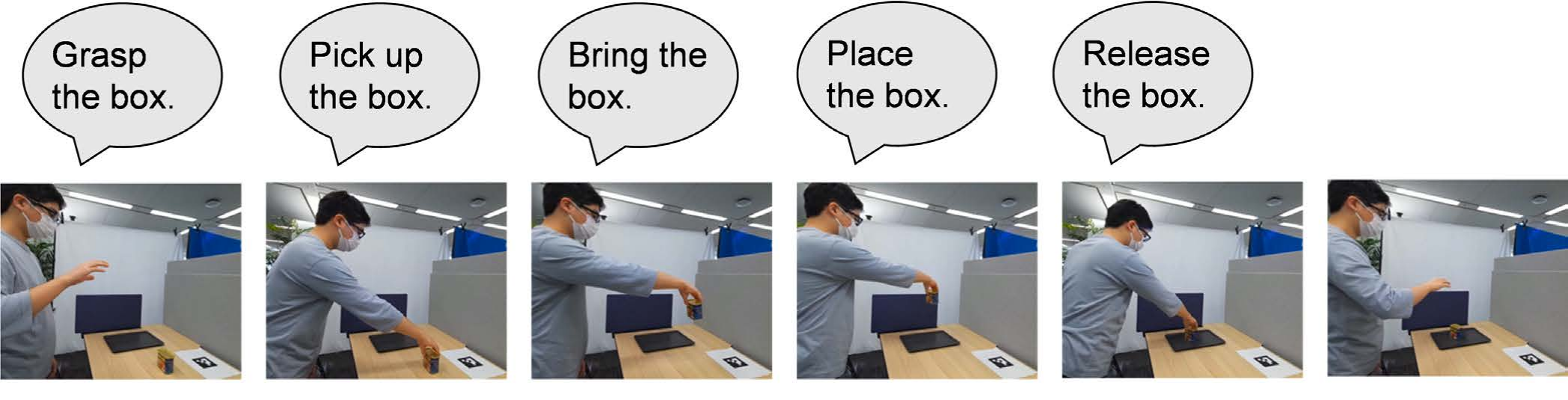}
    \caption{Overview of the demonstration}
    \label{fig:demonstration}
\end{figure}

The demonstration employed a stop-and-go approach, allowing the demonstrator to explicitly instruct the system to break down the action sequence into a task sequence. Furthermore, the demonstrator can teach the system collision avoidance paths when carrying objects by explicitly adding way points as stop motions. At each stop, the demonstrator provided verbal descriptions of the task, such as ``grasp the box" or ``pick up the box," thereby assisting the system in task recognition.

From the explanation and the demonstration image sequences, the tasks and their skill parameters can be estimated~(\cite{wake2021learning}). 
Initially, the type of each task is estimated. Subsequently, the skill parameters for each task are determined by analyzing the task sequence again using the daemon functions. %Typically, the skill parameters encompass the hand position and Labanotation pose at the beginning and end of the task. Additionally, in grasping, the approach direction are also included as the skill parameters. %In some skills such as NC-PC (PTG13 placing) task, the surface normal (assuming an upright direction in the robot coordinates, \ie, z-axis) is also the skill parameter.

\subsection{Hand motion to body motion under hardware-level reusability}

\if 0
\begin{figure}
    \centering
    %\includegraphics{}
    \caption{Caption}
    \label{fig:unexpected_output}
\end{figure}
\fi

Using IK, we transform hand motions outputted by the skill agent into body motions. A typical IK solver, such as~\cite{Beeson2015}, minimizes the difference between the desired hand pose and the pose of the hand obtained through forward kinematics under a certain joint configuration. Generally, there are multiple solutions that satisfy the target hand pose and IK sometimes outputs unexpected body pose. % as shown in Figure~\ref{fig:unexpected_output}. 
As the degrees of freedom of a robot increase, the likelihood of unexpected poses occurring becomes larger. This issue is addressed by using a Labanotation-based IK solver with the body-role division algorithm~(\cite{SasabuchiRAL2021}). 

The Labanotation-based IK solver with the body-role division avoids the unexpected poses. In Labanotation~(\cite{GuestBOOK1970,IkeuchiIJCV2018}), the pose of each limb is represented by 26 discretized directions and there is a margin to achieve a certain target Labanotation pose. The Labanotation pose is obtained from the demonstration (one of the skill parameters).
%To apply the Labanotation constraint, we manually set the joint angles to each Labanotation pose.
We solve IK as long as the joint angles do not go outside the specified range given by the Labanotation pose. 

\subsection{Robot testbed}

\begin{figure}
    \centering
    \includegraphics[width=\linewidth]{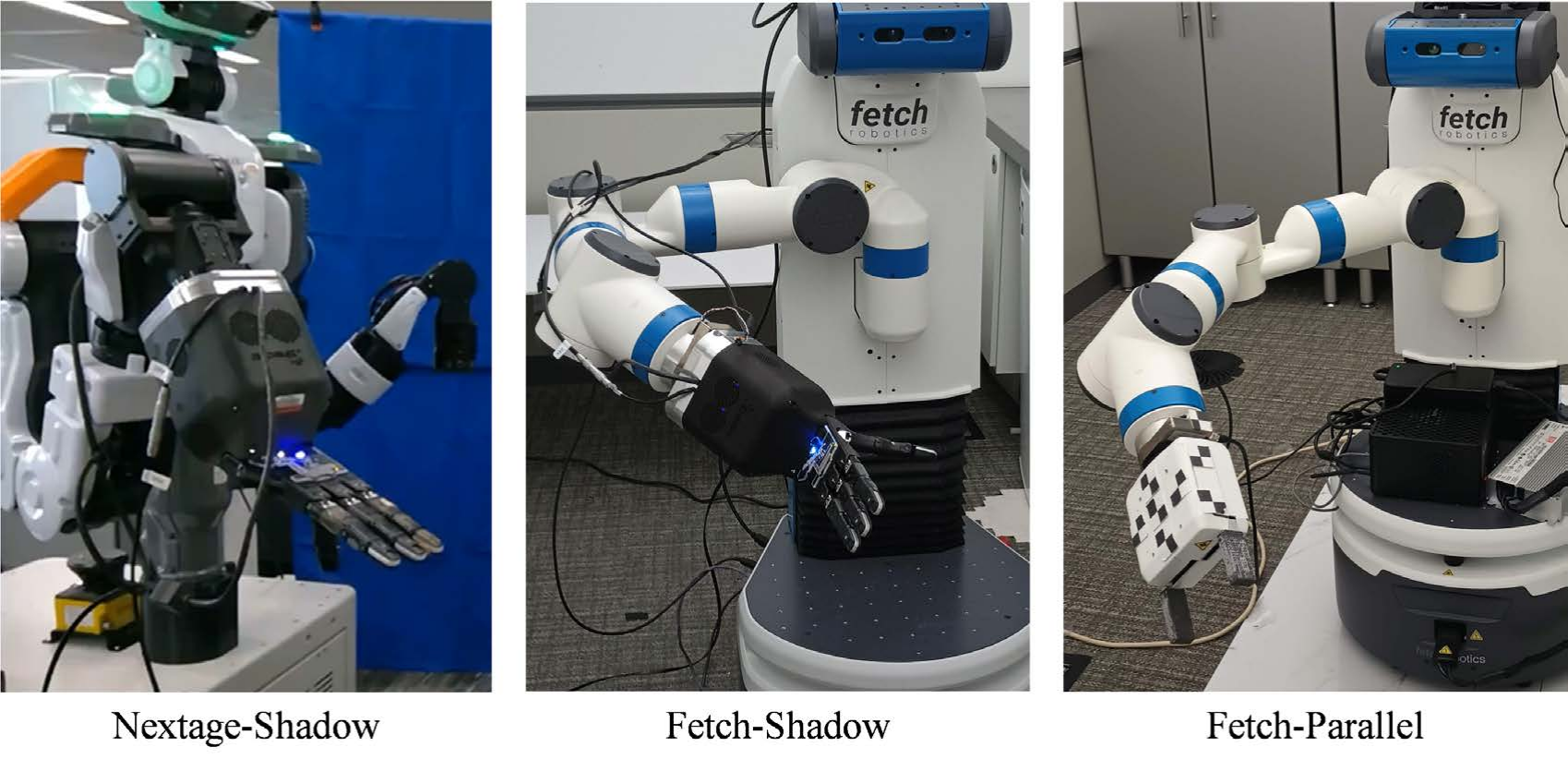}
    \caption{Three testbed robots: Nextage, Kawada Robotics (left) and Fetch Mobile Manipulator, Fetch Robotics (middle and right). The left and the middle are equipped with Shadow Dexterous Hand Lite, Shadow Robotics, as a robot hand. The right is equipped with an original parallel gripper of Fetch Mobile Manipulator.}
    \label{fig:testbed}
\end{figure}

We used three testbed robots shown in Figure~\ref{fig:testbed}. Both robots run on ROS~(\cite{Quigley2009ROSAO}). The first (refered to as Nextage-Shadow) was Nextage, Kawada Robotics\footnote{https://www.kawadarobot.co.jp/en/nextage/}. It has two arms and each arm has six DOF. It also has one DOF in the waist (rotation around the vertical axis). In this paper, we used only the right arm, neither the left arm nor waist, to perform tasks. The right arm was equipped with 6-axis force/torque sensor, FFS Series, Leptrino\footnote{https://www.leptrino.co.jp/product/6axis-force-sensor (Japanese)} and Shadow Dexterous Hand Lite, Shadow Robotics\footnote{https://www.shadowrobot.com/dexterous-hand-series/}, as a robot hand. Nextage was equipped with a stereo camera to observe an environment in a 3-dimensional space. 

The second (referred to as Fetch-Shadow) was Fetch Mobile Manipulator, Fetch Robotics\footnote{https://fetchrobotics.com/fetch-mobile-manipulator/}. It has one arm with 7 DOF, 1 DOF in the waist (moving up and down), and 2 DOF in a mobile base. It was also equipped with Leptrino FFS Series and Shadow Dexterous Hand Lite. We do not use a mobile base during manipulation. It was equipped with an RGB-D camera, Primesense Carmine 1.09, to observe an environment. The third (referred to as Fetch-Parallel) was also Fetch Mobile Manipulator. But it was equipped with an original parallel gripper in place of Shadow Dexterous Hand Lite.

\if 0
As described above, we need to set the Labanotation constraint for each robot. 
The number of degrees of Labanotation constraints should depend on the DOF of the robot. In the case of Nextage with the fixed waist joint, DOF is six in a single-arm manipulation. Even if we do not apply the Labanotation constraint, the unexpected solution of the IK seldom occurs. On the other hand, in the case of Fetch's arm with a lifter, DOF is eight. Thus, we apply the full Labanotation constraint, \ie, the constraint of four joints (two from an upper arm direction and two from a lower arm direction), especially when generating initial posture\footnote{We can apply Labanotation constraint in every IK solution. Due to the difference between the human body configuration and the robot's configuration, a Labanotation pose obtained from the demonstration may not be suitable every time. Thus, we apply the constraint when deciding the initial posture and use the current posture as the initial guess in IK after that.}. Note that though Fetch Mobile Manipulator has several joints that can rotate infinitely in the arm, the Labanotation pose can be set to limit the angle of the elbow joint to the positive direction to reduce unnecessary movement.
\fi

\section{Demonstration}

\begin{figure*}[t]
    \centering
    \includegraphics[width=0.7\linewidth]{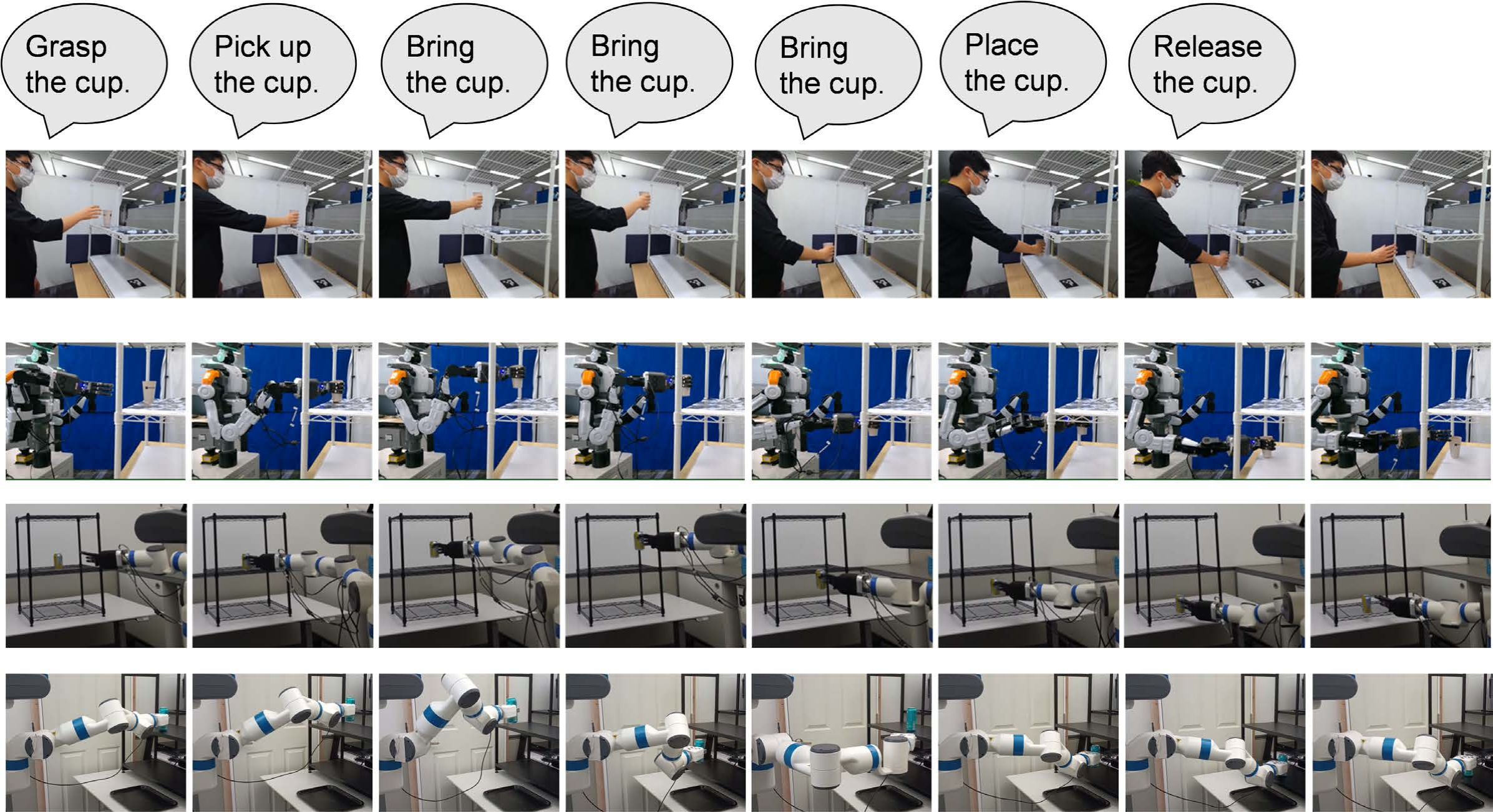}
    \caption{Shelf-sequence demo. The first row: human demonstration. The second row: execution by Nextage-Shadow. The third row: execution by Fetch-Shadow. The fourth row: execution by Fetch-Parallel.}
    \label{fig:shelf_demo}
\end{figure*}

To demonstrate hardware-level reusability in the learning-from-observation framework, we applied the four task sequences, {\em place-on-plate} demo, {\em shelf-sequence} demo, {\em throw-away} demo, and {\em open-fridge} demo to the three different robots, Nextage-Shadow, Fetch-Shadow, and Fetch-Parallel. 
%three demonstrations. 
The place-on-plate demo consists of grasping the box, picking up the box from a table, bringing the box, placing the box on a plate, and releasing the box (See Figure~\ref{fig:demonstration}). The shelf-sequence demo consists of grasping the cup, picking up the cup, three repetitions of bringing the cup, and releasing the cup (See Figure~\ref{fig:shelf_demo}). 
The throw-away demo consists of picking up the red can, bringing the can, and releasing the can (See Figure~\ref{fig:throw_away_demo}).
The open-fridge demo consists of grasping the handle and opening the fridge (see Figure~\ref{fig:open_fridge_demo}).
All the execution videos can be seen from \url{https://j-taka.github.io/research/hardware_level_reusability.html}.

\subsection{Place-on-plate Demo}

\if 0
\begin{figure}
    \centering
    %\includegraphics{}
    \caption{Obtained demonstration parameters}
    \label{fig:demonstration_parameters}
\end{figure}
\fi

\begin{figure}
    \centering
    \includegraphics[width=\linewidth]{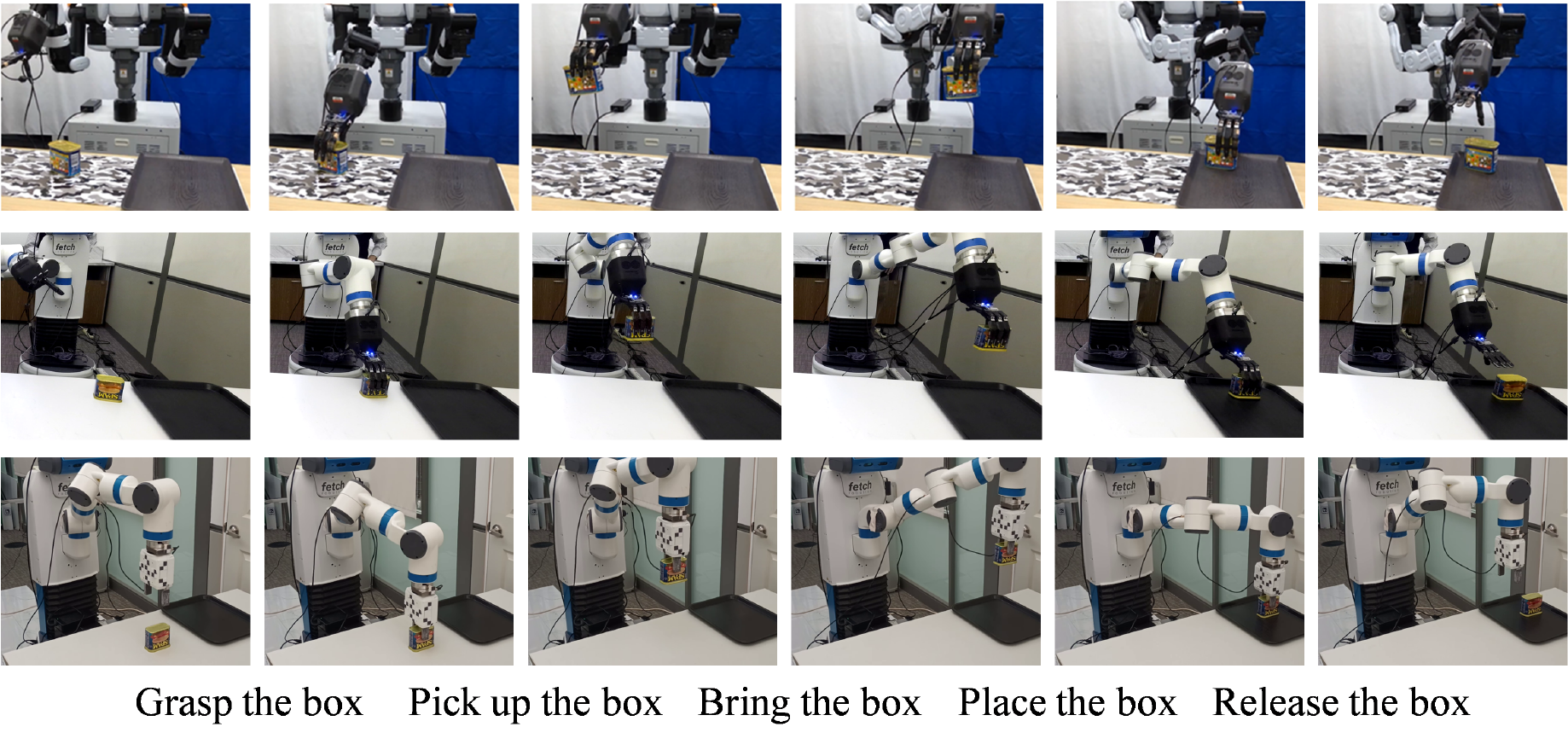}
    \caption{Place-on-plate demo. The first row: execution by Nextage-Shadow. The second row: execution by Fetch-Shadow. The third row: execution by Fetch-Parallel. These are the reproduction of the demonstration in Figure~\ref{fig:demonstration}.}
    \label{fig:place_on_plate_demo}
\end{figure}

For performing the place-on-plate demonstration, we first demonstrated the task sequence in front of Azure Kinect. As the result, the task sequence was recognized as Active-force grasp, PC-NC-a (PTG11), NC-NC (STG12), NC-PC-a (PTG13), and Release. After obtaining the task sequence, the skill parameters were estimated by observing the task sequence again. 
%Figure~\ref{fig:demonstration_parameters} shows the obtained skill parameters. 
Figure~\ref{fig:place_on_plate_demo} shows the execution by the three robots. By % giving each task the demonstration parameters and 
executing the task sequence as the same as the observed one, the three robots executed the same task sequence.  

\subsection{Shelf-sequence demo}

Figure~\ref{fig:shelf_demo} shows the result from the demonstration to the execution of the shelf-sequence demo by the three robots. The task sequence was recognized as Passive-force grasp, PC-NC-a (PTG11), three pieces of NC-NC (PTG12), NC-PC-a (PTG13), and Release. And the skill parameters were estimated by observing the task sequence as similar to the place-on-plate demo. Using the obtained skill parameters and executing the task sequence, the three robots executed the same task sequence. In this demonstration, the demonstrator used three pieces of NC-NC (PTG12) to teach the robots the trajectory for avoiding the collision with the shelf and the robots successfully avoided the collisions.

\begin{figure}
    \centering
    \includegraphics[width=\linewidth]{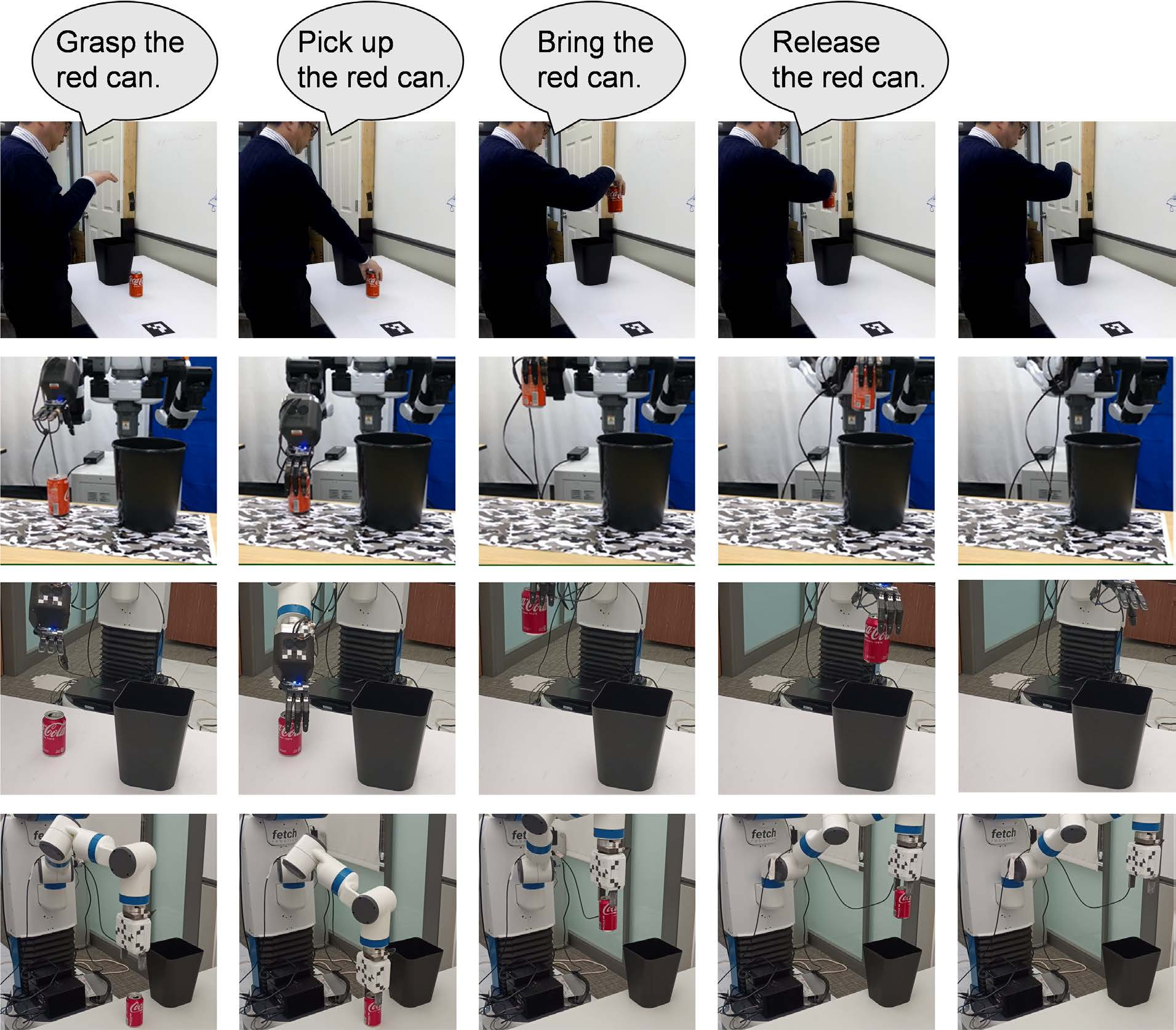}
    \caption{Throw-away demo. The first row: human demonstration. The second row: execution by Nextage-Shadow. The third row: execution by Fetch-Shadow. The fourth row: execution by Fetch-Parallel.}
    \label{fig:throw_away_demo}
\end{figure}

\subsection{Throw-away demo}

Figure~\ref{fig:throw_away_demo} shows the result from the demonstration to the execution of the throw-away demo by the three robots. The task sequence was recognized as Active-force grasp, PC-NC-a (PTG11), NC-NC (PTG12), and Release. The difference between the place-on-plate demo and the throw-away demo is to release an object before placing it. The three robots executed the same task sequence. 

\begin{figure}
    \centering
    \includegraphics[width=0.8\linewidth]{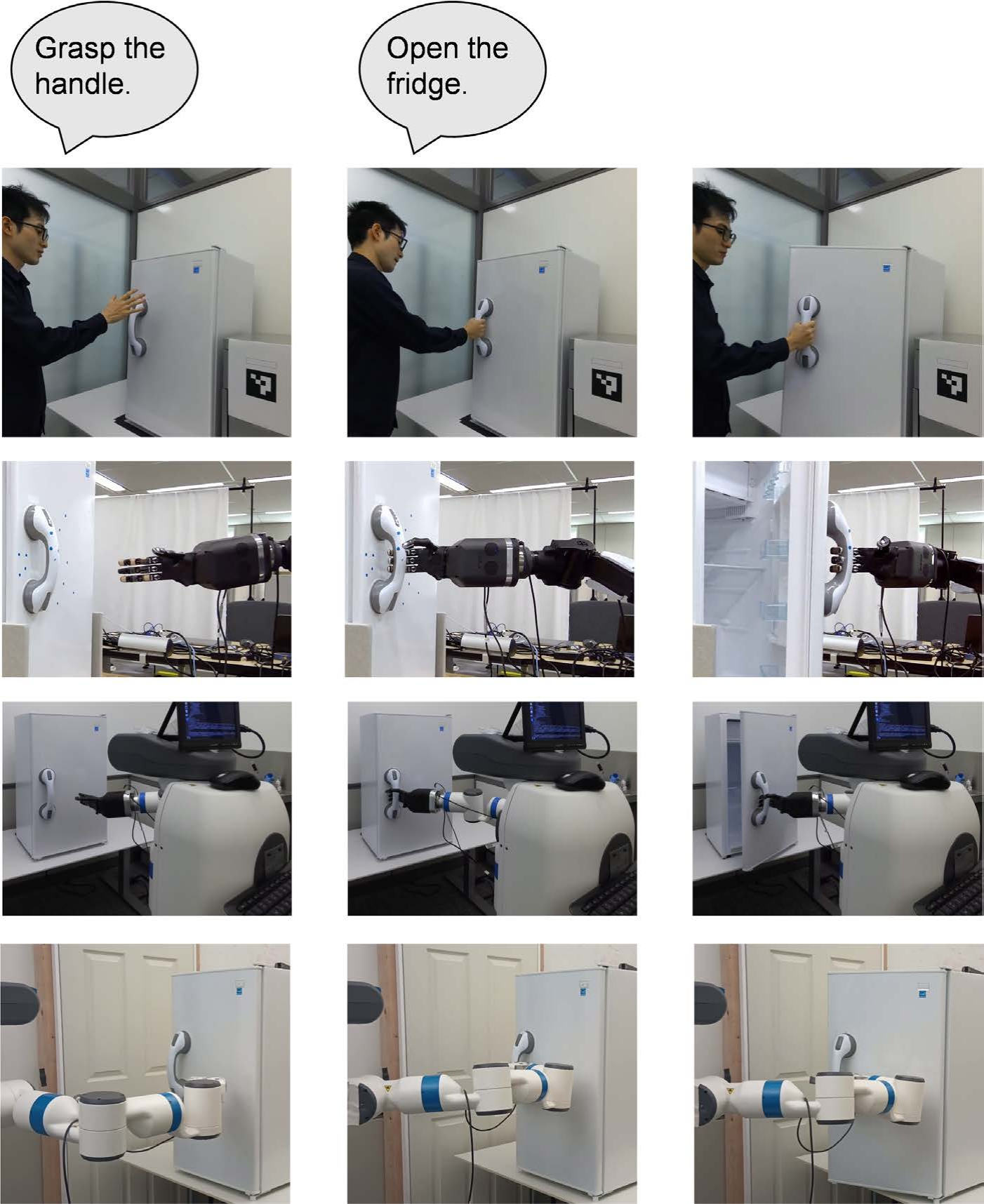}
    \caption{Open-fridge demo. The first row: human demonstration. The second row: execution by Nextage-Shadow. The third row: execution by Fetch-Shadow. The fourth row: execution by Fetch-Parallel.}
    \label{fig:open_fridge_demo}
\end{figure}

\subsection{Open-fridge demo}

Figure~\ref{fig:open_fridge_demo} shows the result from the demonstration to the execution of the open-fridge demo by the three robots. The task sequence was recognized as Lazy-closure grasp and OR-RV (PTG51). All robots succeeded in opening the door. %, although there were differences in the angle at which the door opened.

\subsection{Findings from demonstration}

\subsubsection{How reusable?}

Since the hand is the same in Nextage-Shadow and Fetch-Shadow, the implemented skill agents (both manipulation and grasping skill agents) are reusable by changing the IK solver for Nextage or Fetch. Since the hand is different in Fetch-Shadow and Fetch-Parallel, it is necessary to change the grasping skill agents. % such as Active-force grasp, Passive-force grasp, and Lazy-closure grasp. 
Fortunately, the parallel gripper has only one DOF, just opening and closing, and thus, the grasping is achieved by  
positioning the hand and closing. That is not so difficult to implement. Furthermore, the manipulation skill agents can be used as is; the finger joints are fixed in the execution of these skills and the execution is completed by body motion. 
%Series of PC-NC-a (PTG11) and NC-NC (PTG12) can be achieved by following the target trajectory, which is the IK issue. In PTG13, by replacing the part that determines the moment of placement in each hand, the remaining implementation could be used as is. 

Through the experiments, we found the following two things. First, we found that each hardware has a different range of acceptable skill parameters. For example, in the place-on-plate demo, the approach direction of the parallel gripper is changed to the upward to increase the success rate of the task sequence. Conversely, if the acceptable range is known, the success rate can be increased by adjusting the parameters without changing the nature of the target task sequence. Second, we found that it may be better not to use human grasping strategies as they are in the parallel gripper, because the parallel gripper differs significantly from a human hand in the shape and DOF. The OR-RV (PTG51) task can be achieved under satisfying the assumption where a hand and an object are integrated into one unit. Though Fetch-Parallel was able to open the fridge, the excessive feedback may be applied due to the deviation from the assumption at the beginning of the opening. Although Shadow Dexterous Hand Lite can be hooked while wrapping the handle, the parallel gripper touch the handle at small number of contact points.

\subsubsection{Comparison to related work}

\cite{pyrobot2019} proposed the open-source Robotic framework, {\em PyRobot}, which is aware of hardware independent. The hardware independent is the concept similar to the hardware-level reusability. That target robotic tasks of the framework includes manipulation and navigation. There is the difference in the target users, such as ordinary person in ours and the person who has a programming skill in theirs. In their framework, it is necessary to write a code, such as detecting the grasping point using an RGB image, set a preparation pose and the pose to grasp using the detection results, and send the command to close the hand. Thanks to the framework, each operation can be written by a one-line code, but the user needs to write a several-line code in total. Furthermore, the user needs to consider the mathematics to decide the preparation and grasping poses. When considering such things, it is required to be aware of hardware, such as which approach direction eases grasping. That reduces the hardware-level reusability of the written code.

On the other hand, the target user of the proposed system is an ordinary person, who does not have a programming skill. For such a person, it is desirable to enable to make a program for the place-to-plate demo using the instruction, such as grasping a box and pick it up. That requires the robot system to remove the effort of programming such as use of the vision and mathematics to decide the pose of the gripper. 
The proposed system provides those using the hint from the human demonstration. Using more abstracted instructions can remove awareness of the hardware. For example, issues related to approach direction in grasping are solved by the system.

\section{Summary and discussion}

This paper focused on the second step of Learning-from-observation (LfO), emphasizing the portion where the task models, obtained from demonstrations, were executed on the various robots. In line with the spirit of LfO, an effort was made to create a system that is as independent on the robot's hardware as possible. Attention was directed towards hand motions to design skill agents that move the hand to satisfy conditions derived from the task requirements and physical constraints from the environment. By taking hand motions as a reference, differences in hardware among robots are absorbed into inverse kinematics (IK), allowing the easy substitution of an appropriate IK when hardware changes, ensuring many skill agents can be used.

In environments with physical constraints, adjustments to hand trajectories are often necessary; such a concept is known as compliant manipulation (\cite{MasonITSMC1981}). Recently, reinforcement learning (RL) has been applied to address household tasks, including compliant manipulation. However, traditional RL methods have primarily focused on designing policies for specific tasks %, limiting their applicability 
and requiring individual learning for each new task. We proposed policies that consider constraints applicable to various tasks by grouping multiple tasks based on the types of physical constraints. The type of physical constraints are determined by the characteristics of the imposed force directions. Consequently, a general policy is learned based on these characteristics.

End-to-end generation of robot programs using large language models (LLM) has been proposed in general, such as~\cite{vemprala2023chatgpt, wake2023chatgpt, yu2023language}. However, many robot tasks require compliant manipulation. In the case of compliant manipulation, adjusting the hand trajectory using force feedback becomes necessary. While systems based on LLMs can provide an overview of program design, generating programs with adjusting capabilities using local feedback is challenging. To address this issue, we proposed preparing a standard set of machine-independent executable skill agents using reinforcement learning or similar methods. This standard set can be provided as a prompt to the LLMs, enabling the design of a robot system that can actually execute. We consider this approach holds promise for resolving the challenge of incorporating local feedback in robot program generation using LLMs.

In this paper, we utilized Shadow Hand on two different robots. As a result, we were able to share the same grasp skill-agent library. 
In the case of the parallel gripper of Fecth, we resolved the difference in the hardware by only changing the grasp skill-agent library. 
Standardizing grasp skill agents across different robotic hands remains a challenge for future research.

\bibliographystyle{SageH}
\bibliography{paper}

\appendix
% \section*{Appendix}

\section{Interstate transition details}
\label{sec:interstate_transition}
Analysis of the remaining part of the interstate transition is given here.

\subsection{OP-PR and PR-OP}

\paragraph{OP-PR}
A typical example of an object in OP (M=0, D=1, C=2) is a peg fully inserted into a hole with touching the bottom of the hole. The possible motion of the peg is limited to the motion of pulling it out from the hole. When the peg is slightly pulled out, it detaches from the bottom surface. In other words, the transition occurs from the detachment d-state to the maintenance d-state. A3 is applicable to this direction.
\if 0
\begin{small}
\begin{verbatim}
  A3: if F+s < delta-zero and S = goal-s,
        then reward
\end{verbatim}
\end{small}
\fi
Two dimensions orthogonal to the motion direction remain constrained both before and after the transition, requiring control to maintain the constraint d-state throughout the transition, as indicated by B3.
\if 0
\begin{small}
\begin{verbatim}
  B3: if F-t > delta-collision, then penalty
  B3: if F-u > delta-collision, then penalty
\end{verbatim}
\end{small}
\fi 

In summary, 
\begin{small}
\begin{verbatim}
  Reward OP-PR (PTG31 (Drawer-open) task)
    if F-t > delta-collision, penalty
    if F-u > delta-collision, penalty
    if F+s < delta-zero AND S = goal-s,
      then reward
\end{verbatim}
\end{small}

\paragraph{PR-OP}

An example of a transition from PR (M=1, D=0, C=2) to OP (M=0, D=1, C=2) is inserting a peg partway into a hole until it reaches the bottom of the hole, in contrast to the previous example. Concerning the direction of motion, the transition occurs from the maintenance d-state to the detachment d-state. In other words, A2 is applicable, with the termination condition being the onset of the drag force.
\if 0
\begin{small}
\begin{verbatim}
  A2: if F-s > delta-zero, then reward
\end{verbatim}
\end{small}
\fi
The two dimensions orthogonal to the motion constrained by the environment both before and after the transition. B3 is applicable, necessitating control to maintain the constraint d-state throughout the transition.
\if 0
\begin{small}
\begin{verbatim}
  B3: if F-t > delta-collision, then penalty
  B3: if F-u > delta-collision, then penalty
\end{verbatim}
\end{small}
\fi 

In summary, 
\begin{small}
\begin{verbatim}
  Reward PR-OP (PTG33 (Drawer-close) task)
    if F-t > delta-collision, penalty
    if F-u > delta-collision, penalty
    if F-s > delta-zero, reward
\end{verbatim}
\end{small}

\subsection{PR-NC and NC-PR}

\paragraph{PR-NC}
An example of a transition from PR (M=1, D=0, C=2) to NC (M=3, D=0, C=0) involves a peg partially inserted into a hole suddenly popping out of the hole. Concerning the direction of motion, it remains in the maintenance d-state before and after the transition. In other words, A1 is applicable, and the terminal condition can be defined solely based on positional information, given by the demonstration.
\if 0
\begin{small}
\begin{verbatim}
  A1: if S = goal-s, then reward
\end{verbatim}
\end{small}
\fi
Regarding the directions orthogonal to the motion, both directions are constrained before the transition and the constraints from the environment are lifted, entering the maintenance d-state after the transition. B8 is applicable.
\if 0
\begin{small}
\begin{verbatim}
  B8: if F-t > delta-collision, then penalty 
      if F-t < delta-zero and T = goal-t, 
        then reward
  B8: if F-u > delta-collision, then penalty
      if F-u < delta-zero and U = goal-u, 
        then reward
\end{verbatim}
\end{small}
\fi

In summary, 
\begin{small}
\begin{verbatim}
  Reward PR-NC
    if F-t > delta-collision, then penalty
    if F-u > delta-collision, then penalty
    if S = goal-s AND F-t < delta-zero AND 
       T = goal-t AND F-u < delta-zero AND
       U = goal-u, then reward
\end{verbatim}
\end{small}

\paragraph{NC-PR}

The transition from NC (M=3, D=0, C=0) to PR (M=1, D=0, C=2) corresponds to the opposite scenario of the previous example, where a peg in the air is suddenly inserted into a hole. Concerning the direction of motion, the maintenance d-state is maintained. In other words, A1 is applicable, and the terminal condition is defined solely based on positional information, given by the demonstration.
\if 0
\begin{small}
\begin{verbatim}
  A1: if S = goal-s, then reward
\end{verbatim}
\end{small}
\fi
Regarding the directions orthogonal to the motion, both directions are unconstrained before the transition, and after the transition, they transit into the constrained d-state due to environmental constraints. Specifically, B9 is applicable, indicating that visual feedback in those dimensions is necessary.

In summary, 
\begin{small}
\begin{verbatim}
  Reward NC-PR
    if NOT(AfterTransition):
      if |T - feature-t| > delta-gap, 
        then penalty
      if |U - feature-u| > delta-gap, 
        then penalty
    else:
      if F-t > delta-collision, then penalty
      if F-u > delta-collision, then penalty
      if S = goal-s, then reward
\end{verbatim}
\end{small}

\subsection{PR-PC and PC-PR}

\paragraph{PR-PC}
An example of a transition from PR (M=1, D=0, C=2) to PC (M=2, D=1, C=0) involves pulling a peg out from within a hole. As the peg moves, one direction orthogonal to the motion direction maintains contact with the continuous surface %, such as the bottom of the hole, 
even after leaving the hole, while the other direction becomes unconstrained and enters the maintenance d-state. In this case, the detachment d-state occurs in one direction (with the bottom surface), and the maintenance d-state is reached in the other. 

Concerning the direction of motion, it remains in the maintenance d-state, making A1 applicable. On the other hand, in dimensions orthogonal to the motion, for one dimension, there is a transition from the constraint d-state to the maintenance d-state (B8 is applicable), and for the other dimension, there is a transition from the constraint d-state to the detachment d-state (B7 is applicable).

In summary, 
\begin{small}
\begin{verbatim}
   Reward PR-PC
    if F-t > delta-collision, then penalty
    if F-u > delta-collision, then penalty
    if F-u < delta-zero, then penalty
    if S = goal-s AND F-t < delta-zero AND 
       T = goal-t, then reward
\end{verbatim}
\end{small}

\paragraph{PC-PR}

In the transition from PC (M=2, D=1, C=0) to PR (M=1, D=0, C=2), for one dimension orthogonal to the direction of motion, the maintenance d-state transits to the constraint d-state, while in the other dimension, the detachment d-state transits to the constraint d-state. This scenario could be exemplified by sliding a peg on a table, causing the peg into a hole and all faces of the peg become constrained.

The direction of motion, both before and after the transition, maintains the maintenance d-state, making A1 applicable. In one dimension orthogonal to the motion, where the detachment d-state transits to the constraint d-state, B6 is applicable. Namely, by maintaining the detachment d-state, the system automatically enters the constraint d-state. For the other dimension, the transition from the maintenance d-state to the constraint d-state occurs and B9 is applicable, indicating the need for visual feedback.

In summary,
\begin{small}
\begin{verbatim}
  Reward PC-PR
    if F-t > delta-collision, then penalty
    if F-t < delta-zero, then penalty
    if NOT(AfterTransition) AND 
       |U - feature-u| > delta-gap, 
      then penalty
    if AfterTransition AND 
       F-u > delta-collision, then penalty
    if S = goal-s, then reward    
\end{verbatim}
\end{small}

\subsection{PR-TR and TR-PR}
\paragraph{PR-TR}

The transition from PR (M=1, D=0, C=2) to TR (M=2, D=0, C=1) involves a scenario where, upon pulling the peg out of the hole, constraints in the one direction of the hole remains.
In terms of the direction of motion, it remains in the maintenance d-state, and A1 is applicable.
\if 0
\begin{small}
\begin{verbatim}
  A1: if S = goal-s, then reward
\end{verbatim}
\end{small}
\fi
For one dimension orthogonal to the direction of motion, the constraint d-state transits to the maintenance d-state, making B8 applicable.
\if 0
\begin{small}
\begin{verbatim}
  B8: if F-t > delta-collision, then penalty
      if F-t < delta-zero AND T = goal-t, 
        then reward
\end{verbatim}
\end{small}
\fi
For the other dimension, the constraint d-state is maintained, so B3 is applicable:
\if 0
\begin{small}
\begin{verbatim}
  B3: if F-u > delta-collision, then penalty
\end{verbatim}
\end{small}
\fi 

In summary, 
\begin{small}
\begin{verbatim}
  Reward PR-TR
    if F-t > delta-collision, then penalty
    if F-u > delta-collision, then penalty
    if S = goal-s AND F-t < delta-zero AND 
       T = goal-t, then reward
\end{verbatim}
\end{small}
\paragraph{TR-PR}

The transition from TR (M=2, D=0, C=1)  to PR (M=1, D=0, C=2)  maintains the maintenance d-state in the direction of motion, and A1 can be applied.
\if 0
\begin{small}
\begin{verbatim}
  A1: if S = goal-s, then reward
\end{verbatim}
\end{small}
\fi
For one dimension orthogonal to the direction of motion, there is a transition from the maintenance d-state to the constraint d-state, requiring visual feedback. In other words, B9 is applicable.
\if 0
\begin{small}
\begin{verbatim}
  B9: if NOT(AfterTransition) AND
         |T - feature-t| > delta-gap, 
        then penalty
      if AfterTransition AND 
         F-t > delta-collision,
        then penalty.
\end{verbatim}
\end{small}
\fi 
For the other dimension, it remains to be in the constraint d-state, and B3 can be applied.
\if 0
\begin{small}
\begin{verbatim}
  B3: if F-u > delta-collision, then penalty
\end{verbatim}
\end{small}
\fi 

In summary,
\begin{small}
\begin{verbatim}
  Reward TR-PR
    if F-u > delta-collision, then penalty
    if NOT(AfterTransition) AND 
       |T - feature-t| > delta-gap, 
      then penalty
    if AfterTransition AND 
       F-t > delta-collision, then penalty
    if S = goal-s, then reward
\end{verbatim}
\end{small}
\subsection{PR-OT and OT-PR}
\paragraph{PR-OT}

The transition from PR (M=1, D=0, C=2) to OT (M=1, D=1, C=1) involves the transition from the constraint d-state to the detachment d-state in one of the two constraint d-state dimensions. For example, when a peg in a hole is pulled out, a portion of one set of opposing constraint surfaces is removed, resulting in a detachment d-state, while the other dimension remains in a constraint d-state. 

Regarding the direction of motion, it remains in the maintenance d-state, and A1 can be applied.
\if 0
\begin{small}
\begin{verbatim}
  A1: if S = goal-s, then reward
\end{verbatim}
\end{small}
\fi
As for the second dimension to maintain the constraint d-state, B3 can be applied.
\if 0 
\begin{small}
\begin{verbatim}
  B3: if F-t > delta-collision, then penalty
\end{verbatim}
\end{small}
\fi 
The third dimension transits from the constraint d-state to the detachment d-state and B7 can be applied.
\if 0
\begin{small}
\begin{verbatim}
  B7: if F-u > delta-collision, then penalty
      if F-u < delta-zero, then penalty
\end{verbatim}
\end{small}
\fi

In summary,
\begin{small}
\begin{verbatim}
  Reward PR-OT
    if F-t > delta-collision, then penalty
    if F-u > delta-collision, then penalty
    if F-u < delta-zero, then penalty
    if S = goal-s, then reward
\end{verbatim}
\end{small}
\paragraph{OT-PR}

The transition from OT (M=1, D=1, C=1) to PR (M=1, D=0, C=2)  involves the dimension that was in the detachment d-state transiting to the constraint d-state. For this dimension, maintaining contact alone is sufficient to naturally transit from the detachment d-state to the constraint d-state.

As for the direction of motion, the maintenance d-state remains throughout the transition and A1 can be applied.
\if 0
\begin{small}
\begin{verbatim}
  A1: if S = goal-s, then reward
\end{verbatim}
\end{small}
\fi 
One dimension orthogonal to the motion remains in the constraint d-state and B3 can be applied.
\if 0
\begin{small}
\begin{verbatim}
  B3: if F-t > delta-collision, then penalty
\end{verbatim}
\end{small}
\fi
In the other orthogonal dimension to the motion, a transition occurs from the detachment d-state to the constraint d-state. B6 is applicable.
\if 0
\begin{small}
\begin{verbatim}
  B6: if F-u > delta-collision, then penalty
      if F-u < delta-zero, then penalty
\end{verbatim}
\end{small}
\fi

Summarizing these, we obtain:
\begin{small}
\begin{verbatim}
  Reward OT-PR
    if F-t > delta-collision, then penalty
    if F-u > delta-collision, then penalty
    if F-u < delta-zero, then penalty
    if S = goal-s, then reward
\end{verbatim}
\end{small}
\subsection{OT-NC and NC-OT}
\paragraph{OT-NC}

\begin{figure}[t]
    \centering
    \begin{tabular}{ccc}
    \begin{minipage}{0.28\hsize}
    \includegraphics[width=\linewidth]{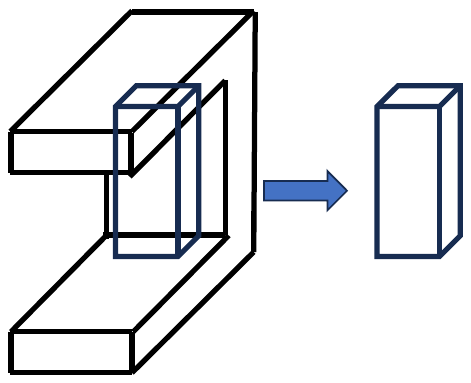}
    \end{minipage}
    &
    \begin{minipage}{0.28\hsize}
    \includegraphics[width=\linewidth]{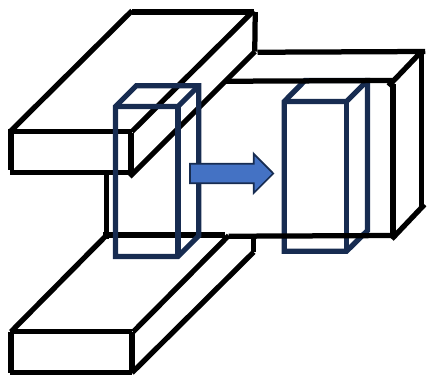}
    \end{minipage}
    &
    \begin{minipage}{0.28\hsize}
    \includegraphics[width=\linewidth]{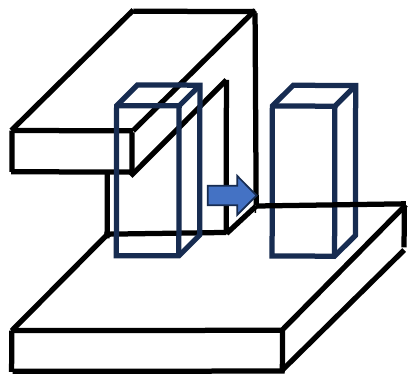}
    \end{minipage}
    \\
    (a) & (b) & (c)
    \end{tabular}
    \caption{OT transitions. (a) OT-NC transition. (b) OT-PC where the detachment dimension remains. (c) OT-PC where the constraint dimension transitions to the detachment dimension. }
    \label{fig:OT-NC-example}
\end{figure}

The transition from OT (M=1, D=1, C=1) to NC (M=3, D=0, C=0) occurs due to the shape of the environment while moving in the maintenance direction rather than the detachment direction. See Figure~\ref{fig:OT-NC-example}~(a).

In such case, concerning the direction of motion, the maintenance d-state is maintained. Therefore A1 can be applied.
\if 0
\begin{verbatim}
    A1: if S = goal-s, then reward    
\end{verbatim}
\fi
In one dimension orthogonal to the motion, a transition occurs from the detachment d-state to the maintenance d-state. Therefore, B5 can be applied.
\if 0
\begin{small}
\begin{verbatim}
  B5: if F-t > delta-collision, then penalty
      if T = goal-t, then reward
\end{verbatim}
\end{small}
\fi
In the other orthogonal dimension to the motion, the constraint d-state transits to the maintenance d-state. Therefore, B8 is applicable.
\if 0
\begin{small}
\begin{verbatim}
  B8: if F-u > delta-collision then penalty
      if F-u < delta-zero and U = goal-u, 
        the reward
\end{verbatim}
\end{small}
\fi

In summary,
\begin{small}
\begin{verbatim}
  Reward OT-NC
    if F-u > delta-collision, then penalty
    if NOT(AfterTransition):
      if F-t > delta-collision, then penalty
      if F-t < delta-zero, then penalty
    else:
      if S = goal-s AND F-t < delta-zero AND 
         T = goal-t AND F-u < delta-zero AND 
         U = goal-u, then reward
\end{verbatim}
\end{small}
\if 0
\begin{small}
\begin{verbatim}
  Reward OT-NC
    if F-t > delta-collision, then penalty
    if F-u > delta-collision, then penalty
    if S = goal-s AND T = goal-t AND 
       F-t < delta-zero AND U = goal-u AND 
       F-u < delta-zero, then reward
\end{verbatim}
\end{small}
\fi

\paragraph{NC-OT}

In contrast to the previous scenario, a transition occurs where a peg enters a hook-shaped hole that is partially open in mid-air. In two dimensions orthogonal to the motion, both the detachment d-state and the constraint d-state occur simultaneously from the maintenance d-state. 

As for the motion direction, the maintenance d-state is maintained, and A1 can be applied.
\if 0
\begin{small}
\begin{verbatim}
  A1: if S = goal-s, then reward
\end{verbatim}
\end{small}
\fi 
In one dimension orthogonal to the motion, there is a transition from the maintenance d-state to the detachment d-state and B4 can be applied; a visual sensor is required for this transition.
\if 0
\begin{small}
\begin{verbatim}
  B4: if NOT(AfterTransition):
        if |T - feature-t| > delta-gap,
          then penalty, 
      else:
        if F-t > delta-collision, 
          then penalty
        if F-t < delta-zero, then penalty.
\end{verbatim}
\end{small}
\fi 
In the other dimension orthogonal to the motion, there is a transition from the maintenance d-state to the constraint d-state and B9 can be applied. This also requires a visual sensor.
\if 0
\begin{small}
\begin{verbatim}
  B9: if NOT(AfterTransition) AND 
         |U - feature-u| > delta-gap, 
        then penalty
      if AfterTransition AND 
         F-u > delta-collision, then penalty
\end{verbatim}
\end{small}
\fi

In summary, 
\begin{small}
\begin{verbatim}
  Reward NC-OT
    if NOT(AfterTransiton):
      if |T - feature-t| > delta-gap,
        then penalty
      if |U - feature-u| > delta-gap,
        then penalty
    else:
      if F-t > delta-collision, then penalty
      if F-t < delta-zero, then penalty
      if F-u > delta-collision, then penalty
      if S = goal-s, then reward
\end{verbatim}
\end{small}

\subsection{OT-PC and PC-OT}

The transition from OT (M=1, D=1, C=1) to PC (M=2, D=1, C=0) can occur in two cases: one
where the detachment surface remains in contact while the constraint surfaces disappear, as shown in Figure~\ref{fig:OT-NC-example}~(b), and the other where a portion of constraint surfaces disappear, leading to a detachment state in this dimension, as shown in Figure~\ref{fig:OT-NC-example}~(c).

\paragraph{OT-PC-a: the detachment surface remains in contact}
Regarding the direction of motion, the maintenance d-state is maintained. Therefore A1 can be applied.
\if 0
\begin{small}
\begin{verbatim}
  A1: if S = goal-s, then reward
\end{verbatim}
\end{small}
\fi
In one dimension orthogonal to the motion, the detachment d-state is maintained. Therefore B2 can be applied.
\if 0
\begin{small}
\begin{verbatim}
  B2: if F-t > delta-collision, then penalty
      if F-t < delta-zero, then penalty
\end{verbatim}
\end{small}
\fi 
In the other orthogonal dimension, the constraint d-state transits to the maintenance d-state. Therefore B8 can be applied.
\if 0
\begin{small}
\begin{verbatim}
  B8: if F-u > delta-collision, then penalty
      if F-u < delta-zero AND U = goal-u,
        then reward
\end{verbatim}
\end{small}
\fi

In summary,
\begin{small}
\begin{verbatim}
  Reward OT-PC-a
    if F-t > delta-collision, then penalty
    if F-t < delta-zero, then penalty
    if F-u > delta-collision, then penalty
    if S = goal-s AND F-u < delta-zero AND 
       U = goal-u, then reward
\end{verbatim}
\end{small}

\paragraph{OT-PC-b: a portion of a constraint surface transits to a detachment surface}
Regarding the direction of motion, the maintenance d-state is maintained and A1 can be applied.
\if 0
\begin{small}
\begin{verbatim}
  A1: if S = goal-s, then reward
\end{verbatim}
\end{small}
\fi 
In one dimension orthogonal to the motion, the detachment d-state transits to the maintenance d-state and B5 can be applied.
\if 0
\begin{small}
\begin{verbatim}
  B5: If F-t > delta-collision, then penalty
      if T = goal-t, then reward
\end{verbatim}
\end{small}
\fi
In the other dimension, the constraint d-state transits to the detachment d-state. Therefore B7 is applicable.
\if 0
\begin{small}
\begin{verbatim}
  B7: if F-u > delta-collision, then penalty
      if F-u < delta-zero, then penalty
\end{verbatim}
\end{small}
\fi 

In summary, 
\begin{small}
\begin{verbatim}
  Reward OT-PC-b
    if F-u > delta-collision, then penalty
    if F-u < delta-zero, then penalty
    if NOT(AfterTransition):
        if F-t > delta-collision, 
          then penalty
        if F-t < delta-zero, then penalty
    else:
      if S = goal-s AND F-t < delta-zero AND
         T = goal-t, then reward
\end{verbatim}
\end{small}
\if 0
\begin{small}
\begin{verbatim}
  Reward OT-PC-b
    if F-t > delta-collision, then penalty
    if F-u > delta-collision, then penalty
    if F-u < delta-zero, then penalty
    if S = goal-s AND T = goal-t, then reward
\end{verbatim}
\end{small}
\fi 

\paragraph{PC-OT-a: the detachment surface remains in contact}
As for the transition from PC (M=2, D=1, C=0) to OT (M=1, D=1, C=1), there are also two scenarios.
In the PC-OT-a case, regarding to the direction of motion, the maintenance d-state is maintained. A1 can be applied.
\if 0
\begin{small}
\begin{verbatim}
  A1: if S = goal-s, then reward
\end{verbatim}
\end{small}
\fi
In one orthogonal direction to the motion, the detachment d-state is maintained. B2 can be applied.
\if 0
\begin{small}
\begin{verbatim}
  B2: if F-t > delta-collision,
        then penalty 
      if F-t < delta-zero, then penalty
\end{verbatim}
\end{small}
\fi 
In the other orthogonal direction, the maintenance d-state transits to the constraint d-state and B9 can be applied.
\if 0
\begin{small}
\begin{verbatim}
  B9: if NOT(AfterTransition) AND 
         |U - feature-u| > delta-gap,
        then penalty 
      if AfterTransition AND 
         F-u > delta-collision, then penalty
\end{verbatim}
\end{small}
\fi

In summary,
\begin{small}
\begin{verbatim}
  Reward PC-OT-a
    if F-t > delta-collision, then penalty
    if F-t < delta-zero, then penalty
    if NOT(AfterTransition) AND 
       |U - feature-u| > delta-gap,
      then penalty
    if AfterTransition AND 
       F-u > delta-collision, then penalty
    if S = goal-s, then reward
\end{verbatim}
\end{small}
\paragraph{PC-OT-b: the detachment d-state transits to the constraint d-state}
Regarding to the motion direction, the maintenance d-state is maintained and A1 can be applied.
\if 0
\begin{small}
\begin{verbatim}
  A1: if S = goal-s, then reward
\end{verbatim}
\end{small}
\fi 
In one orthogonal dimension to the motion, the maintenance d-state transits to the detachment d-state and B4 can be applied.
\if 0
\begin{small}
\begin{verbatim}
  B4: if NOT(AfterTransition):
        if |T - feature-t| > delta-gap,
          then penalty
      else:
        if F-t > delta-collision,
          then penalty
        if F-t < delta-zero, then penalty.
\end{verbatim}
\end{small}
\fi 
In the other orthogonal dimension to the motion, the detachment d-state transits to the constraint d-state and B6 can be applied.
\if 0 
\begin{small}
\begin{verbatim}
  B6: if F-u > delta-collision, then penalty
      if F-u < delta-zero, then penalty
\end{verbatim}
\end{small}
\fi

In summary,
\begin{small}
\begin{verbatim}
  Reward PC-OT-b
    if F-u > delta-collision, then penalty
    if F-u < delta-zero, then penalty
    if NOT(AfterTransition):
      if |T- feature-t| > delt-gap,
        then penalty
    else:
      if F-t > delta-collision, then penalty
      if F-t < delta-zero, then penalty
      if S = goal-s, then reward
\end{verbatim}
\end{small}

\subsection{OT-TR and TR-OT}

The transition from OT (M=1, D=1, C=1) to TR (M=2, D=0, C=1) also occurs through motion in two directions: motion in the detachment direction and motion along the detachment surface.
\paragraph{OT-TR-a: motion in the detachment direction}
In the motion direction, the detachment d-state transits to the maintenance d-state. A3 can be applied.
\if 0
\begin{small}
\begin{verbatim}
  A3: if F+s < delta-zero and S = goal,
    then reward
\end{verbatim}
\end{small}
\fi 
In one orthogonal dimension, the maintenance d-state is maintained and B1 can be applied.
\if 0
\begin{small}
\begin{verbatim}
  B1: if T = goal, then reward
\end{verbatim}
\end{small}
\fi
In the other orthogonal direction, the constraint d-state is maintained and B3 can be applied.
\if 0
\begin{small}
\begin{verbatim}
  B3: if F-u > delta-collision, then penalty
\end{verbatim}
\end{small}
\fi 

In summary,
\begin{small}
\begin{verbatim}
  Reward OT-TR-a
    if F-u > delta-collision, then penalty
    if F+s < delta-zero AND S = goal-s AND 
       T = goal-t, then reward
\end{verbatim}
\end{small}
\paragraph{OT-TR-b motion along the detachment surface}
In the direction of motion, the maintenance d-state is maintained. A1 can be applied.
\if 0
\begin{small}
\begin{verbatim}
  A1: if S = goal-s, then reward
\end{verbatim}
\end{small}
\fi 
In one orthogonal direction to the motion, the constraint d-state is maintained. B3 can be applied.
\if 0
\begin{small}
\begin{verbatim}
  B3: if F-t > delta-collision, then penalty
\end{verbatim}
\end{small}
\fi 
In the other orthogonal dimension, the detachment d-state transits to the maintenance d-state. B5 can be applied.
\if 0
\begin{small}
\begin{verbatim}
  B5: if F-u > delta-collision, then penalty
      if U = goal-u, then reward
\end{verbatim}
\end{small}
\fi 

In summary,
\begin{small}
\begin{verbatim}
  Reward OT-TR-b
    if F-t > delta-collision, then penalty
    if NOT(AfterTransition):
      if F-u > delta-collision, then penalty
      if F-u < delta-zero, then penalty
    else:
      if S = goal-s AND F-u < delta-zero AND
         U = goal-u, then reward
\end{verbatim}
\end{small}

\paragraph{TR-OT-a motion toward the detachment surface}
The transition from TR (M=2, D=0, C=1) to OT (M=1, D=1, C=1) also occurs in two scenario. In TR-OT-a case, along the motion direction,
the maintenance d-state transits to the detachment d-state. A2 can be applied.
\if 0
\begin{small}
\begin{verbatim}
  A2: if F-s > delta-zero, then reward
\end{verbatim}
\end{small}
\fi
In one orthogonal dimension, the maintenance d-state is maintained. B1 can be applied.
\if 0
\begin{small}
\begin{verbatim}
  B1: if T = goal-t, then reward
\end{verbatim}
\end{small}
\fi 
In the other orthogonal dimension, the constraint d-state is maintained. B3 can be applied.
\if 0
\begin{small}
\begin{verbatim}
  B3: if F-u > delta-collision, then penalty
\end{verbatim}
\end{small}
\fi

In summary,
\begin{small}
\begin{verbatim}
  Reward TR-OT-a
    if F-u > delta-collision, then penalty
    if F-s > delta-zero AND T = goal-t,
      then reward
\end{verbatim}
\end{small}

\paragraph{TR-OT-b motion along the detachment surface}
In the direction of motion, the maintenance d-state is maintained. A1 can be applied.
\if 0
\begin{small}
\begin{verbatim}
  A1: if S = goal-s, then reward
\end{verbatim}
\end{small}
\fi 
In one orthogonal direction, the constraint d-state is maintained. B3 can be applied.
\if 0
\begin{small}
\begin{verbatim}
  B3: if F-t > delta-collision, then penalty
\end{verbatim}
\end{small}
\fi 
In the other orthogonal direction, the maintenance d-state transits to the detachment d-state. B4 can be applied.
\if 0
\begin{small}
\begin{verbatim}
  B4: if NOT(AfterTransition):
        if |U - feature-u| > delta-gap, 
          then penalty
      else:
        if F-u > delta-collision, 
          then penalty
        if F-u < delta-zero, then penalty
\end{verbatim}
\end{small}
\fi 

In summary,
\begin{small}
\begin{verbatim}
  Reward TR-OT-b
    if F-t > delta-collision, then penalty
    if NOT(AfterTransition):
      if |U - feature-u| > delta-gap, 
        then penalty
    else:
      if F-u > delta-collision, then penalty
      if F-u < delta-zero, then penalty
      if S = goal-s, then reward
\end{verbatim}
\end{small}

\section{Intrastate transition details}
\subsection{TR-TR}
Regarding the transition from TR (M=2, D=0, C=1) to TR, in the direction of motion, the maintenance d-state is maintained. A1 can be applied. 
\if 0
\begin{small}
\begin{verbatim}
  A1: if S = goal-s, then reward
\end{verbatim}
\end{small}
\fi 
In one orthogonal direction to the motion, the maintenance d-state is maintained. B1 can be applied.
\if 0
\begin{small}
\begin{verbatim}
  B1: if T = goal-t, then reward
\end{verbatim}
\end{small}
\fi 
In the other orthogonal direction, the constraint d-state is maintained. B3 can be applied.
\if 0
\begin{small}
\begin{verbatim}
  B3: if F-u > delta-collision, then penalty
\end{verbatim}
\end{small}
\fi 

In summary,
\begin{small}
\begin{verbatim}
  Reward TR-TR
    if F-u > delta-collision, then penalty
    if S = goal-s AND T = goal-t,
      then reward
\end{verbatim}
\end{small}

\subsection{OT-OT}
The OT state consists of a set of two Kuhn-Tucker solution classes. OT1 (M=1, D=1, C=1) has the solution domain on a semi-circular arc on the great circle, while OT2 (M=0, D=2, C=1) has a partial arc of the great circle as the domain of solutions. %In this paper, we assume that the interstate transitions only occur from OT1.
\paragraph{OT1-OT1}

In OT1, there are two possible directions of motion: motion in the detachment direction and motion along the detachment surface. However, to transit to OT1, motion along the detachment surface is required. In this case, the d-state of motion direction is such that the maintenance d-state is preserved. A1 can be applied.
\if 0
\begin{small}
\begin{verbatim}
  A1: if S = goal-s, then reward
\end{verbatim}
\end{small}
\fi 
On the other hand, in one orthogonal direction to the motion, the detachment d-state is maintained. B2 can be applied.
\if 0
\begin{small}
\begin{verbatim}
  B2: if F-t > delta-collision, then penalty
      if F-t < delta1-zero, then penalty
\end{verbatim}
\end{small}
\fi
In the other orthogonal dimension, the constraint d-state is maintained. B3 can be applied.
\if 0
\begin{small}
\begin{verbatim}
  B3: if F-u > delta-collision, then penaly
\end{verbatim}
\end{small}
\fi 

In summary, 
\begin{small}
\begin{verbatim}
  Reward OT1-OT1
    if F-t > delta-collision, then penalty
    if F-t < delta-zero, then penalty
    if F-u > delta-collision, then penalty
    if S = goal-s, then reward
\end{verbatim}
\end{small}

\paragraph{OT1-OT2}
The region of solutions for OT1, with further constraints, becomes a partial arc, resulting in OT2. Therefore, the transition from OT1 (M=1, D=1, C=1) to OT2 (M=0, D=2, C=1) occurs in cases where motion along the detachment surface results in encountering another contact surface. Consequently, concerning the direction of motion, the maintenance d-state transits to the detachment d-state. A2 can be applied.
\if 0
\begin{verbatim}
    A2: F-s > delta-zero, then reward
\end{verbatim}
\fi
One orthogonal direction to the motion maintains the detachment d-state. B2 can be applied.
\if 0
\begin{small}
\begin{verbatim}
  B2: if F-t > delta-collision, then penalty
      if F-t < delta-zero, then penalty
\end{verbatim}
\end{small}
\fi
The other orthogonal direction to the motion maintains the constraint d-state. B3 can be applied.
\if 0
\begin{small}
\begin{verbatim}
  B3: if F-u > delta, then penaly
\end{verbatim}
\end{small}
\fi 

In summary,
\begin{small}
\begin{verbatim}
  Reward OT1-OT2
    if F-t > delta-collision, then penalty
    if F-t < delta-zero, then penalty
    if F-u > delta-collision, then penalty
    if F-s > delta-zero, then reward
\end{verbatim}
\end{small}
\paragraph{OT2-OT1}

The transition from OT2 (M=0, D=2, C=1) to OT1 (M=1, D=1, C=1) is the reverse of the previous transition, where the solution domain restricted to a partial arc, transits through detachment motion to a semi-circle region. Therefore, concerning the direction of motion, a transition from the detachment d-state to a maintenance d-state occurs. Therefore A3 can be applied.
\if 0
\begin{small}
\begin{verbatim}
  A3: if F+s < delta-zero AND S = goal-s,
        then reward
\end{verbatim}
\end{small}
\fi 
One orthogonal direction to the motion maintains the detachment d-state. B2 can be applied.
\if 0
\begin{small}
\begin{verbatim}
  B2: if F-t > delta-collision, then penalty
      if F-t < delta-zero, then penalty
\end{verbatim}
\end{small}
\fi 
The other orthogonal direction to the motion maintains the constraint d-state. B3 can be applied.
\if 0
\begin{small}
\begin{verbatim}
  B3: if F-u > delta-collision, then penaly
\end{verbatim}
\end{small}
\fi 

In summary,
\begin{small}
\begin{verbatim}
  Reward OT2-OT1
    if F-t > delta-collision, then penalty
    if F-t < delta-zero, then penalty
    if F-u > delta-collision, then penalty
    if F+s < delta-zero AND S = goal-s, 
      then reward
\end{verbatim}
\end{small}
\end{document}